\documentclass[11pt,a4paper]{article}

\usepackage{adjustbox}
\usepackage{amsfonts}
\usepackage{amssymb}
\usepackage{amsmath}
\usepackage{amsthm}
\usepackage{authblk}
\usepackage{array}
\usepackage{bbm}
\usepackage{bigdelim}
\usepackage{booktabs}
\usepackage{caption}
\usepackage{subcaption}
\usepackage{dsfont}
\usepackage{enumerate}
\usepackage{framed,multirow}
\usepackage{graphics}
\usepackage{graphicx}
\usepackage{hyperref}
\usepackage{inputenc}
\usepackage{latexsym}
\usepackage{makecell}
\usepackage{mathrsfs}  
\usepackage[table]{xcolor}
\usepackage{tikz}
\usepackage[normalem]{ulem}
\usepackage{multirow}

\usetikzlibrary{calc,fit}

\def\OOO{\mathcal{O}}

\theoremstyle{definition}

\title{Fitting and recognition of geometric primitives in segmented 3D point clouds using a localized voting procedure}

\author[ ]{Andrea Raffo}
\author[ \thanks{Corresponding author}]{Chiara Romanengo}
\author[ ]{Bianca Falcidieno}
\author[ ]{Silvia Biasotti}

\affil[ ]{Istituto di Matematica Applicata e Tecnologie Informatiche  ``E. Magenes", Consiglio Nazionale delle Ricerche, Via de Marini 6, 16149 Genova, Italy.}

\date{}                     
\newcommand\Tstrut{\rule{0pt}{2.6ex}}         
\newcommand\Bstrut{\rule[-0.9ex]{0pt}{0pt}}   

\newcolumntype{a}{>{\columncolor{blue!12}}m}
\newcolumntype{z}{>{\columncolor{teal!25}}m}

\begin{document}
\maketitle

\begin{abstract}
The automatic creation of geometric models from point clouds has numerous applications in CAD (e.g., reverse engineering, manufacturing, assembling) and, more in general, in shape modelling and processing. Given a segmented point cloud representing a man-made object, we propose a method for recognizing simple geometric primitives and their interrelationships. Our approach is based on the Hough transform (HT) for its ability to deal with noise, missing parts and outliers. In our method we introduce a novel technique for processing segmented point clouds that, through a voting procedure, is able to  provide an initial estimate of the geometric parameters characterizing each primitive type. By using these estimates, we localize the search of the optimal solution in a dimensionally-reduced parameter space thus making it efficient to extend the HT to more primitives than those that are generally found in the literature, i.e. planes and spheres. Then, we extract a number of geometric descriptors that uniquely characterize a segment, 
and, on the basis of these descriptors, we show how to aggregate parts of primitives (segments). Experiments on both synthetic and industrial scans reveal the robustness of the primitive fitting method and its effectiveness for inferring relations among segments.  \\
\textbf{Keywords}: point clouds, surface primitives, standard forms, geometric descriptors.
\end{abstract}

\section{Introduction\label{sec:intro}}
In many applications, fitting and recognition are intrinsically intertwined problems. Indeed, two aspects are involved in the identification of the surface primitives: the classification of the type of primitive that best approximates (a part of) the surface and the parameters that identify it within that type of primitive. 

The survey \cite{Kaiser2019} gives an overview on the classification and  comparison of methods developed over the years to detect simple geometric primitives in 3D data, captured from different possible sources. In their paper, the authors examine several algorithms that extract simple geometric primitives from raw dense 3D data, dividing them into three families: i) stochastic methods, such as RANSAC \cite{Schnabel2007}; ii) methods that exploit the parameter space, as the Hough-like voting methods, such as \cite{hulik2014continuous}; and iii) other clustering techniques used to discover primitives in 3D data, for instance growing primitives from a seeds or segments, such as \cite{Le:2017}. 

However, the segmentation in simple geometric primitives is often not sufficient to characterize the complexity of a model made up of multiple repeated elements, patterns and components. Following the assumptions made in \cite{Li2011} that global relationships are more stable than local relationships between point neighbours, and that man-made engineering objects are commonly rich in basic primitives (especially planes, cylinders, cones, spheres and tori, often aligned or parallel to each other), in this work we focus on the recognition of these segments in terms of a mathematical representation and the explicit presentation  of their relationships. We assume that these segments are assigned in input or possibly extracted by existing software, e.g., by using the pre-segmentations provided by other methods, such as RANSAC, \cite{Schnabel2007} or some recent learning-based approaches, \cite{LiSDYG19,ParseNet,Yang2021}.

We have observed that the general problem of segmentation and fitting of primitives is often specific to the type of primitive under consideration, and usually one must limit oneself to identifying very simple primitives because otherwise the number of parameters involved in its formulation increases too much. If we were able to conduct the primitive recognition using only its standard form, it would be possible to consider a larger number of primitives and the computational cost would be substantially reduced. For this reason, we have identified the small availability of techniques for working on a point cloud with primitives in a standard form as a weakness of the primitive fitting pipeline, and in this paper we focus primarily on these aspects rather than the general pipeline.

To address the recognition and fitting of parts of segments, we adopt an approach based on the Hough Transform (HT). Originally introduced for images in \cite{Hough1962,DH72}, this type of description is known to be robust to noise, missing parts and outliers, see for instance \cite{Mukhopadhyay2015}, and is naturally oriented towards the recognition of a mathematical expression of the primitives. 
Unfortunately, the variety of primitives that can be used within the HT framework is limited by the number of parameters of the primitive itself,  if it is considered in its generic space embedding. This fact makes the HT in space  very popular for the recognition of generic planes \cite{LIMBERGER20152043} and spheres \cite{Camurri2014}, but even for ellipsoids it is required that the cloud is centred in the origin of the Cartesian axes \cite{BELTRAMETTI2020}. 

To overcome this limitation, we propose a new method for fitting, recognizing and clustering geometric primitives in segmented 3D point clouds, which allows us: i) to automatically centre and orient a (spherical, cylindrical, conical or toric) segment so that it can be fitted with a primitive in standard form; ii) to estimate the primitive parameters so as to localize the search for the optimal solution. Once each segment has been properly roto-translated, we are able to apply the HT technique for a number of primitives otherwise non affordable in terms of computational cost and memory space occupied (see Section \ref{sec:pipeline}).  By exploiting the mathematical representation found and the parameters of each primitive, we are then able to evaluate the global relationships between the parts, identify elements that are aligned, parallel, etc., and eventually aggregate them (see Section \ref{sec:HTclustering}).

The main contributions of this paper include:
\begin{itemize}
    \item A novel pipeline to compute initial estimates for the parameters of spheres, cylinders, cones, and tori.
    \item A new technique to reduce the dimension of the parameters space and localize the search for the optimal solution, thus making the application of the HT algorithm possible to a larger number of primitives.
    \item Recognition of global relations among geometric primitives extracted from CAD objects, such as lying on the same primitive, sharing the same axis or radius, or being on parallel planes, etc..
\end{itemize}

Our implementation is currently based on MATLAB and is available as a GitHub repository\footnote{\url{https://github.com/chiararomanengo/fitting_geometric_primitives.git}}, see also \cite{Romanengo:2022}.

The remainder of this paper is organized as follows. Section \ref{sec:standardForm} lists the simple geometric primitives that will be considered throughout the paper, as well as their standard forms. In Section \ref{sec:pipeline}, the core of our paper, we first remind the reader of some preliminary notions on the Hough Transform; we then proceed to describe the pipeline of our algorithm and, finally, we provide initial examples on synthetic data. Section \ref{sec:HTclustering} exploits the geometric descriptors returned by our methodology and hierarchical clustering to recognize possible interrelationships; to test the robustness of our method, numerical simulations are performed on segmented point clouds from different datasets and  obtained by two segmentation methods. Final conclusions end the paper.

\section{List of primitives and their standard form}\label{sec:standardForm}

The expression \emph{standard form} -- sometimes referred to as \emph{canonical form} -- refers to a standard way of presenting some set of mathematical objects. A standard form is usually expected to be simpler than the elements it is equivalent to, in some way: for instance, such (simplified) expressions usually require less memory, and possess nice features which make it possible to design cleaner and more precise algorithms  \cite{Caviness:1970}.
In analytic geometry, standard forms can help study curves and surfaces. Conics and quadrics can be classified by their orbits under roto-translation. Their general implicit equation can be simplified greatly by a suitable rotation -- which eliminates mixed terms -- and a suitable translation -- which removes one or more monomials of degree $1$; from a purely geometric perspective, vertices and centers are translated to the origin of the coordinate system, while axes are aligned to the coordinate axes. When dealing with parametric representations, standard forms allow the setting of some of the parameters.  The number of parameters is reduced, in both implicit and parametric cases.  A similar argument can be applied to tori and, more in general, to surfaces of revolution.

\begin{figure}[t]
     \begin{center}
     \includegraphics[scale=0.325,trim={0cm 1cm 0cm 1cm}, clip]{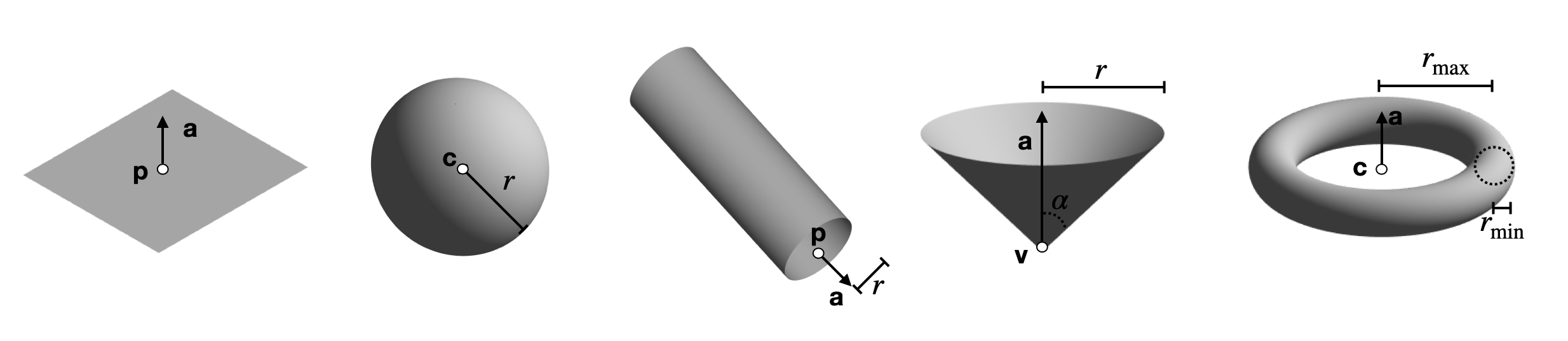}
     \end{center}
     \caption{Types of basic primitives considered in this paper: plane, cylinder, cone, sphere and torus respectively, along with their attributes (geometric descriptors).}
     \label{fig:basic_primitives}
 \end{figure}

The remainder of this section introduces the parametric equations used for cylinders, cones, spheres, and tori. For each primitive type, we specify the constraints used to obtain the corresponding standard forms. Planes are not used in standard form and are represented with the Hesse normal form, see Section \ref{sec:preproc_tangpla}.
Figure \ref{fig:basic_primitives} shows the basic primitives considered in this paper with their attributes (geometric descriptors).

\paragraph*{Spheres} The points on the sphere of radius $r>0$ and center $\mathbf{c}\in\mathbb{R}^3$ can be parametrized as
    \begin{equation*}
        \mathbf{p}(u,v)=\mathbf{c}+r\cos(v)\left(\cos(u)\mathbf{e}_1+\sin(u)\mathbf{e}_2\right)+r\sin(v)\mathbf{e}_3,
    \end{equation*}
    where $\{\mathbf{e}_1,\mathbf{e}_2,\mathbf{e}_3\}$ denotes the standard basis for $\mathbb{R}^3$. The standard form is obtained by setting $\mathbf{c}=\mathbf{0}$.

\paragraph*{Cylinders} The parametric equations of a cylinder may be written as
    \begin{equation*}
        \mathbf{p}(u,v)=\mathbf{l}+r\cos(u)\mathbf{v}_1+r\sin(u)\mathbf{v}_2+v\mathbf{a}, 
    \end{equation*}
    where: $\mathbf{l}$ is the location vector defining the base plane; $r$ is the cylinder radius; $\mathbf{a}$ is a unit vector that gives the direction of the rotational axis; $\mathbf{v}_1$ and $\mathbf{v}_2$ are chosen so that $\{\mathbf{v}_1,\mathbf{v}_2,\mathbf{a}\}$ forms an orthonormal basis. We define the standard form by setting $\mathbf{l}=\mathbf{0}$, $\mathbf{v}_i=\mathbf{e}_i$ for any $i$, and $\mathbf{a}=\mathbf{e}_3$. Note that this choice is purely individual, as one could impose any of the vectors $\mathbf{e}_i$ to be the standard rotational axis.

\paragraph*{Cones} Cones can be parametrically represented by
    \begin{equation*}
        \mathbf{p}(u,v)=\mathbf{l}+\left(r+v\sin(\alpha)\right)\left(\cos(u)\mathbf{v}_1+\sin(u)\mathbf{v}_2
        \right)+v\cos(\alpha)\mathbf{a},
    \end{equation*}
    with $\mathbf{l}$, $\mathbf{v}_i$ and $\mathbf{a}$ having the same geometric meaning as for the cylinders, while $r$ denotes the radius of the circle found by intersecting the cone and the base plane, and $\alpha$ gives the half-angle at the apex of the cone. We call standard form any parametrization obtained by imposing $\mathbf{v}_i=\mathbf{e}_i$ for any $i$, $\mathbf{a}=\mathbf{e}_3$, and by moving the cone vertex to the origin; note that the latter corresponds to set $r$ to zero (which means that the cone vertex lies on the base plane) and further impose $\mathbf{l}=\mathbf{0}$.
    
\paragraph*{Tori} Tori are parametrized as
     \begin{equation*}
        \mathbf{p}(u,v)=\mathbf{c}+\left(r_{\max}+r_{\min}\cos(v)\right)\left(\cos(u)\mathbf{v}_1+\sin(u)\mathbf{v}_2\right)+r_{\min}\sin(v)\mathbf{a},
    \end{equation*}
    where $r_{\min}$ and $r_{\max}$ are the minor and the major radii, $\mathbf{c}$ is the center of the torus, $\mathbf{a}$ is its rotational axis, and $\mathbf{v}_1$ and $\mathbf{v}_2$ are the remaining axes of the torus coordinate system. The standard forms are here expressed by setting $\mathbf{c}=\mathbf{0}$, $\mathbf{v}_i:=\mathbf{e}_i$ for any $i$, and $\mathbf{a}:=\mathbf{e}_3$.

\section{Fitting and recognizing spheres, cylinders, cones, and tori in a pre-segmented point cloud}\label{sec:pipeline}
In this section we introduce a novel method to fit and recognise geometric primitives in a pre-segmented point cloud $\mathcal{X}:=\cup_i\mathcal{P}_i$. The segmentation can be provided by any existing method: more precisely, in our experiments we use segmentations created by RANSAC \cite{Schnabel2007} and a recent learning-based approach \cite{ParseNet}; on the other hand, it is equally possible to employ other approaches. Note that RANSAC additionally provides the primitive type of a segment, but can be easily deceived, e.g., in the presence of point cloud artifacts (see, for example, \cite{Le:2017}).
We start by introducing preliminary concepts on Hough transform (see Section \ref{sec:preliminary}) and, then, describing how the generic segment $\mathcal{P}_i$ can be pre-processed (see Section \ref{sec:preproc}). Then, we  explain how such a pre-processing can be subsequently applied -- in combination with the HT -- in our fitting and recognition problem, without requiring any a-priori knowledge of the primitive type.

\subsection{Preliminary concepts on the Hough Transform}
\label{sec:preliminary}
\begin{figure}[h!]
    \centering
    \begin{tabular}{c}
     \includegraphics[width=\columnwidth]{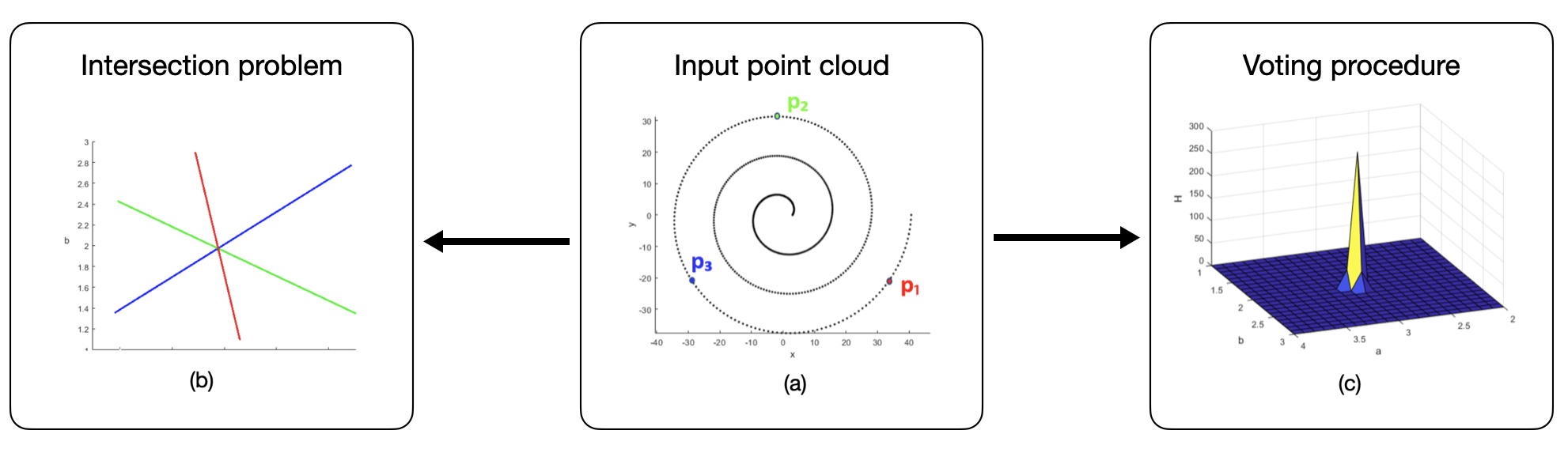}
    \end{tabular}
    \caption{The Hough paradigm illustrated for a plane curve.}
    \label{fig:duality}
\end{figure}

The core idea behind the Hough transform is to interpret the problem of shape recognition into a dual problem. To better grasp this concept, let us consider the plane point cloud shown in Figure \ref{fig:duality}(a), which is defined by an equation that depends on two parameters $a$ and $b$. Points on this curve correspond to lines in the parameter space, that meet at a single point; this intersection point uniquely identifies the coefficients of the original line equation, see Figure \ref{fig:duality}(b). While in theory it is sufficient to compute the intersection of at least $n$ Hough transforms -- being $n$ the dimension of the parameter space -- such an intersection is generally empty in practice: this occurs, for example, because of numerical reasons or for the presence of point cloud artefacts in the input data (e.g., noise, outliers). To overcome this issue, the parameter space is usually discretized into cells and a voting procedure is adopted, see Figure \ref{fig:duality}(c).

More generally, given an input point cloud and a collection of shapes, the crucial idea is to consider the parameter space defining the collection as the domain of the problem, to discretize it into cells, and to vote a cell of such a discretized space every time an HT crosses it. Practically, this translates into the voting of the most representative shape(s). 

The HT has two considerable advantages. Firstly, the adoption of a voting strategy makes it robust to most point cloud artifacts. Secondly, when the Hough-regularity property holds, it naturally provides geometric descriptors that can be used in shape analysis. However, voting strategies suffer from multiple drawbacks. As the number of parameters increases, tensor-product grids cause the memory cost to scale exponentially; moreover, they are notably affected by the curse of dimensionality which leads, in turn, to numerical instability  (see, for example, \cite{Dahyot:2009}). The surge in memory requirement is accompanied by that of the computational complexity. All these difficulties have curbed the use of the HT to families of equations with no more than $4$ parameters, making the recognition of surfaces rather unpractical. Among the various strategies based on the HT paradigm,  here we focus on the extension to general algebraic (hyper)surfaces proposed in \cite{beltrametti2012algebraic}, as it is suitable for dealing with a large set of mathematical primitives.
The algebraic HT deals with the problem of finding the surface -- within a given family of surfaces $\mathcal{S}_{\boldsymbol{\beta}}$ -- that best approximates a shape, in the form of a set of sampled points $\mathcal{P}\subset\mathbb{R}^3$; here, $\boldsymbol{\beta}$ denotes the parameter vector. The voting procedure can be summarised as follows:
\begin{enumerate}
    \item A region of the space of parameters is selected and discretized in cells which are uniquely identified by the coordinates of their center. An accumulator function is initialized.
    \item Each point in $\mathcal{P}$ gives rise to a Hough transform, i.e., a (hyper)surface in the parameter space. The accumulator value at a cell corresponds to the number of HTs that cross such a cell. 
    \item The parameter values corresponding to the maximum value of the accumulator function are chosen as the parameters of the best fitting surface.
\end{enumerate}

\subsection{Segment pre-processing\label{sec:preproc}} 
Some good initial estimate of the position in space of the point cloud would be a good starting point for (recognizing and) fitting a surface into the point cloud. For example, \cite{Le:2017} already used global estimates of the principal directions of the input data to reduce the number of primitives that can fit a point cloud.
In our case, initial estimates are fundamental for two reasons. Firstly, they allow you to put a segment in its standard form, thus reducing the number of parameters to be handled by the HT technique. Secondly, they provide a guess to the HT technique of where the optimal solution is, thus solving the problem of unboundedness of the parameter space. We now describe, for each family of geometric primitives introduced in Section \ref{sec:standardForm}, how the initial estimates can be computed. 

\subsubsection{Segment centering and normal estimation via local HT fits\label{sec:preproc_tangpla}} 
The steps described in this section are performed independently from the primitive type; they are preliminary to the computation of our initial estimates and, subsequently, to our final recognition.

\paragraph*{Point cloud centering} At first, the input point cloud $\mathcal{P}$ is centered, i.e., it is translated so that its barycenter coincides with the origin of the Cartesian coordinate system.

\paragraph*{Point cloud downsampling} The axis-aligned minimum bounding box of $\mathcal{P}$ is split into $s$ equal-sized boxes. Points within the same box $j=1,...,s$ will be replaced (i.e., downsampled) by the point which is closest to their barycenter, so as to obtain a uniformly downsampled point cloud. This step is optional, and is meant to handle very dense point clouds with a lower execution time.

\paragraph*{Normal estimation} For each $\mathbf{p}_j\in\mathcal{P}$, we select all points within a given distance $D$ w.r.t. the usual Euclidean metric; we denote this neighbourhood by $\mathcal{N}_j:=\mathcal{N}(\mathbf{p}_j,D)$. We then apply the HT technique to $\mathcal{N}_j$ and select the most voted plane $\hat{\pi}_j$, which gives an approximation of the true tangent plane $\pi_j$ at $\mathbf{p}_j$ or, equivalently, of the normal vector $\mathbf{n}_j$ at the same point.
    To this end, we consider the Hesse normal form  
    \begin{equation*}
        x\cos\theta\sin\phi+y\sin\theta\sin\phi+z\cos\phi-\rho=0,
    \end{equation*}
    where: $x$, $y$ and $z$ are the Cartesian coordinates of a sample point; $\theta\in [0,2\pi)$ and $\phi\in [0,\pi]$ are the polar coordinates of the normal vector to the plane; $\rho\in\mathbb{R}_{\geq0}$ is the distance from the plane to the origin of the coordinate system. 
    The normal at $\mathbf{p}_j$ is approximated by the vector $\hat{\mathbf{n}}_j=[\cos\hat{\theta}_j\sin\hat{\phi}_j,\sin\hat{\phi}_j\sin\hat{\theta}_j, \cos\hat{\phi}_j]$, being $\hat{\theta}_j$ and $\hat{\phi}_j$ estimates of $\theta_j$ and $\phi_j$ obtained via the Hough transform. More details on the application of the HT based on this parametrization can be found in \cite{LIMBERGER20152043}. 
\paragraph*{Normal accuracy} As the last step, we compute the accuracy of each candidate tangent plane by using the \emph{Mean Fitting Error} (MFE), defined as:
    \begin{equation}
        \text{MFE}(\mathcal{N}_j,\pi_j):=\dfrac{1}{|\mathcal{N}_j|}\sum_{\mathbf{x}\in \mathcal{N}_j}d(\mathbf{x},\hat{\pi}_j)/l_j,
        \label{eqn:MFE}
    \end{equation}
    where $d$ is the Euclidean distance and $l_j$ is the diagonal of the axis-aligned bounding box containing $\mathcal{N}_j$. From here on, we denote $\text{MFE}(\mathcal{N}_j,\pi_j)$ by $\text{MFE}_j$ and $\mathbf{MFE}:=[\text{MFE}_1, \text{MFE}_2, \ldots]$. Finally, we select the points corresponding to the lowest entries in $\mathbf{MFE}$, i.e., having the most accurate estimations of the normal vector. More precisely, the entries are selected through a threshold that depends on a percentage (a typical value is 2$\%$) of the maximum among the length, width, and height of the bounding box of $\mathcal{P}$. Let $\mathbf{p}_1, \ldots, \mathbf{p}_k$ denote such points.

\subsubsection{Initial estimates\label{sec:init_estim}}
Once the candidate tangent planes have been estimated, we specialize the processing for each type of primitive. For the sake of clarity, in this section we show the procedure on complete primitives; however, the method is also able to deal with parts of primitives, as shown in Section \ref{sec:part-noise}.

\begin{figure}[h!]
\centering
\begin{tikzpicture}

        \begin{scope}[xshift=0cm]
            \node[minimum width=2.75cm,minimum height=2.75cm,inner sep=0pt] (C1) {\includegraphics[width=2.4cm]{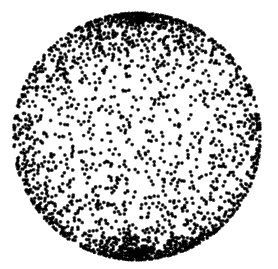}};
        \end{scope}
        
        \begin{scope}[xshift=4.0cm]
        \node[minimum width=2.75cm,minimum height=2.75cm,inner sep=0pt] (C2) {\includegraphics[width=2.6cm]{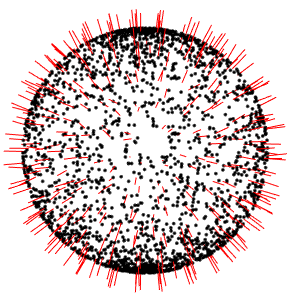}};
        \end{scope}
        
        \begin{scope}[xshift=9cm]
            \node[minimum width=3cm,minimum height=3cm,inner sep=0pt] (C3) {\includegraphics[width=3.65cm, trim={0 0 0 0.25cm}, clip]{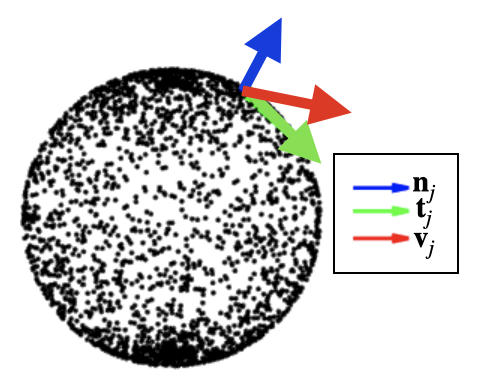}};
            \end{scope}
            
        	 \begin{scope}[xshift=14cm]
         \node[minimum width=2.75cm,minimum height=2.75cm,inner sep=0pt] (C4) {\includegraphics[width=3cm]{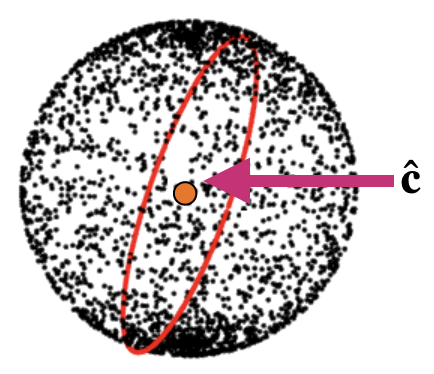}};
       	 \end{scope}
        
        \draw[very thick,->,>=stealth] ($(C1.east)+(0.2,0)$) -- ($(C2.west)+(-0.2,0)$) node [text width=2.5cm,midway,above,align=center,font=\tiny] {normal} node [text width=2.5cm,midway,below,align=center,font=\tiny] {estimation};
        \draw[very thick,->,>=stealth] ($(C2.east)+(0.2,0)$) -- ($(C3.west)+(-0.2,0)$) node [text width=2.5cm,midway,above,align=center,font=\tiny] {sphere} node [text width=2.5cm,midway,below,align=center,font=\tiny] {slicing};
        \draw[very thick,->,>=stealth] ($(C3.east)+(0.2,0)$) -- ($(C4.west)+(-0.2,0)$) node [text width=2.5cm,midway,above,align=center,font=\tiny] {center + radius} node [text width=2.5cm,midway,below,align=center,font=\tiny] {estimation};
        
        \node[align = center, below] at ($(C1.south)+(0,0.1)$) {(a)};
        \node[align = center, below] at ($(C2.south)+(0,0.1)$) {(b)};
        \node[align = center, below] at ($(C3.south)+(0,0.1)$) {(c)};
        \node[align = center, below] at ($(C4.south)+(0,0.1)$) {(d)};

        \draw[style=dashed, rounded corners] (-1.45,-2.00) rectangle (5.40,1.75); 
        \node[draw,align=left, overlay] at (2,1.1) {Pre-processing};

    	\end{tikzpicture}
	\caption{Initial estimates for a sphere. The pre-processing step centers the point cloud and approximates the normal vectors at a set of points, see (b). Given a point $\mathbf{p}_j$ and a point $\mathbf{q}_j$ on its tangent plane, (c) shows the vectors $\mathbf{n}_j$, $\mathbf{t}_j$ and $\mathbf{v}_j$ in blue, green and red, respectively. Finally, (d) shows the estimate $\mathbf{\hat{c}}$ of the center.}
	\label{fig:sphere}
\end{figure}

\paragraph*{Sphere}
For each $\mathbf{p}_j\in\mathcal{P}$, we sample a point $\mathbf{q}_j$ on the (candidate) tangent plane $\pi_j$. Then, we consider the\footnote{Despite the existence of infinitely many planes that are perpendicular to $\pi_j$, here we choose one by fixing a point on $\pi_j$ and its normal vector.} plane $\tilde{\pi}_j$ passing through $\mathbf{p}_j$ and having normal vector $\mathbf{v}_j:=\mathbf{n}_j\times{}\mathbf{t}_j$, where $\mathbf{t}_j=\mathbf{p}_j-\mathbf{q}_j$. A graphical illustration of the three vectors involved is given in Figure \ref{fig:sphere}(c): the blue, green and red vectors are, respectively, $\mathbf{n}_j$, $\mathbf{t}_j$ and $\mathbf{v}_j$. The plane $\tilde{\pi}_j$ intersects the sphere into a set of points outlining a circle, which can be recognized through the classical HT procedure for circles, and whose approximation is evaluated by the Mean Fitting Error. In exact arithmetic, $\tilde{\pi}_j$ passes through the sphere center, which corresponds to the circle center as well; in floating-point arithmetic (or when the input segment is perturbed), the center and the radius of the circle give an estimate of the center and the radius of the sphere. By repeating this procedure for all points in $\mathcal{P}$ -- or at least, for a representative subset of points -- we obtain a set of estimates of the sphere center and radius. By thresholding the MFE, we can discard low-quality estimates; finally, by averaging over the remaining centers and radii we obtain the final estimates of the sphere center and  radius, which will be denoted by $\hat{\mathbf{c}}$ and $\hat{r}$. 

\begin{figure}[h!]
\centering
\begin{tikzpicture}

        \begin{scope}[xshift=0cm]
            \node[minimum width=2.75cm,minimum height=2.75cm,inner sep=0pt] (C1) {\includegraphics[width=2.5cm,trim={2.5cm 2.5cm 2.5cm 2.5cm}, clip]{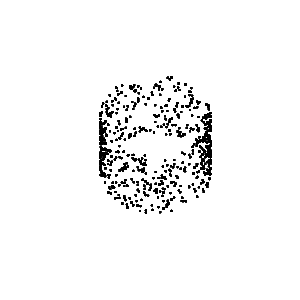}};
        \end{scope}
        
        \begin{scope}[xshift=4.0cm]
        \node[minimum width=2.75cm,minimum height=2.75cm,inner sep=0pt] (C2) {\includegraphics[width=2.50cm,trim={2.5cm 2.5cm 2.5cm 2.5cm}, clip]{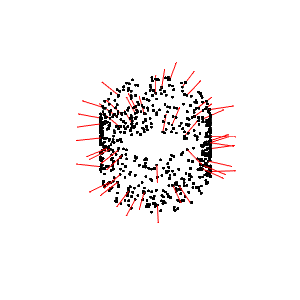}};
        \end{scope}
        
        \begin{scope}[xshift=9cm]
            \node[minimum width=2.75cm,minimum height=2.75cm,inner sep=0pt] (C3) {\includegraphics[width=3.75cm,trim={0cm 0cm 0cm 0cm}, clip]{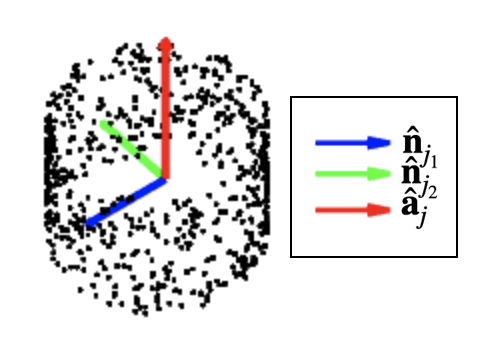}};
            \end{scope}
            
            \begin{scope}[xshift=14cm]
            \node[minimum width=2.75cm,minimum height=2.75cm,inner sep=0pt] (C4) {\includegraphics[width=2.3cm]{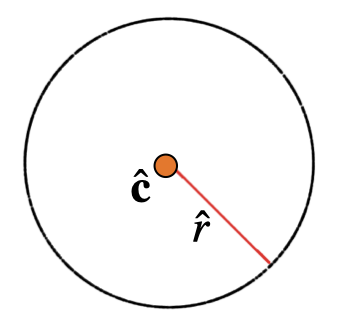}};
        \end{scope}
        
        \draw[very thick,->,>=stealth] ($(C1.east)+(0.2,0)$) -- ($(C2.west)+(-0.2,0)$) node [text width=2.5cm,midway,above,align=center,font=\tiny] {normal} node [text width=2.5cm,midway,below,align=center,font=\tiny] {estimation};
        \draw[very thick,->,>=stealth] ($(C2.east)+(0.2,0)$) -- ($(C3.west)+(-0.2,0)$) node [text width=2.5cm,midway,above,align=center,font=\tiny] {axis} node [text width=2.5cm,midway,below,align=center,font=\tiny] {estimation};
        \draw[very thick,->,>=stealth] ($(C3.east)+(0.2,0)$) -- ($(C4.west)+(-0.2,0)$) node [text width=2.5cm,midway,above,align=center,font=\tiny] {radius} node [text width=2.5cm,midway,below,align=center,font=\tiny] {estimation};
        
        \node[align = center, below] at ($(C1.south)+(0,0.25)$) {(a)};
        \node[align = center, below] at ($(C2.south)+(0,0.25)$) {(b)};
        \node[align = center, below] at ($(C3.south)+(0,0.25)$) {(c)};
        \node[align = center, below] at ($(C4.south)+(0,0.25)$) {(d)};

        \draw[style=dashed, rounded corners] (-1.45,-1.75) rectangle (5.40,1.5); 
        \node[draw,align=left, overlay] at (2,1.1) {Pre-processing};
        
    \end{tikzpicture}
    \caption{Initial estimates for a circular cylinder. The pre-processing step centers the point cloud and approximates the normal vectors at a set of points, see (b). Given two points $\mathbf{p}_{j_1}$ and $\mathbf{p}_{j_2}$, (c) shows the corresponding normal vectors $\hat{\mathbf{n}}_{j_1}$ and $\hat{\mathbf{n}}_{j_2}$, in blue and green respectively, and, in red, their cross product $\hat{\mathbf{a}}_{j_1,j_2}$. Finally, (d) shows the estimate $\hat{r}$ of the radius.}
    \label{fig:cylinder}
    \end{figure}

\paragraph*{Cylinder}
For each pair of points $\mathbf{p}_{j_1}$ and $\mathbf{p}_{j_2}$, where $j_1,j_2=1,\ldots,k$ and $j_1\ne{}j_2$, we consider the corresponding normal vectors $\hat{\mathbf{n}}_{j_1}$, $\hat{\mathbf{n}}_{j_2}$: their cross product, denoted $\hat{\mathbf{a}}_{j_1,j_2}:=\hat{\mathbf{n}}_{j_1}\times{}\hat{\mathbf{n}}_{j_2}$, is an approximation of the rotational axis of the cylinder up to a translation. Figure \ref{fig:cylinder}(c) represents a simplified situation, where the two normals (in green and blue) and their cross product (in red) are positioned so that $\hat{\mathbf{a}}_{j_1,j_2}$ determines the rotational axis; note that this choice is purely illustrative.  By iterating over all possible combinations, one can obtain multiple estimates of the rotational axis; we average over all these estimates and return the resulting vector, denoted $\hat{\mathbf{a}}$.

The point cloud $\mathcal{P}$ is now rotated so that $\hat{\mathbf{a}}$ is parallel to the $z-axis$ and, subsequently, projected onto the $xy$-plane. To estimate the radius $r$ and the center $\mathbf{c}$ of the projected points, which outline (arcs of) a circle if the initial point cloud originated from a circular cylinder, we detect the most voted circle by applying the HT-based recognition process, see Figure \ref{fig:cylinder}(d).

\paragraph*{Cone}
From basic geometry we know that, in exact arithmetic, the vertex of a cone can be found by intersecting (at least) three tangent planes. In case of data perturbation, however, such an intersection will be most likely empty. To overcome this problem, we define a voting procedure that exploits the representations of the tangent planes of the points $\mathbf{p}_1, \ldots, \mathbf{p}_k$. More specifically, since for each tangent plane $\pi_j$ the normal $\mathbf{n}_j$ and the term $\rho_j$ are known, the voting procedure considers the coordinates $x$, $y$ and $z$ of the vertex as the parameters to be estimated. The most voted coordinates correspond to the vertex $\hat{\mathbf{v}}$.

\begin{figure}[h!]
\centering
\begin{tikzpicture}

        \begin{scope}[xshift=0cm]
            \node[minimum width=2.75cm,minimum height=2.75cm,inner sep=0pt] (C1) {\includegraphics[width=2.8cm]{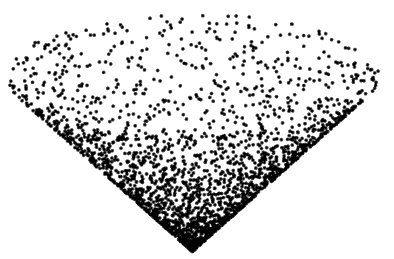}};
        \end{scope}
        
        \begin{scope}[xshift=4.0cm]
        \node[minimum width=2.75cm,minimum height=2.75cm,inner sep=0pt] (C2) {\includegraphics[width=3cm]{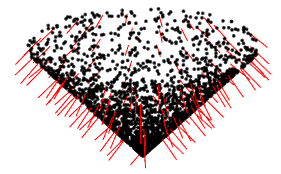}};
        \end{scope}
        
        \begin{scope}[xshift=9.5cm]
            \node[minimum width=2.85cm,minimum height=2.75cm,inner sep=0pt] (C3) {\includegraphics[width=4.2cm]{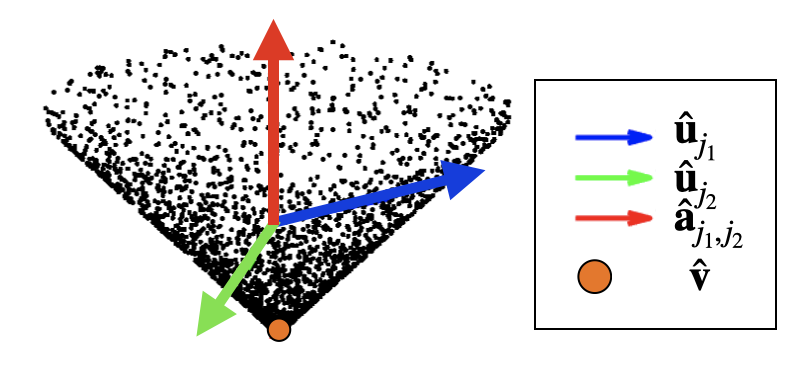}};
            \end{scope}
        
        \draw[very thick,->,>=stealth] ($(C1.east)+(0.2,0)$) -- ($(C2.west)+(-0.2,0)$) node [text width=2.5cm,midway,above,align=center,font=\tiny] {normal} node [text width=2.5cm,midway,below,align=center,font=\tiny] {estimation};
        \draw[very thick,->,>=stealth] ($(C2.east)+(0.2,0)$) -- ($(C3.west)+(-0.2,0)$) node [text width=2.5cm,midway,above,align=center,font=\tiny] {axis and vertex} node [text width=2.5cm,midway,below,align=center,font=\tiny] {estimation};
        
        \node[align = center, below] at ($(C1.south)+(0,0.3)$) {(a)};
        \node[align = center, below] at ($(C2.south)+(0,0.3)$) {(b)};
        \node[align = center, below] at ($(C3.south)+(0,0.3)$) {(c)};
        
        \draw[style=dashed, rounded corners] (-1.45,-1.75) rectangle (5.40,1.5); 
        \node[draw,align=left, overlay] at (2,1.1) {Pre-processing};
    \end{tikzpicture}
    \caption{Initial estimates for a circular cone. The pre-processing step centers the point cloud and approximates the normal vectors at a set of points, see (b). The intersection of the tangent planes estimates the coordinates of the vertex $\hat{\mathbf{v}}$. Given two points $\mathbf{p}_{j_1}$ and $\mathbf{p}_{j_2}$, in (d) the corresponding blue and green vectors $\hat{\mathbf{u}}_{j_1}$ and $\hat{\mathbf{u}}_{j_2}$ and the resulting cross product $\hat{\mathbf{a}}_{j_1,j_2}$ in red.}
    \label{fig:cone}
    \end{figure}

Then, for each pair of points $\mathbf{p}_{j_1}$ and $\mathbf{p}_{j_2}$, where $j_1,j_2=1,\ldots,k$ and $j_1\ne{}j_2$, we consider the corresponding normal vectors $\hat{\mathbf{n}}_{j_1}$, $\hat{\mathbf{n}}_{j_2}$ and the vectors $\hat{\mathbf{t}}_{j_1}:=\mathbf{p}_{j_1}-\hat{\mathbf{v}}$, $\hat{\mathbf{t}}_{j_2}:=\mathbf{p}_{j_2}-\hat{\mathbf{v}}$. We compute the cross products  $\hat{\mathbf{u}}_{j_1}=\hat{\mathbf{n}}_{j_1}\times{}\hat{\mathbf{t}}_{j_1}$ and $\hat{\mathbf{u}}_{j_2}=\hat{\mathbf{n}}_{j_2}\times{}\hat{\mathbf{t}}_{j_2}$. By taking the cross product between $\hat{\mathbf{u}}_{j_1}$ and $\hat{\mathbf{u}}_{j_2}$ we obtain an estimate $\hat{\mathbf{a}}_{j_1,j_2}$ of the rotational axis of the cone, up to a translation by the cone vertex. A simplified graphical illustration, where a triplet of vectors $\hat{\mathbf{n}}_{j}$ (in blue),  $\hat{\mathbf{t}}_{j}$ (in green) and $\hat{\mathbf{a}}_{j_1,j_2}$ (in red) are moved to rotational axis, is shown in Figure \ref{fig:cone}(c). By iterating over all possible combinations, one can obtain multiple estimates of the rotational axis; we average over all these estimates and return the resulting vector, denoted $\hat{\mathbf{a}}$. To put the point cloud $\mathcal{P}$ in its canonical form, we apply a roto-translation so that the vertex $\hat{\mathbf{v}}$ is moved to the origin of the coordinate axes  and $\hat{\mathbf{a}}$ coincides with the $z$-axis. The estimate $\hat{\alpha}$ is obtained by computing the angle between the $\mathbf{e}_3$ and the vector [$\frac{z_{max}}{r_{max}}\,, 0 \,, 1]$, where $z_{max}$ and $r_{max}$ are, respectively, the maximum value of the $z$-coordinates and the maximum distance from the origin of the projection on the $xy$-plane of $\mathcal{P}$. 

\begin{figure}[h!]
\centering
\begin{tikzpicture}

        \begin{scope}[xshift=0cm]
            \node[minimum width=2.55cm,minimum height=2.55cm,inner sep=0pt] (C1) {\includegraphics[width=2.2cm]{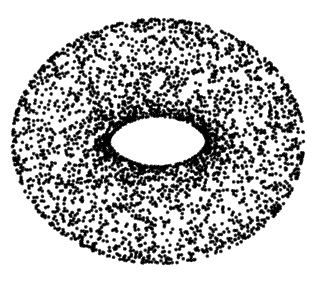}};
        \end{scope}
        
        \begin{scope}[xshift=4.0cm]
        \node[minimum width=2.55cm,minimum height=2.55cm,inner sep=0pt] (C2) {\includegraphics[width=2.7cm]{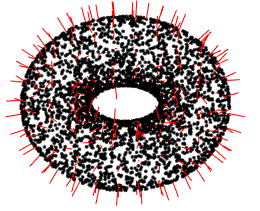}};
        \end{scope}
        
        \begin{scope}[xshift=10cm]
            \node[minimum width=2.55cm,minimum height=2.55cm,inner sep=0pt] (C3) {\includegraphics[width=4.5cm]{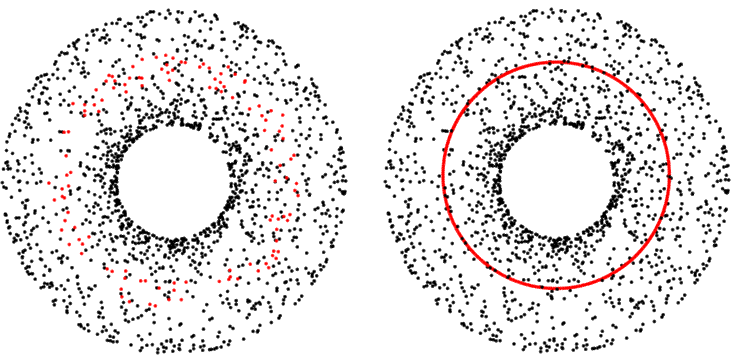}};
            \end{scope}
            
        \begin{scope}[xshift=0cm, yshift=-2.6cm]
            \node[minimum width=2.55cm,minimum height=2.55cm,inner sep=0pt] (C4) {\includegraphics[width=2.4cm]{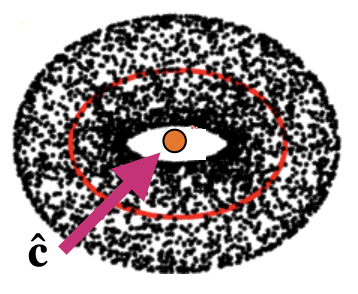}};
        \end{scope}
        
        \begin{scope}[xshift=4.0cm, yshift=-2.6cm]
        \node[minimum width=2.55cm,minimum height=2.55cm,inner sep=0pt] (C5) {\includegraphics[width=2.4cm]{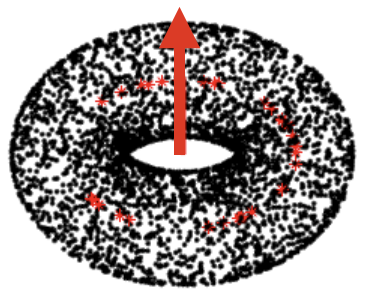}};
        \end{scope}
        
        \begin{scope}[xshift=10cm, yshift=-2.6cm]
            \node[minimum width=2.55cm,minimum height=2.55cm,inner sep=0pt] (C6) {\includegraphics[width=6.25cm]{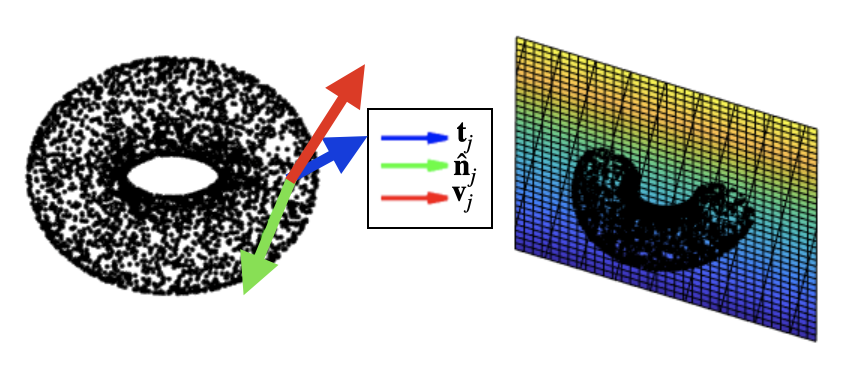}};
            \end{scope}

        \draw[very thick,->,>=stealth] ($(C1.east)+(0.2,0)$) -- ($(C2.west)+(-0.2,0)$) node [text width=2.5cm,midway,above,align=center,font=\tiny] {normal} node [text width=2.5cm,midway,below,align=center,font=\tiny] {estimation};
        \draw[very thick,->,>=stealth] ($(C2.east)+(0.2,0)$) -- ($(C3.west)+(-0.2,0)$) node [text width=2.5cm,midway,above,align=center,font=\tiny] {step} node [text width=2.5cm,midway,below,align=center,font=\tiny] {1};
        \draw[very thick,->,>=stealth] ($(C3.south)+(3,0.5)$) -- ($(C6.north)+(3,-0.5)$) node [text width=2.5cm,midway,above,align=center,font=\tiny, sloped] {step} node [text width=2.5cm,midway,below,align=center,font=\tiny, sloped] {2};
        \draw[very thick,->,>=stealth] ($(C6.west)+(-0.2,0)$) -- ($(C5.east)+(+0.2,0)$) node [text width=2.5cm,midway,above,align=center,font=\tiny] {step} node [text width=2.5cm,midway,below,align=center,font=\tiny] {3};
        \draw[very thick,->,>=stealth] ($(C5.west)+(-0.2,0)$) -- ($(C4.east)+(+0.2,0)$) node [text width=2.5cm,midway,above,align=center,font=\tiny] {step} node [text width=2.5cm,midway,below,align=center,font=\tiny] {4};
        
        \node[align = center, below] at ($(C1.south)+(0,0.3)$) {(a)};
        \node[align = center, below] at ($(C2.south)+(0,0.3)$) {(b)};
        \node[align = center, below] at ($(C3.south)+(0,0.3)$) {(c)};
        \node[align = center, below] at ($(C4.south)+(0,0.3)$) {(f)};
        \node[align = center, below] at ($(C5.south)+(0,0.3)$) {(e)};
        \node[align = center, below] at ($(C6.south)+(0,0.3)$) {(d)};
        
        \draw[style=dashed, rounded corners] (-1.45,-1.75) rectangle (5.40,1.5); 
        \node[draw,align=left, overlay] at (2,1.1) {Pre-processing};

    \end{tikzpicture}
    \caption{Initial estimates for a torus. The pre-processing step centers the point cloud and approximates the normal vectors at a set of points, see (b). In (c) the upper/lower circle recognition, while (d) shows the plane that identify small circles. Given the centers of small circles, in (e) the estimation of the axis $\mathbf{\hat{a}}$ and in (f) the estimation of center $\mathbf{\hat{c}}$ of the torus. }
    \label{fig:torus}
    \end{figure}

\paragraph*{Torus}
In line with the increase in the number of unknown parameters, this primitive requires a more complex handling, as summarized in the following four steps:
\begin{itemize}
    \item \emph{Upper (or lower) circle recognition.} We search for the best fitting plane to the entire point cloud $\mathcal{P}$ which -- unless pathological cases (e.g., very small segments) -- intersects the torus in (possibly perturbed arcs of) a circle, as shown in Figure \ref{fig:torus}(c, left). This circle can be recognized by the standard HT for circles; the parameters found can be used to generate a new dense set of points, which we will denote by $\mathcal{Q}$; an example is shown in Figure \ref{fig:torus}(c, right).
    \item \emph{Recognition of small circles.}
For each of the points $\mathbf{p}_j$, $j=1,\dots,k$, we find its nearest neighbour $\mathbf{q}_j\in\mathcal{Q}$; we then define the vector $\mathbf{v}_j$ as the cross product between the estimated normal vector $\hat{\mathbf{n}}_j$ at $\mathbf{p}_j$ and the vector $\mathbf{t}_j:=\mathbf{p}_j-\mathbf{q}_j$. An example is shown in Figure \ref{fig:torus}(d, left image): the green, blue and red vectors represent, respectively, $\mathbf{t}_j$, $\hat{\mathbf{n}}_j$ and $\mathbf{v}_j$. The just-computed vector $\mathbf{v}_j$ identifies a plane -- see Figure \ref{fig:torus}(d, right image) -- that intersects $\mathcal{P}$ in a set of points outlining two circles, up to some data perturbation. We apply the standard HT to recognise such circles and, more importantly, their radii and centers. For each recognised circle, we compute its Mean Fitting Error and store its center in $\mathcal{C}$ if the MFE is below some given threshold. By averaging the circle radii, we can get an estimate $\widehat{r}_{\min}$ of $r_{\min}$.
    \item \emph{Towards axis estimation.} We use the HT to find the best fitting plane to $\mathcal{C}$. The  normal vector to this plane is an estimate of the rotational axis of the torus, up to a translation (see Figure \ref{fig:torus}(e)). 
    \item \emph{Recognition of the big circle for center estimation.} Finally, we recognize the circle outlined by the points in $\mathcal{C}$, see Figure \ref{fig:torus}(f). The center of the torus is approximated by the circle center, which can be also used to fix the rotational axis. The radius of the circle gives us an estimate $\widehat{r}_{\max}$ of $r_{\max}$.
\end{itemize}

\subsection{Recognising primitives using the Hough transform}
\label{sec:algHT}
\begin{figure}[t!]
    \centering
    \includegraphics[width=15.0cm, trim={0 1cm 3cm 1cm}]{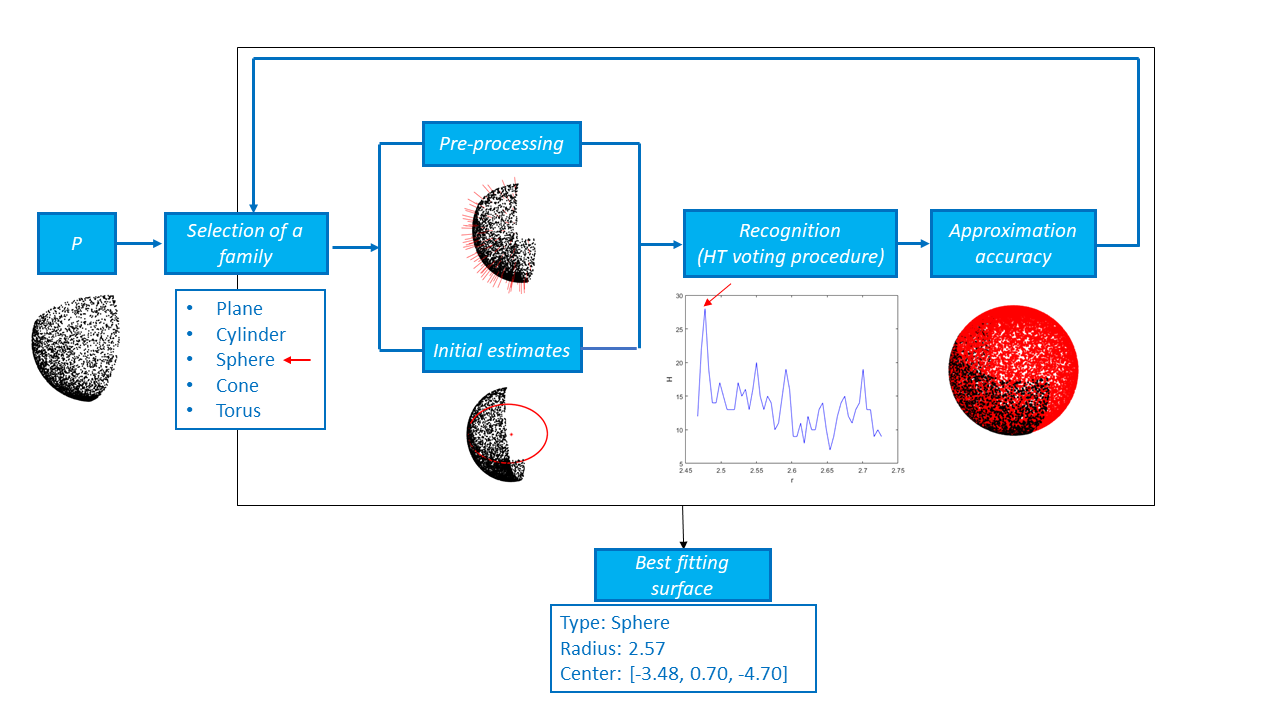}
    \caption{Pipeline of the method for fitting and recognising primitives using the Hough transform} 
    \label{fig:algPipeline}
\end{figure}

We can now exploit the pre-processing step (Section \ref{sec:preproc_tangpla}) and the parameter estimation procedures (Section \ref{sec:init_estim}) to apply subsequently the HT technique to each segment $\mathcal{P}_i$; this is particularly relevant when no prior information on the primitive type to look for is available. For each segment, our pipeline returns its type (i.e., plane, cylinder, cone, sphere, torus), and its geometric descriptors. Figure \ref{fig:algPipeline} illustrates the flow we follow to recognize primitives using the Hough transform.

Given a segment $\mathcal{P}_i$ and a specific family of surfaces $\mathcal{F}=\{\mathcal{S}_{\boldsymbol{\beta}} \}$ of the dictionary of surface primitives (e.g., family of tori), our method runs in three main steps: 

\begin{description}
    \item[Step 1: Pre-processing and initial estimates.] First, the segment $\mathcal{P}_i$ is pre-processed as described in Section  \ref{sec:preproc_tangpla}. Then, we find initial estimates for the current family of surfaces by applying the corresponding procedure from Section \ref{sec:init_estim}. The point cloud obtained from the pre-processing step, here denoted by $\mathcal{P}'_i$, is roto-translated  in order to put it in the standard position. Note that, by working on $\mathcal{P}'_i$ rather than on $\mathcal{P}_i$, we are able to deal with a dimensionally reduced parameter space. 
    \item[Step 2: HT-based surface recognition.] The new set of points $\mathcal{P}_i'$ is the input of the classical HT-based recognition algorithm introduced in Section \ref{sec:intro}. The output consists of the optimal parameters $\mathbf{\hat{\boldsymbol{\beta}}'}$, i.e., the parameters that best fit $\mathcal{P}_i'$ w.r.t. the given the family of surfaces (in canonical position). The initial estimates from the previous step give a hint to the HT technique about where the optimal solution is, thus eliminating the problem of the unboundedness of the parameter space.
    \item[Step 3: Evaluation of the approximation accuracy.] To measure the recognition accuracy of a specific primitive, we use the Mean Fitting Error $\text{MFE}(\mathcal{P}_i',\mathcal{S}_\mathbf{\hat{\boldsymbol{\beta}}'})$, as defined in Equation \ref{eqn:MFE}.
When the accumulator function exhibits more global maxima, the Hough transform returns more optimal solutions; in our case, we keep only the one having the lowest MFE.
\end{description}

When no prior information on the primitive type is available, the three steps above are repeated for each family of surfaces (in this work: planes, cylinders, spheres, cones and tori); the surface with the lowest MFE is returned as the best fitting surface $\mathcal{S}_\mathbf{\hat{\boldsymbol{\beta}'}}$ for the segment $\mathcal{P'}_i$. We admit that none of the primitive types at our disposal offers a satisfying fit of the segment if the all the computed MFEs are above a global threshold $\varepsilon$, here defined as the $5\%$ of the main diagonal of the model bounding box.

Finally, the roto-translation from Step 2 is applied backwards in order to obtain the parametric representation and the geometric descriptors for the segment $\mathcal{P}_i$ in its original position.

\subsubsection{Computational complexity}
The method includes a segment pre-processing and the actual primitive fitting. The segment pre-processing includes the estimation of distances, the axis-aligned bounding box and an approximation of the tangent plane via the HT. For spheres, cones, cylinders and tori, the parameter estimation includes the recognition of circle(s), again by using the HT. The segment recognition procedure basically consists of the HT procedure for the surface primitives in standard form. In practice, both the pre-processing and the fitting procedure mainly depend on the HT, being the other operations in $\OOO(S)$ or in $\OOO(S \log S)$, where $S$ represents the number of points of the segment. The computational complexity of the HT voting procedure is dominated by the size of the accumulator function: denoting $M$ the number of cells of the space of parameters, the computational complexity of the HT recognition on a segment is $\OOO(MS)$. The number of parameters for the HT directly influences the size of $M$. In our method, their number is 3 for planes and circles (they are used in the pre-processing), 2 for tori, 1 for spheres, cylinders and cones. 

Through this parameter reduction, we are able to deal with parametric primitives that would otherwise have 4 parameters (sphere) or at least 9 parameters (cylinder, cone and torus), as deducible from the representations in Section \ref{sec:standardForm}. This reduction in parameters makes it possible to apply HT to primitives that would otherwise not be computable in practice due to the explosion of the spatial complexity of the accumulator function.

The pre-processing and the HT recognition steps are repeated for each segment and for each geometric primitive. It is worth noting that, being each segment and each primitive fitting performed independently, the task is  \emph{embarrassingly parallel}.

\subsection{Examples on synthetic data}
\label{sec:part-noise}
\begin{figure}[h]
    \centering
    
    \begin{minipage}{.3\linewidth}
      \centering
        \resizebox{0.95\columnwidth}{!}{
    \begin{tabular}{|c|c|}
    \hline
    \rowcolor{blue!30} \multicolumn{2}{|c|}{Noise-free segments}
    \\
    \hline
    \includegraphics[scale=0.6, trim={1.5cm 1.00cm 1.5cm 1.00cm},clip]{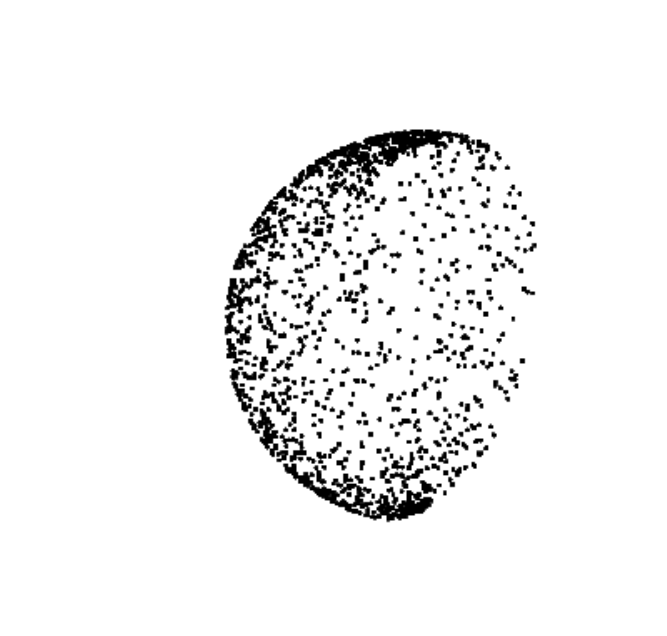}
    & 
    \includegraphics[scale=0.6, trim={1.5cm 1.00cm 1.5cm 1.25cm},clip]{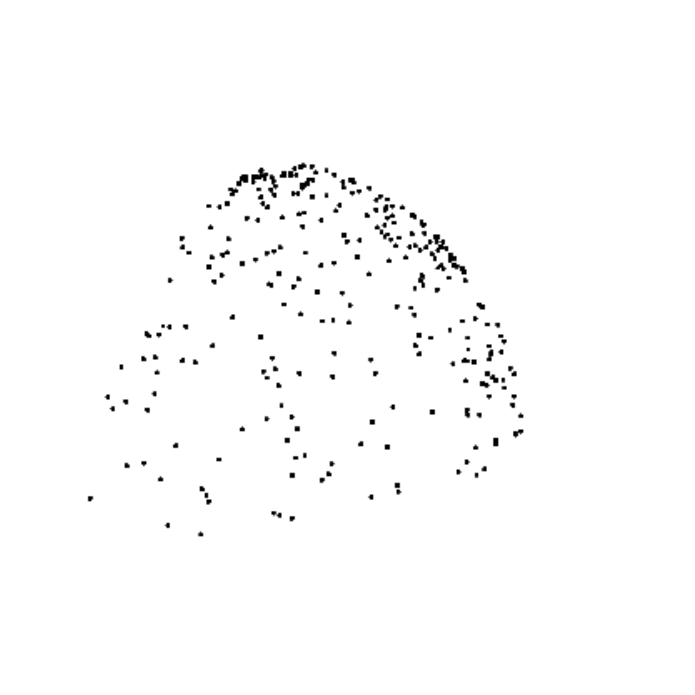}
    \\
    {\small$|r-\hat{r}|=0.00$}
    &
    {\small$|r-\hat{r}|=0.01$}
    \\
    {\small$||\mathbf{c}-\hat{\mathbf{c}}||_2=0.00$}
    &
    {\small$||\mathbf{c}-\hat{\mathbf{c}}||_2=0.02$}
    \\
    \hline
    \includegraphics[scale=0.6, trim={1.5cm 0.75cm 1.5cm 0.50cm},clip]{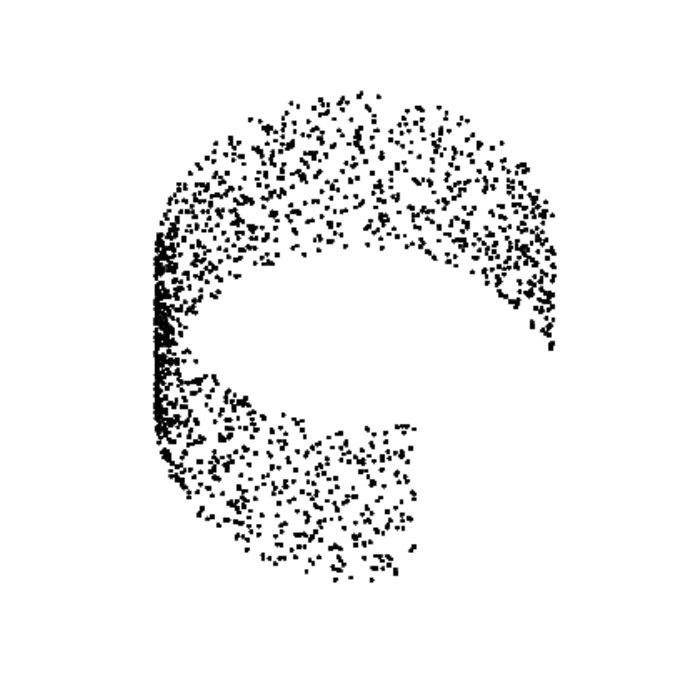}
    & 
    \includegraphics[scale=0.6, trim={1.5cm 0.75cm 1.5cm 0.50cm},clip]{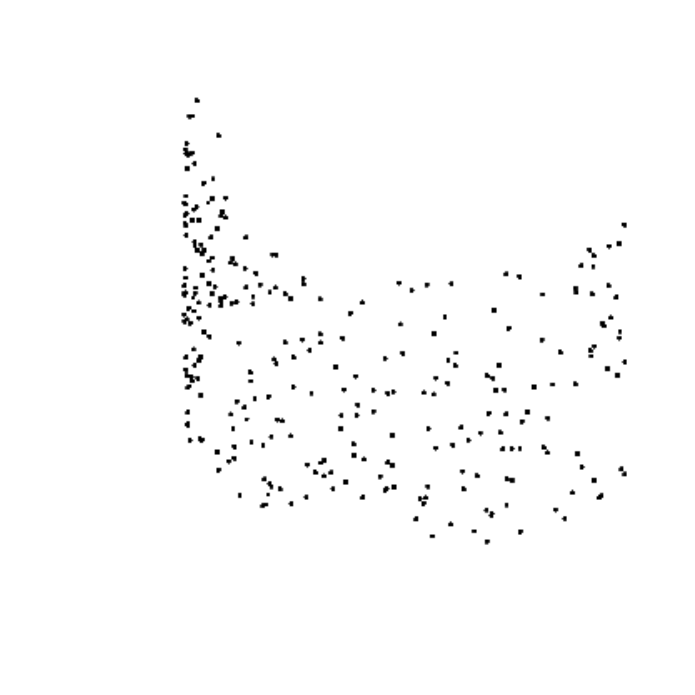}
    \\
    {\small$|r-\hat{r}|=0.00$}
    &
    {\small$|r-\hat{r}|=0.00$}
    \\
    {\small$||\mathbf{a}-\hat{\mathbf{a}}||_2=0.00$}
    &
    {\small$||\mathbf{a}-\hat{\mathbf{a}}||_2=0.00$}
    \\
    \hline
    \includegraphics[scale=0.6, trim={1.0cm 1.0cm 1.0cm 1.0cm},clip]{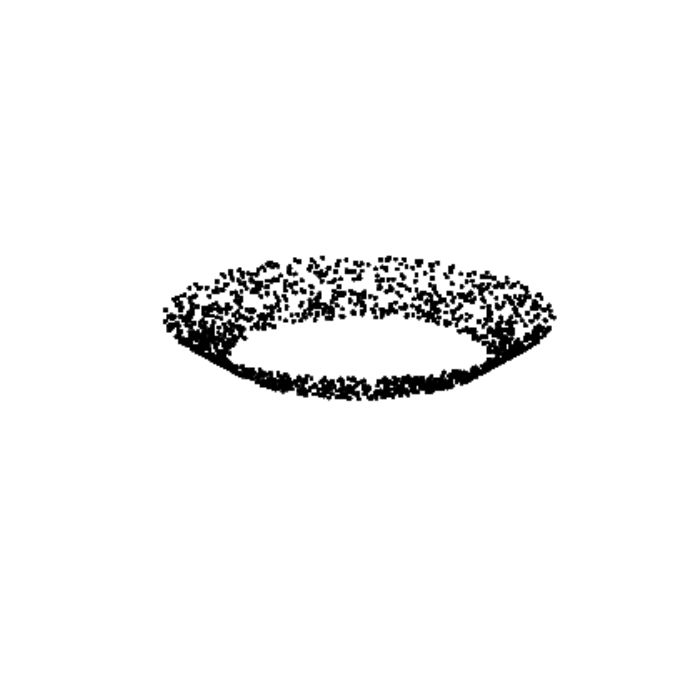}
    &
    \includegraphics[scale=0.6, trim={1.0cm 1.0cm 1.0cm 1.0cm},clip]{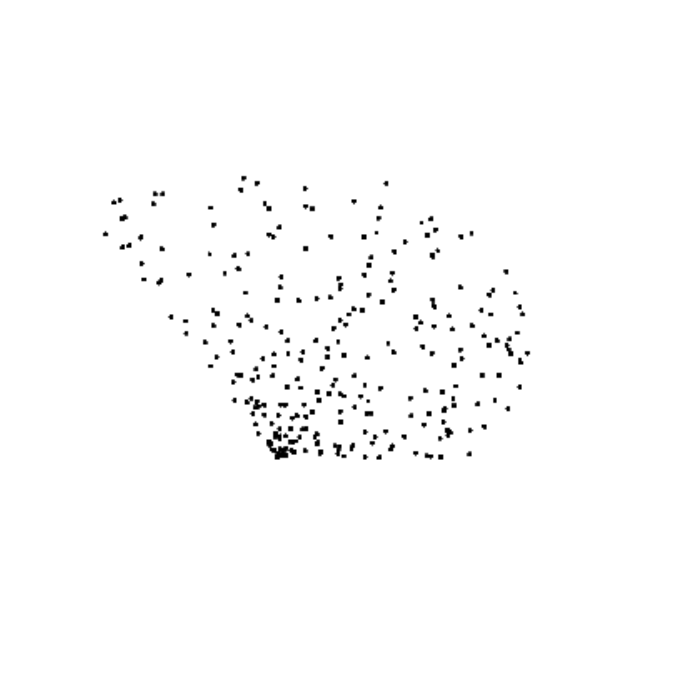}
    \\
    {\small$|\alpha-\hat{\alpha}|=0.00$}
    &
    {\small$|\alpha-\hat{\alpha}|=0.01$}
    \\
    {\small$||\mathbf{a}-\hat{\mathbf{a}}||_2=0.00$}
    &
    {\small$||\mathbf{a}-\hat{\mathbf{a}}||_2=0.00$}
    \\
    {\small$||\mathbf{v}-\hat{\mathbf{v}}||_2=0.01$}
    &
    {\small$||\mathbf{v}-\hat{\mathbf{v}}||_2=0.02$}
    \\
    \hline
    \includegraphics[scale=0.6, trim={1.0cm 1.45cm 1.0cm 1.15cm},clip]{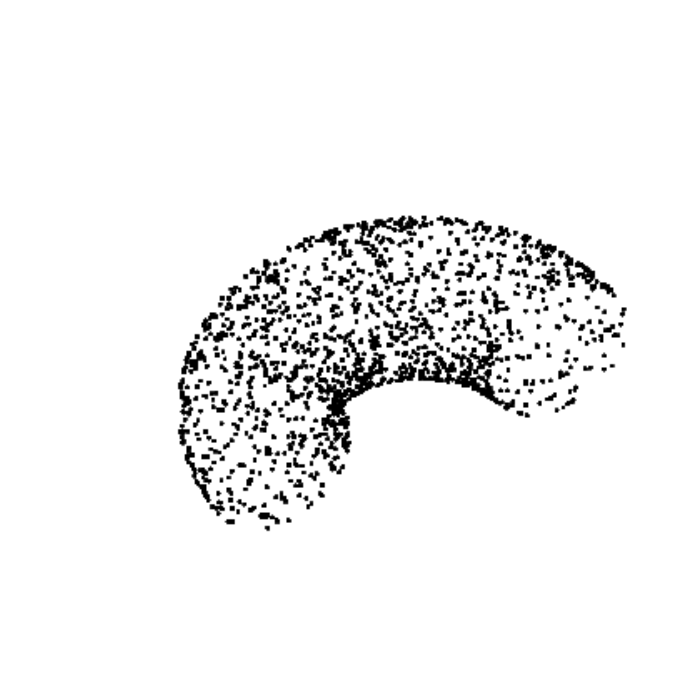}
    & 
    \includegraphics[scale=0.6, trim={1.0cm 1.45cm 1.0cm 1.15cm},clip]{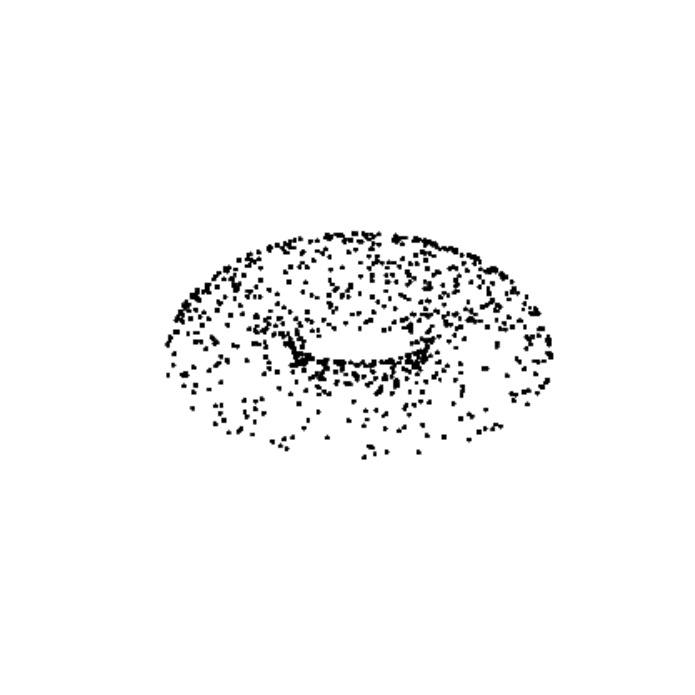}
    \\
    {\small$|r_{\min}-\hat{r}_{\min}|=0.00$}
    & {\small$|r_{\min}-\hat{r}_{\min}|=0.01$}
    \\
    {\small$|r_{\max}-\hat{r}_{\max}|=0.01$}
    & {\small$|r_{\max}-\hat{r}_{\max}|=0.02$}
    \\
    {\small$||\mathbf{c}-\hat{\mathbf{c}}||_2=0.01$}
    & {\small$||\mathbf{c}-\hat{\mathbf{c}}||_2=0.02$}
    \\
    {\small$||\mathbf{a}-\hat{\mathbf{a}}||_2=0.00$}
    & {\small$||\mathbf{a}-\hat{\mathbf{a}}||_2=0.00$}
    \\
    \hline
    \end{tabular}
    }
    \end{minipage}%
    \begin{minipage}{.68\linewidth}
    \resizebox{0.90\columnwidth}{!}{
    \begin{tabular}{|c|c|c|c|}
    \hline
    \rowcolor{blue!30} \multicolumn{4}{|c|}{Segments with increasing noise} \\
    \hline
    \rowcolor{blue!10}
    Noise-free
    &
    $\mathcal{U}(-0.1,0.1)$
    &
    $\mathcal{U}(-0.5,0.5)$
    &
    $\mathcal{U}(-1.0,1.0)$
    \\
    \hline
    \includegraphics[scale=0.6, trim={1.5cm 1.0cm 1.5cm 1.25cm},clip,angle =-90]{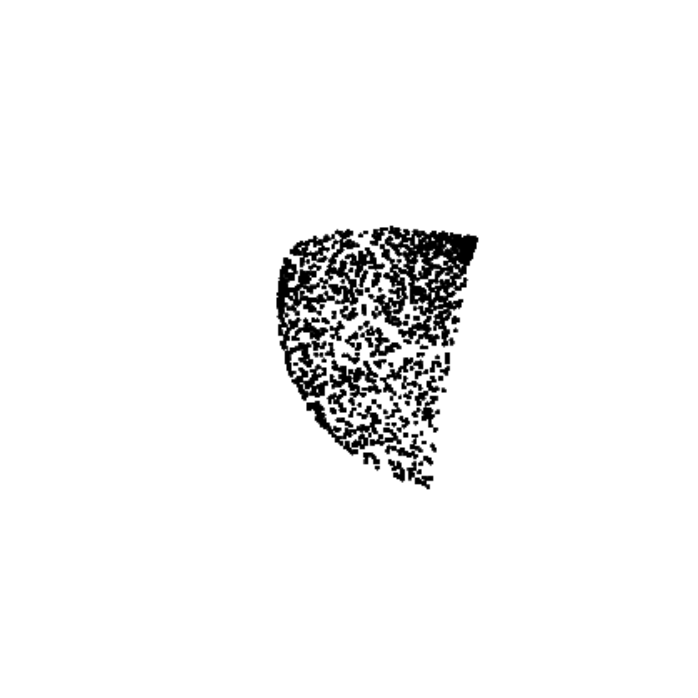}
    & 
    \includegraphics[scale=0.6, trim={1.5cm 1.0cm 1.5cm 1.25cm},clip,angle =-90]{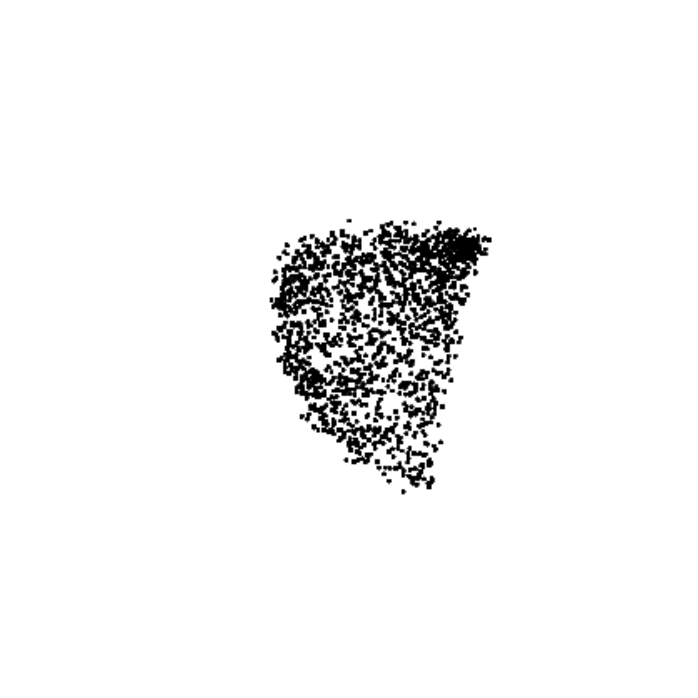}
    &
    \includegraphics[scale=0.6, trim={1.5cm 1.0cm 1.5cm 1.25cm},clip,angle =-90]{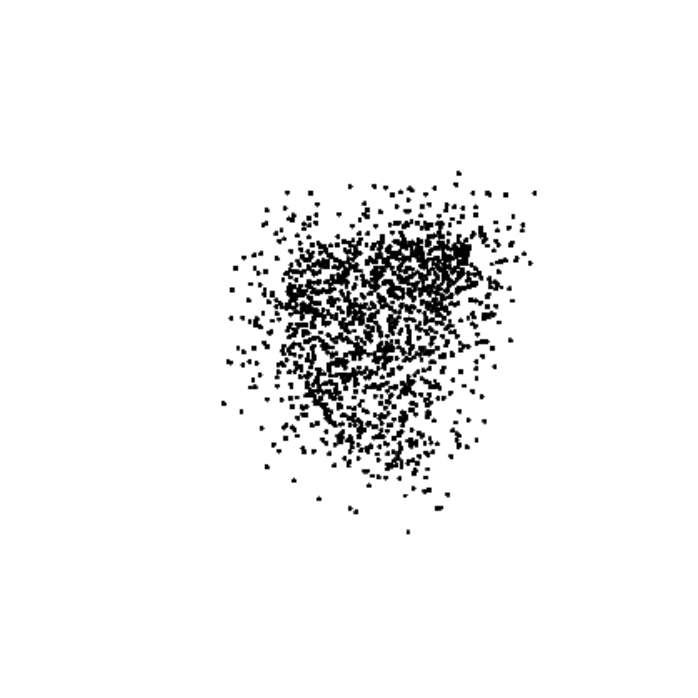}
    &
    \includegraphics[scale=0.6, trim={1.5cm 1.0cm 1.5cm 1.25cm},clip,angle =-90]{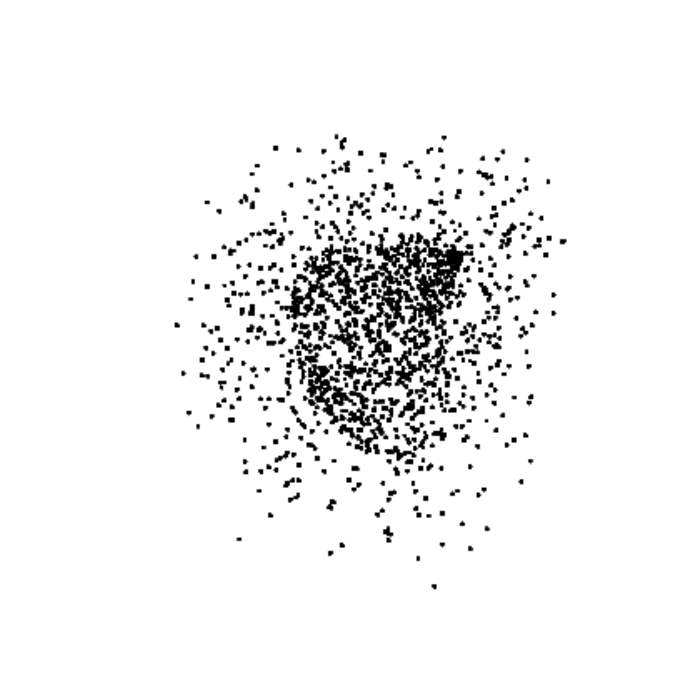}
    \\
    {\small$|r-\hat{r}|=0.00$}
    &
    {\small$|r-\hat{r}|=0.02$}
    &
    {\small$|r-\hat{r}|=0.05$}
    &
    {\small$|r-\hat{r}|=0.03$}
    \\
    {\small$||\mathbf{c}-\hat{\mathbf{c}}||_2=0.01$}
    &
    {\small$||\mathbf{c}-\hat{\mathbf{c}}||_2=0.02$}
    &
    {\small$||\mathbf{c}-\hat{\mathbf{c}}||_2=0.02$}
    &
    {\small$||\mathbf{c}-\hat{\mathbf{c}}||_2=0.06$}
    \\
    \hline
    \includegraphics[scale=0.6, trim={1.5cm 1.45cm 1.5cm 1.25cm},clip]{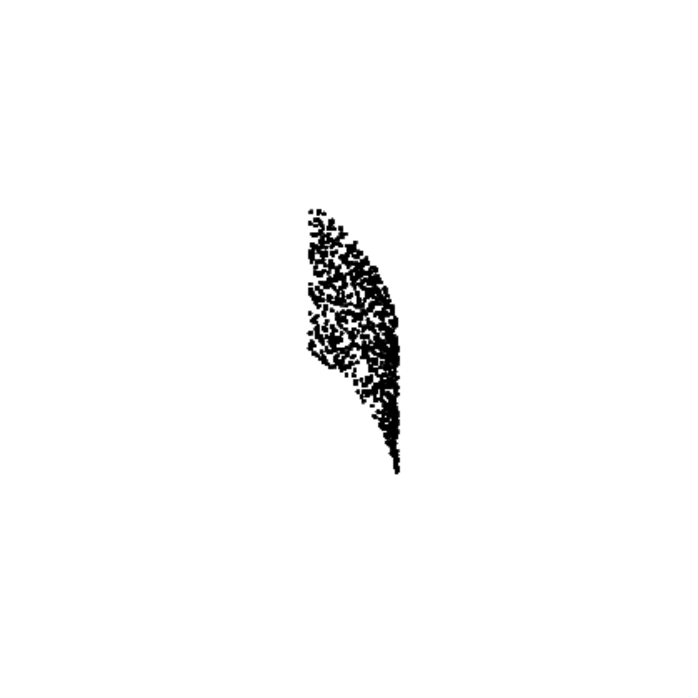}
    & 
    \includegraphics[scale=0.6, trim={1.5cm 1.45cm 1.5cm 1.25cm},clip]{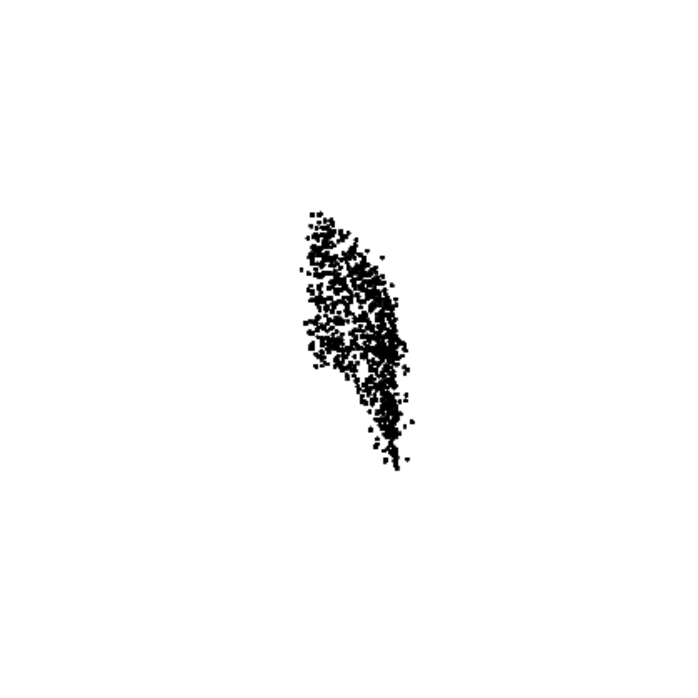}
    &
    \includegraphics[scale=0.6, trim={1.5cm 1.45cm 1.5cm 1.25cm},clip]{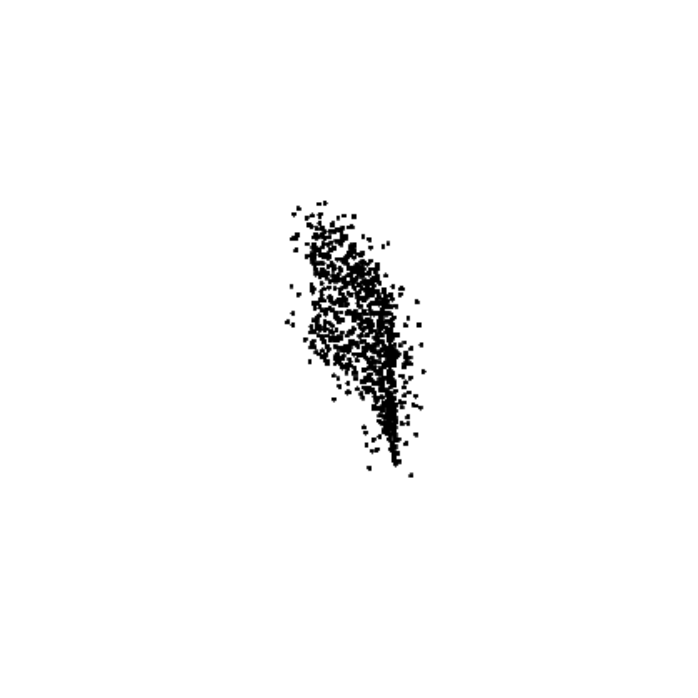}
    &
    \includegraphics[scale=0.6, trim={1.5cm 1.45cm 1.5cm 1.25cm},clip]{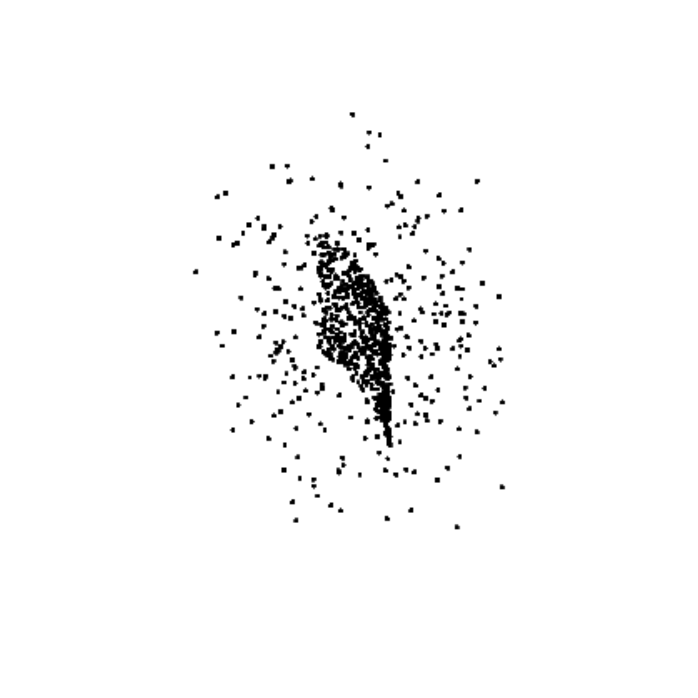}
    \\
    {\small$|r-\hat{r}|=0.00$}
    &
    {\small$|r-\hat{r}|=0.00$}
    &
    {\small$|r-\hat{r}|=0.04$}
    &
    {\small$|r-\hat{r}|=0.47$}
    \\
    {\small$||\mathbf{a}-\hat{\mathbf{a}}||_2=0.00$}
    &
    {\small$||\mathbf{a}-\hat{\mathbf{a}}||_2=0.00$}
    &
    {\small$||\mathbf{a}-\hat{\mathbf{a}}||_2=0.00$}
    &
    {\small$||\mathbf{a}-\hat{\mathbf{a}}||_2=0.44$}
    \\
    \hline
    \includegraphics[scale=0.4, trim={1.0cm 1.0cm 1.0cm 1.0cm},clip]{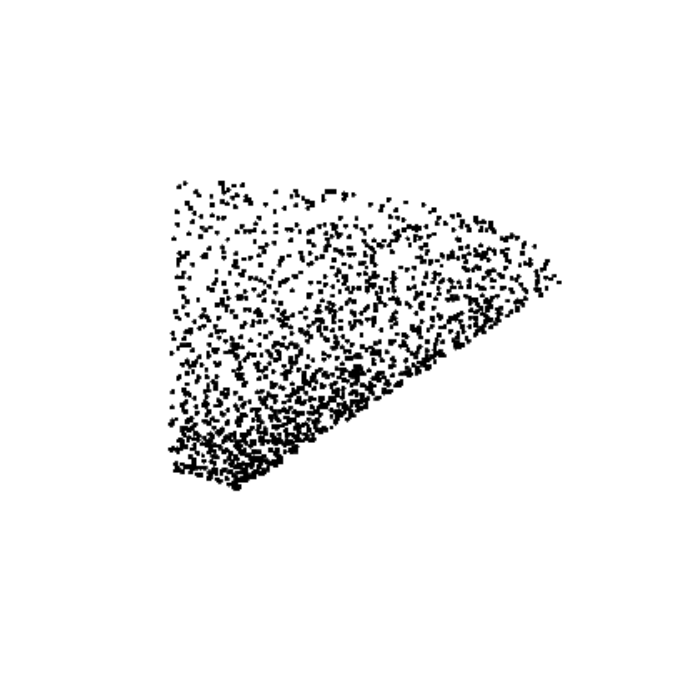}
    & 
    \includegraphics[scale=0.4, trim={1.0cm 1.0cm 1.0cm 1.0cm},clip]{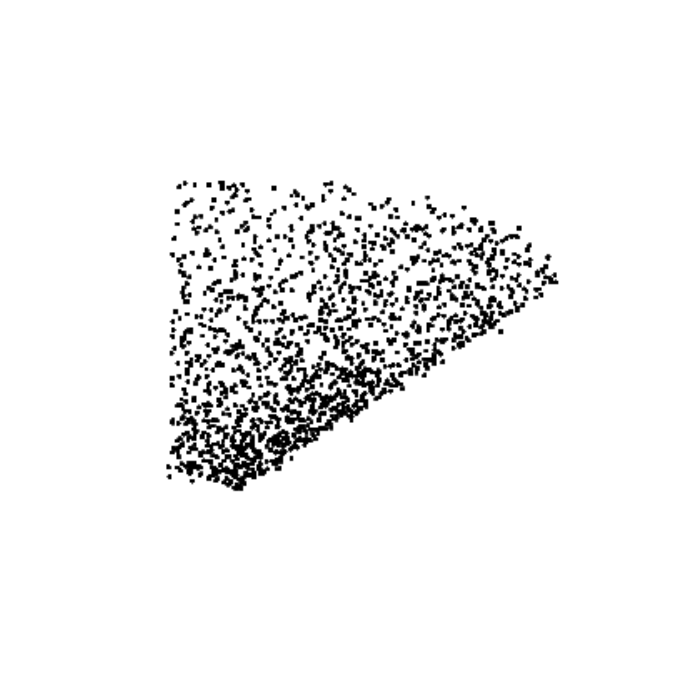}
    &
    \includegraphics[scale=0.4, trim={1.0cm 1.0cm 1.0cm 1.0cm},clip]{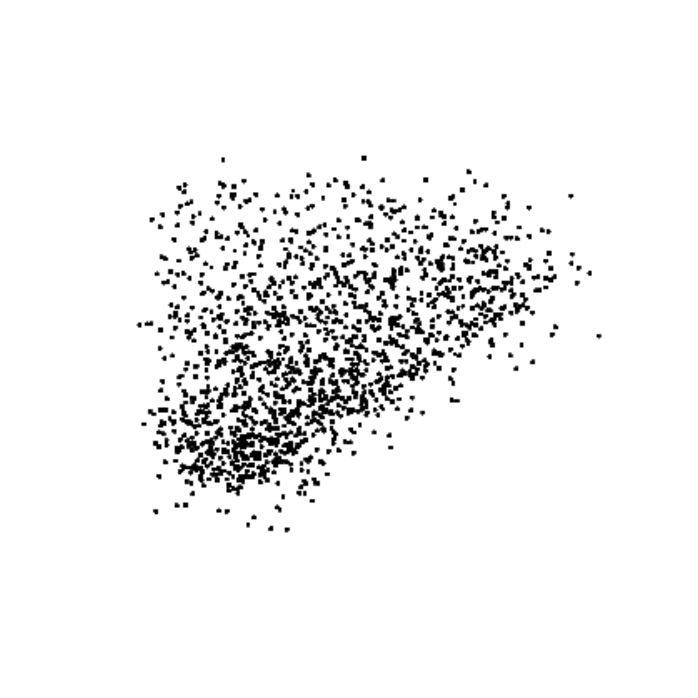}
    &
    \includegraphics[scale=0.4, trim={1.0cm 1.0cm 1.0cm 1.0cm},clip]{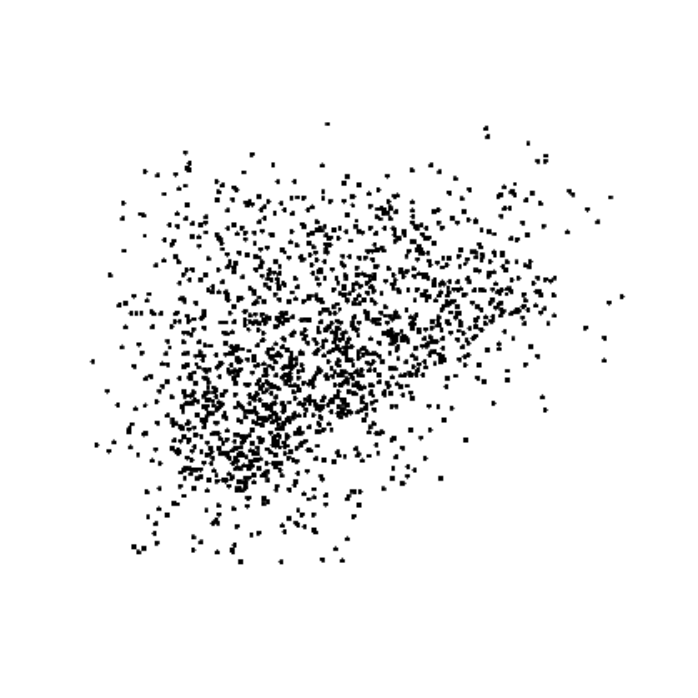}
    \\
    {\small$|\alpha-\hat{\alpha}|=0.00$}
    &
    {\small$|\alpha-\hat{\alpha}|=0.00$}
    &
    {\small$|\alpha-\hat{r}|=0.01$}
    &
    {\small$|\alpha-\hat{\alpha}|=0.03$}
    \\
    {\small$||\mathbf{a}-\hat{\mathbf{a}}||_2=0.00$}
    &
    {\small$||\mathbf{a}-\hat{\mathbf{a}}||_2=0.00$}
    &
    {\small$||\mathbf{a}-\hat{\mathbf{a}}||_2=0.19$}
    &
    {\small$||\mathbf{a}-\hat{\mathbf{a}}||_2=0.37$}
    \\
    {\small$||\mathbf{v}-\hat{\mathbf{v}}||_2=0.02$}
    &
    {\small$||\mathbf{v}-\hat{\mathbf{v}}||_2=0.02$}
    &
    {\small$||\mathbf{v}-\hat{\mathbf{v}}||_2=0.03$}
    &
    {\small$||\mathbf{v}-\hat{\mathbf{v}}||_2=0.03$}
    \\
    \hline
    \includegraphics[scale=0.5, trim={1.0cm 1.45cm 1.0cm 1.15cm},clip]{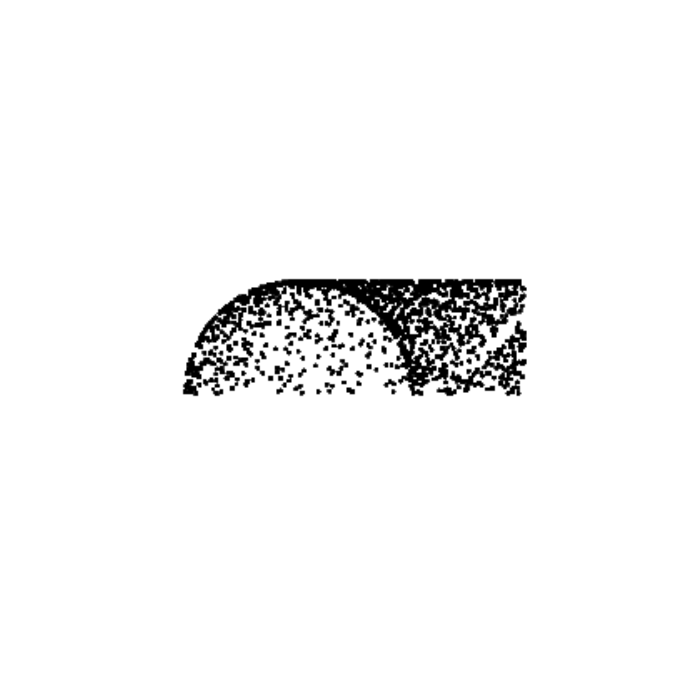}
    & 
    \includegraphics[scale=0.5, trim={1.0cm 1.45cm 1.0cm 1.15cm},clip]{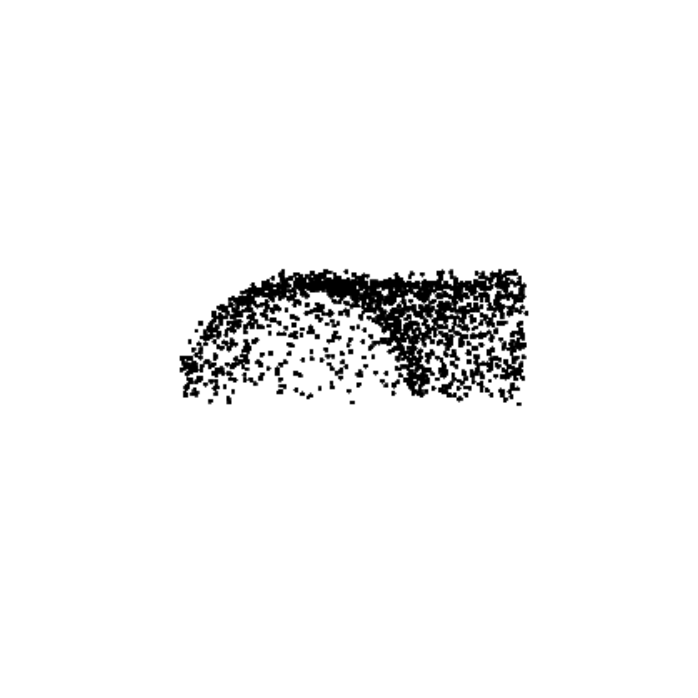}
    &
    \includegraphics[scale=0.5, trim={1.0cm 1.45cm 1.0cm 1.15cm},clip]{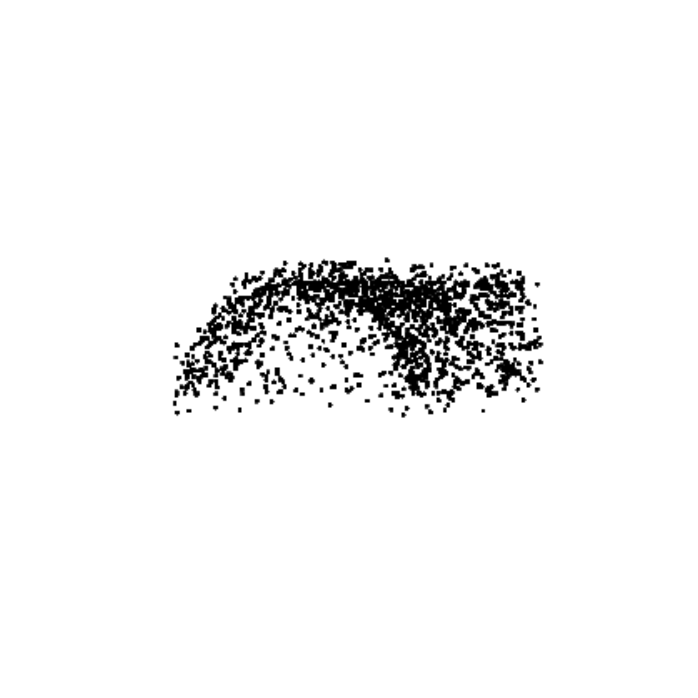}
    &
    \includegraphics[scale=0.5, trim={1.0cm 1.45cm 1.0cm 1.15cm},clip]{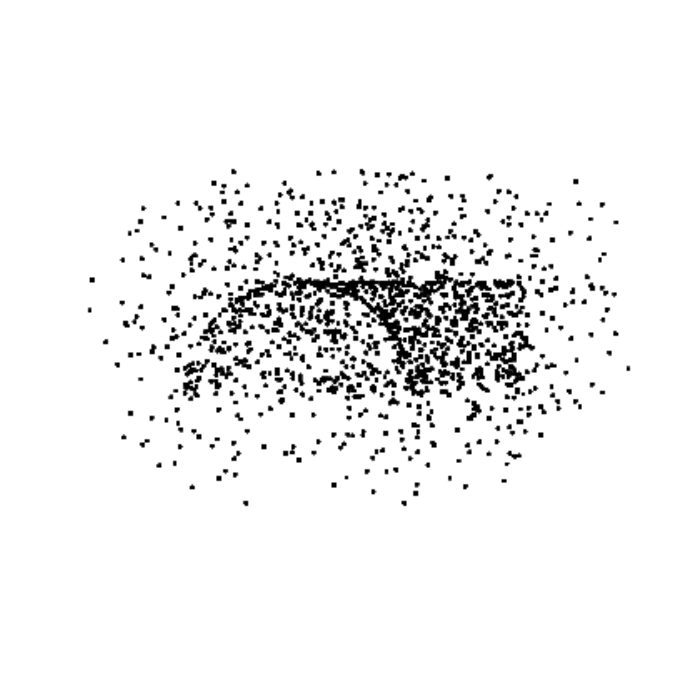}
    \\
    {\small$|r_{\min}-\hat{r}_{\min}|=0.00$}
    & {\small$|r_{\min}-\hat{r}_{\min}|=0.01$}
    & {\small$|r_{\min}-\hat{r}_{\min}|=0.02$}
    & {\small$|r_{\min}-\hat{r}_{\min}|=0.09$}
    \\
    {\small$|r_{\max}-\hat{r}_{\max}|=0.02$}
    & {\small$|r_{\max}-\hat{r}_{\max}|=0.02$}
    & {\small$|r_{\max}-\hat{r}_{\max}|=0.04$}
    & {\small$|r_{\max}-\hat{r}_{\max}|=1.18$}
    \\
    {\small$||\mathbf{c}-\hat{\mathbf{c}}||_2=0.02$}
    & {\small$||\mathbf{c}-\hat{\mathbf{c}}||_2=0.03$}
    & {\small$||\mathbf{c}-\hat{\mathbf{c}}||_2=0.04$}
    & {\small$||\mathbf{c}-\hat{\mathbf{c}}||_2=0.07$}
    \\
    {\small$||\mathbf{a}-\hat{\mathbf{a}}||_2=0.00$}
    & {\small$||\mathbf{a}-\hat{\mathbf{a}}||_2=0.01$}
    & {\small$||\mathbf{a}-\hat{\mathbf{a}}||_2=0.02$}
    & {\small$||\mathbf{a}-\hat{\mathbf{a}}||_2=0.07$}
    \\
    \hline
    \end{tabular}
    }
    \end{minipage}
\caption{Initial estimates for various  segments, without noise (left) and with increasing noise (right). Moving row by row, the segments represent different pieces of: a sphere and a cylinder of radius $r=1.5$, a cone of angle $\alpha=56.24^\circ$ and a torus of radii $r_{min}=1$ and $r_{max}=2$. Further details on the ground truth parameters are given in Section \ref{sec:part-noise}.
\label{fig:syntetic}}
\end{figure}

As a sanity check, we have tested our strategy by computing initial estimates for the point clouds given in Figure \ref{fig:syntetic}. A validation of the whole pipeline, performed by using it to solve a problem in CAD reverse engineering, is left to Section \ref{sec:HTclustering}. The noise-free segments shown in Figure \ref{fig:syntetic} (left) are all successfully handled: by comparing true and estimated parameters, we can conclude that floating-point arithmetic does not affect much their (initial) parameter estimate; given the fairly low error, these cases could avoid undergoing Step 2 from Section \ref{sec:algHT}. Moreover, the results suggest that the parameter estimation procedure can successfully handle undersampled segments, when the data are sufficiently clean.

Figure \ref{fig:syntetic} (right) illustrates the stability of our procedure for parameter estimation against an increasing amount of noise.  As we move row by row from top to bottom, we show: a one-eighth portion of a full sphere of radius $r=1.5$ and center $\mathbf{c}=(0,0,0)$; a small portion of a cylinder of radius $r=1.5$ and aligned to the $z-$axis; a one-eighth portion of a cone of angle $\alpha=56.24^{\circ}$, vertex $\mathbf{v}=(0,0,0)$ and rotational axis coincident with the $z-$axis; and a one-eighth portion of a full torus aligned with the $z$-axis, with $r_{\min}=1$, $r_{\max}=2$, $\mathbf{c}=(0,0,0)$. As we move from left to right, the progressively increasing in noise corresponds to a lower precision of the initial estimates: the cylindrical segment in the second row suggests that the initial estimation can fail when applied to particularly small segments suffering from strong noise levels.

To quantify the performance of our current implementation we have generated, for each primitive type, $100$ point clouds; the size of these point clouds varies from a minimum of 400 points to a maximum of 1,600 points. The average CPU time, for each primitive type, is: 24.8 seconds for cylinders, 17.0 seconds for spheres, 28.9 seconds for cones and 49.4 seconds for tori. These tests were performed on a desktop equipped with an Intel Core i9 processor (at 3.6 GHz), in a Windows system 64 bits; the method is implemented in MATLAB and currently does not take advantage of the parallelizability of the voting procedure.

\section{Recognition of global relations among geometric primitives extracted from CAD objects \label{sec:HTclustering}}
A common problem in CAD reverse engineering is that of deducing global relations among primitives derived by segmentation algorithms.

A paper that first explored how different geometric primitives can satisfy co-planarity and co-axiality relationships is GlobFit \cite{Li2011}. Starting from an initial mesh segmentation into four basic primitives (planes, cylinders, cones and spheres) obtained with RANSAC \cite{Schnabel2007}, GlobFit builds an orientation graph that is used to infer possible relations between pair of primitives, and eventually optimize their alignment. \cite{Monszpart2015} tackles the problem of selecting local plane-based approximations along with their global inter-plane relations, with the purpose of reconstructing buildings from raw scans with missing data, noise, and varying sampling density. Similarly, \cite{Oesau2015} considers the three main relationships that form the major regularities in urban scenes -- parallelism, orthogonality and co-planarity --  to reinforce detection by non-local fitting.  Even if limited to planes, both \cite{Monszpart2015} and \cite{Oesau2015} aim at detecting overall relations among the primitives able to overcome the connectivity-oriented relations presented in \cite{Li2011}. More recently,  \cite{Raffo:2020} introduced  a novel approach to determine primitive shapes by combining clustering analysis with approximate implicitization: given a set of non-overlapping patches, the method  partitions the patches into subsets corresponding to the underlying primitive shape they originate from; despite returning exact results in exact arithmetic, the approach shows limitations -- both theoretically and experimentally -- when it deals with models with particularly small parts, or when noise is present.

In the following, Section \ref{sec:method_clust} provides an overview of the clustering method that exploits the geometric descriptors obtained in the recognition phase to aggregate segments belonging to the same primitive or sharing certain geometric characteristics. Sections \ref{sec:ABC} and \ref{sec:Fit4CAD} evaluate the performance of the proposed method over the ABC dataset,  \cite{Koch:2019} and the Fit4CAD benchmark \cite{Fit4CAD}, respectively. A comparison with a method that adopts a similar pipeline is shown in Section \ref{sec:comparison_Raffo}. Finally, Section \ref{sec:input_methods} provides the performance of our method considering segmented point clouds created by two different methods.

\subsection{Overview of the clustering method}\label{sec:method_clust}

Given a segmented point cloud representing a CAD object, we here propose a method that exploits the parameters obtained in the recognition process described in Section \ref{sec:algHT}. These parameters uniquely characterize the segments and can be used as geometric descriptors allowing  them to be aggregated  using a clustering approach. The pipeline of such a method is built over two consecutive steps.

    \paragraph*{Step 1} Firstly, we apply the algorithm described in Section \ref{sec:algHT} to each input segment. This allows us to label each segment with its most likely primitive type, as well as obtaining its parametric representation and its shape description.
    \paragraph*{Step 2} Once all segments have been processed, we apply a well-known (hierarchical) clustering approach -- the \emph{complete-linkage} -- to compare clusters and build a dendrogram. The use of complete-linkage is here justified by the need of penalizing chaining effects. The method starts with singletons as clusters, and proceeds by merging, step by step, those clusters that are the closest with respect to the map
    \begin{equation*}
        D(C_i,C_j):=\max_{\boldsymbol{\tau}_k\in C_i,\boldsymbol{\tau}_l\in C_j}d(\boldsymbol{\tau}_k,\boldsymbol{\tau}_l),
    \end{equation*}
    where $C_i, C_j$ is a given pair of clusters (of segments) and $d$ is a user-defined distance or dissimilarity. For any pair of segments $\tau_1$, $\tau_2$ belonging to the same family, several distances $d(\tau_1, \tau_2)$ are possible. Table \ref{table:basic_shapes} lists, for each type of primitive, some simple distances and the intuitive concepts they are meant to measure, using the notation introduced in Section \ref{sec:standardForm}. 
    
In practice, we have at our disposal a set of distances, each one corresponding to a condition we are interested to measure. Note that all distances listed in Table \ref{table:basic_shapes} are metrics and $d(\tau_1, \tau_2)=0$ implies that the primitives $\tau_1$ and $\tau_2$ are equal with respect to that criterion.

\begin{table}[h!]
  \centering
  \small
  \caption{Basic distances for the primitives considered in the paper, using the notation from Figure \ref{fig:basic_primitives}. The subscripts $1$ and $2$ in the parameters refer to the two segments $\tau_1$ and $\tau_2$. \label{table:basic_shapes}}
  \begin{tabular}{|l c c r|}
  \hline
    \rowcolor{blue!30}Primitives & &  Distance $d(\tau_1, \tau_2)$ & Query \\\hline
    & & & \\
    \multirow{2}{*}{Planes} & \ldelim\{{2}{3mm}[] & $||\mathbf{a}_1\times{}\mathbf{a}_2||_2$ & parallel \\
    &  & $|\mathbf{a}_1\cdot{}(\mathbf{p}_1-\mathbf{p}_2)|$ & incident \\[0.2cm]
    \multirow{3}{*}{Cylinders} & \ldelim\{{3}{3mm}[]       & $|r_1-r_2|$ & equal radii \\
    &                           & $||\mathbf{a}_1\times{}\mathbf{a}_2||_2$ & parallel rotational axes \\
    &                           & $||\mathbf{a}_1\times{}(\mathbf{p}_1-\mathbf{p}_2)||_2$ & incident rotational axes \\[0.2cm]
    \multirow{3}{*}{Cones} & \ldelim\{{3}{3mm}[]       & $|\alpha_1-\alpha_2|$ & equal apertures \\
    &                           & $||\mathbf{a}_1\times{}\mathbf{a}_2||_2$ & parallel rotational axes \\
    &                           & $||\mathbf{v}_1-\mathbf{v}_2||_2$ & equal vertices \\[0.2cm]
    \multirow{2}{*}{Spheres} & \ldelim\{{2}{3mm}[]       & $|r_1-r_2|$ & equal radius \\
    &                           & $||\mathbf{c}_1-\mathbf{c}_2||_2$ & equal centers \\[0.2cm]
    \multirow{4}{*}{Tori} & \ldelim\{{4}{3mm}[]       & $|r_{\min,1}-r_{\min,2}|$ & equal smallest radii \\
    &                           & $|r_{\max,1}-r_{\max,2}|$ & equal largest radii \\
    &                           & $||\mathbf{a}_1\times{}\mathbf{a}_2||_2$ & parallel rotational axes\\
    &                           & $||\mathbf{c}_1-\mathbf{c}_2||_2$ & equal centers \\[0.2cm]\hline
  \end{tabular}
\end{table}

In addition, more complex queries can be formulated by summing simple distances. For instance, one can check whether two segments lie on the same torus by using the metric $d(\tau_1, \tau_2):=|r_{\min,1}-r_{\min,2}|+|r_{\max,1}-r_{\max,2}|+||\mathbf{a}_1\times{}\mathbf{a}_2||_2+||\mathbf{c}_1-\mathbf{c}_2||_2$.

Note that the sum of distance is yet a distance, and that cutting the dendrogram with increasing thresholds corresponds to weakening the conditions imposed as the query.  

\subsection{Examples of applications}
We are now ready to evaluate the capability of our method to fit and aggregate primitives according to different correlation queries. For all models, we show how primitives are aggregated if they belong to the same geometric primitive or according to  different relations, such as co-planarity, co-axiality, parallelism. For reasons of space, despite a much wider experimentation, we show only some figures with the most significant relationships found. For all examples in this section, the following thresholds for cutting the dendrograms have been selected: $10^{-10}$ for planes; $10^{-1}$ spheres, cylinders, coni and tori.  

\subsubsection{Models from the ABC dataset  (\cite{Koch:2019})}\label{sec:ABC}

To better evaluate the behavior of our method we used datasets available online.
The first one we considered was the ABC dataset, which contains a collection of one million CAD models created for researching geometric deep learning methods and applications.

Figure \ref{fig:tubo} presents a point cloud  of $22,803$ points and containing $18$ segments: planes, cylinders, cones and tori. Segments satisfying the same correlation query are represented by the same colour. As shown in the images, our method can successfully recognize the primitive type and use the segment parameters to infer various relations. The expressions ``same plane"/``same cylinder"/``same cone"/``same sphere"/``same torus" are henceforth used to test the (possible) presence of segments originating from the same underlying primitive (e.g., to group together toric segments having the same center, radii and rotational axis). The mean MFE over all segments is $0.0019$.

The point set in Figure \ref{fig:topolino_pokeball}(a) is composed of $13,455$ points and contains 26 segments, of which: 2 are extracted from planes, 8 from cylinders, and 16 from tori. This model is perfectly handled by our method, without misclassification in any correlation query. The mean MFE over all segments is $0.0036$. The model in Figure \ref{fig:topolino_pokeball}(b) counts $9$ segments for a total of $15,726$ points. In this example, a pair of segments obtained from the same sphere is present. Again, the grouping proceeds smoothly, except for two queries where two cylinders with very similar radii are clustered together. This misclassification can be partly justified by the intrinsic approximation that voting procedures introduce when discretizing the parameter space; on the other hand, it is also worth noting that our procedure for (initial) parameter estimation generally exhibits a lower precision when applied to small and low-sampled segments, due a higher error in the tangent plane approximation.  In this case, the mean MFE over all segments is $0.0042$.

Finally, we test the resilience to increasing noise in Figure \ref{fig:noise} for a point cloud composed of $9,723$ points and $35$ segments; we added synthetic Gaussian noise of fixed mean $\mu=0$ and standard deviation $\sigma$ equals to $0.10$, $0.25$ and $0.50$.  We note that, as the noise intensity increases, some  segments start being misclassified; more specifically, when $\sigma=0.50$, two cylinders are mislabeled as cones and thus wrongly clustered. 
The mean MFE over all segments in the three cases is, respectively, $0.0106$, $0.0212$ and $0.0347$. 

\newcolumntype{P}[1]{>{\centering\arraybackslash}p{#1}}
\begin{figure}[h]
    \begin{center}
        \footnotesize
        \begin{tabular}{|P{2.75cm}|P{2.75cm}|P{2.75cm}|P{2.75cm}|}
            \hline
            \rowcolor{teal!25}\makecell{Model \\ and planes} &  Cylinders & Cones & Tori \\\hline
             \raisebox{-0.50\height}{\includegraphics[scale=0.110, trim={0cm 0cm 0cm 0cm}, clip]{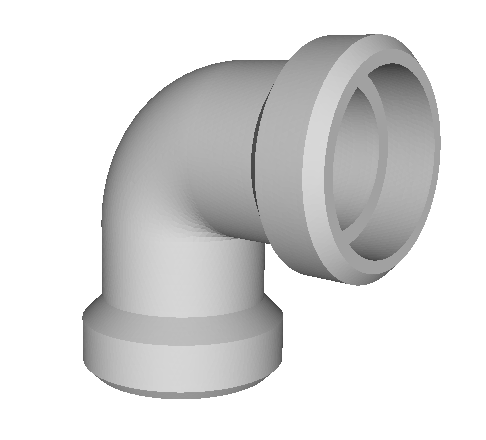}}
             &
             \raisebox{-0.5\height}{\includegraphics[scale=0.25, trim={1cm 1cm 1cm 1cm}, clip]{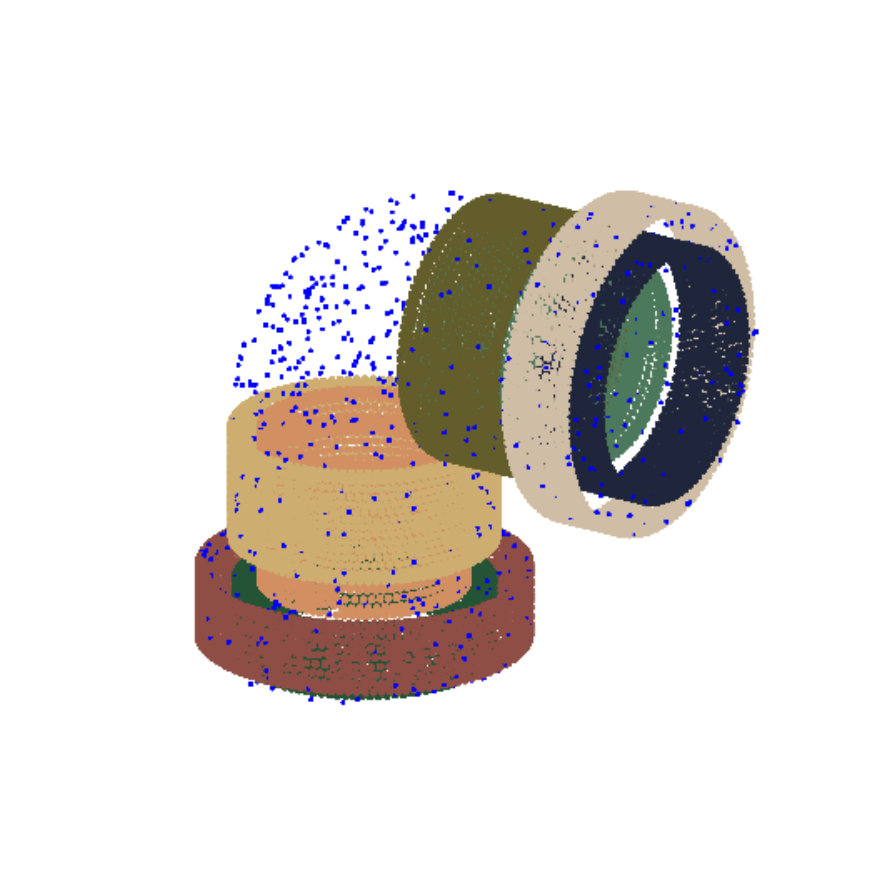}}
             &
             \raisebox{-0.5\height}{\includegraphics[scale=0.25, trim={1cm 1cm 1cm 1cm}, clip]{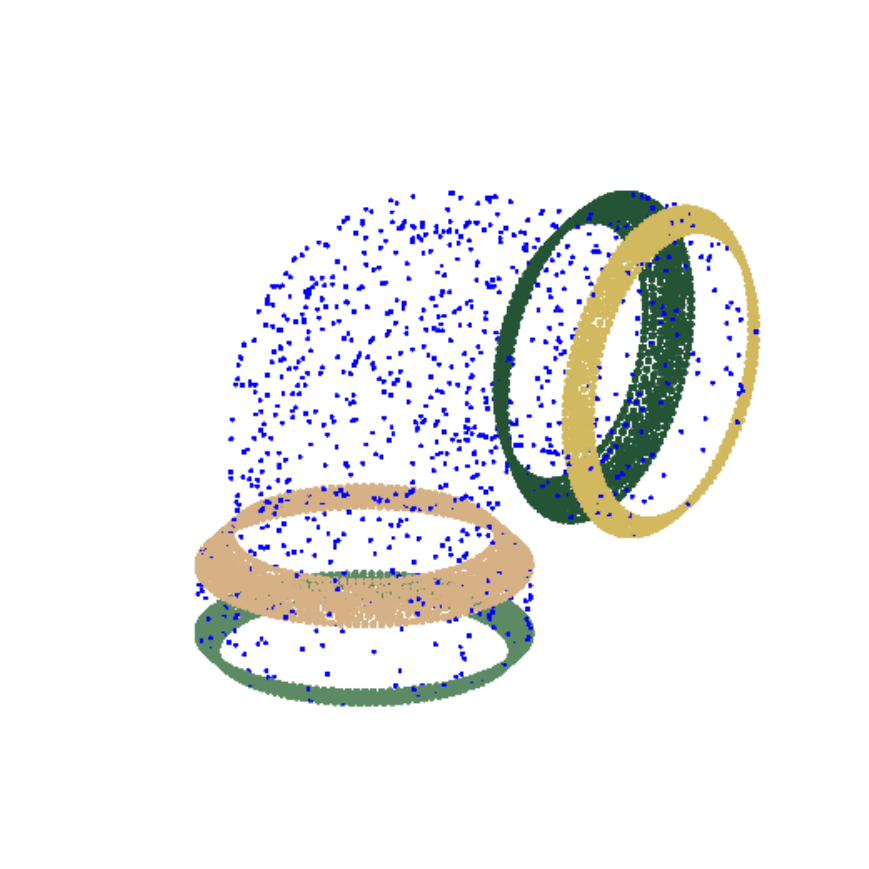}}
             &
             \raisebox{-0.5\height}{\includegraphics[scale=0.25, trim={1cm 1cm 1cm 1cm}, clip]{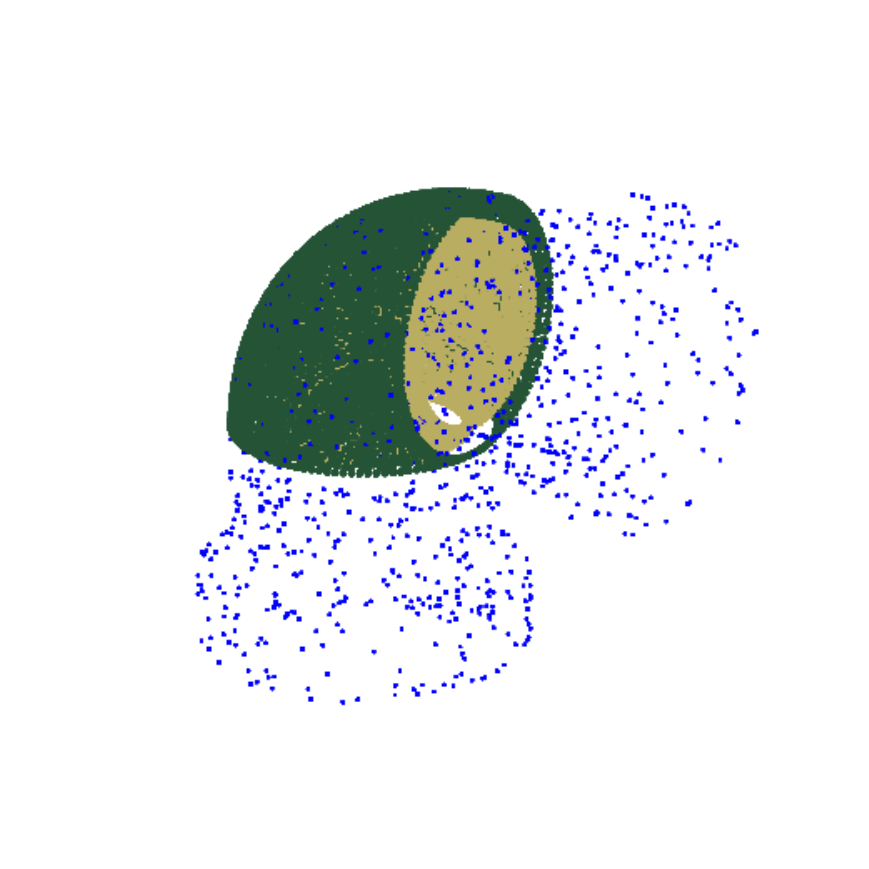}}
             \\
             Original model 
             &
             Same cylinder
             & 
             Same cone
             & 
             Same torus
             \\\hline
             \raisebox{-0.5\height}{\includegraphics[scale=0.25, trim={1cm 1cm 1cm 1cm}, clip]{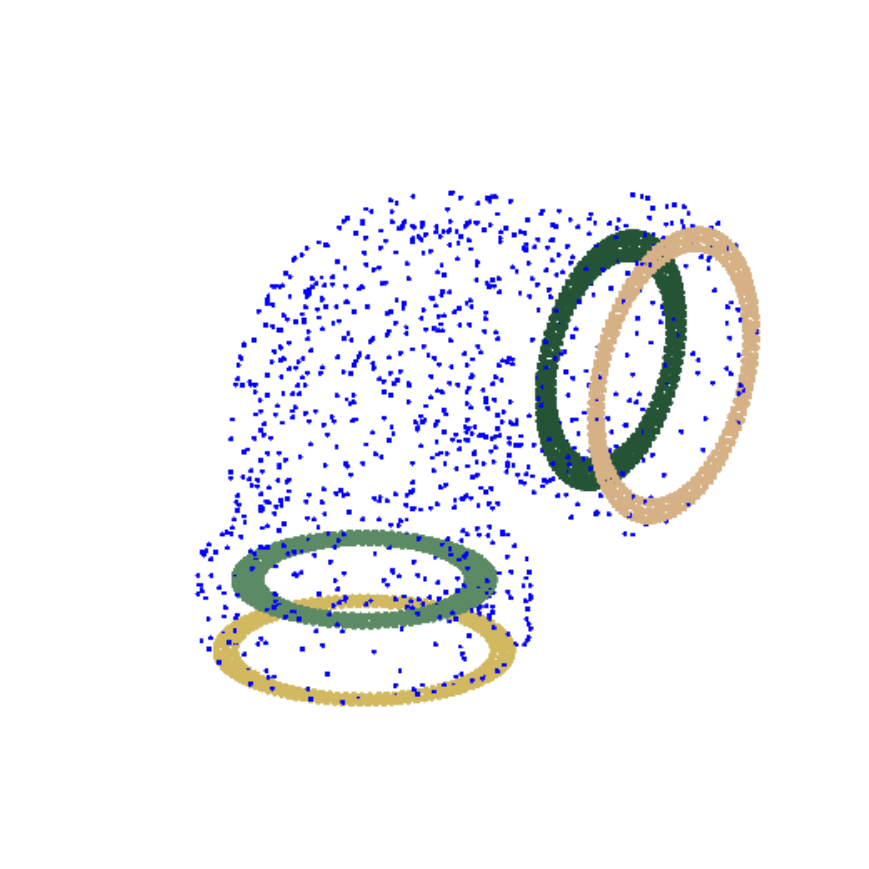}}
             &
             \raisebox{-0.5\height}{\includegraphics[scale=0.25, trim={1cm 1cm 1cm 1cm}, clip]{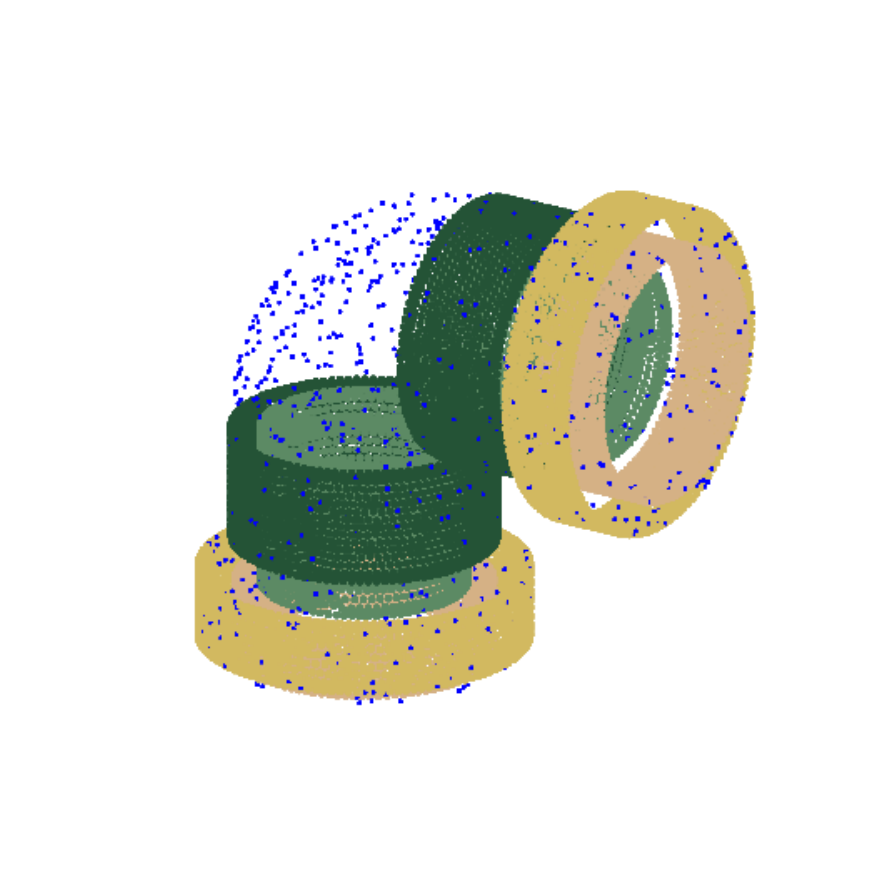}}
             &
             \raisebox{-0.5\height}{\includegraphics[scale=0.25, trim={1cm 1cm 1cm 1cm}, clip]{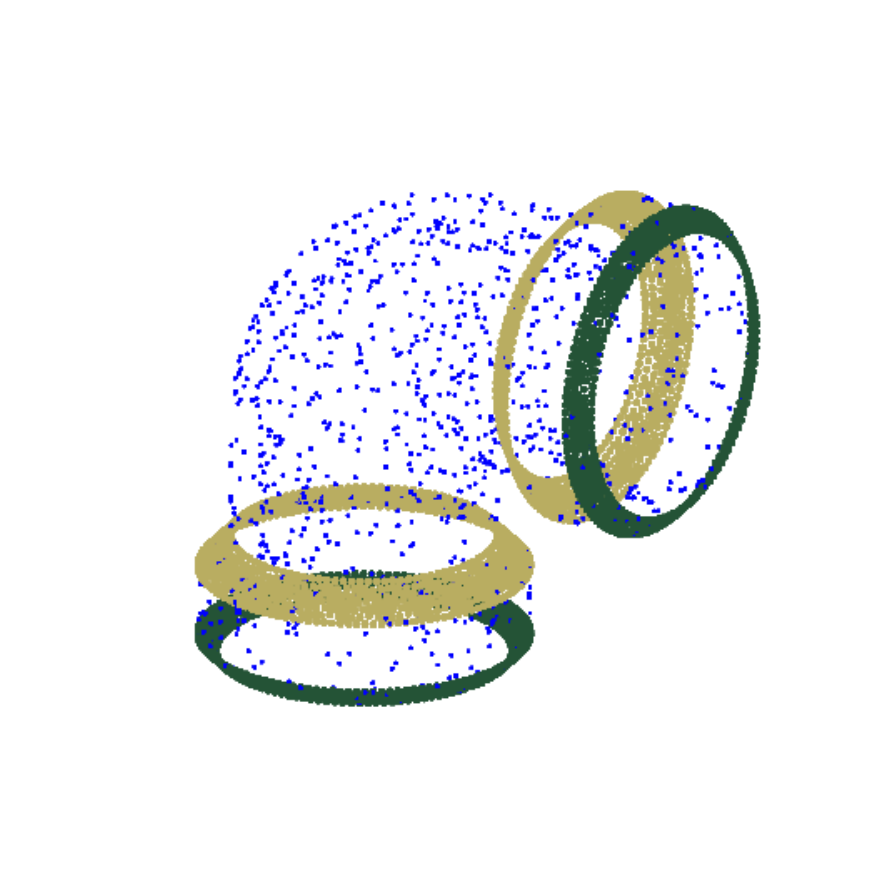}}
             &
             \raisebox{-0.5\height}{\includegraphics[scale=0.25, trim={1cm 1cm 1cm 1cm}, clip]{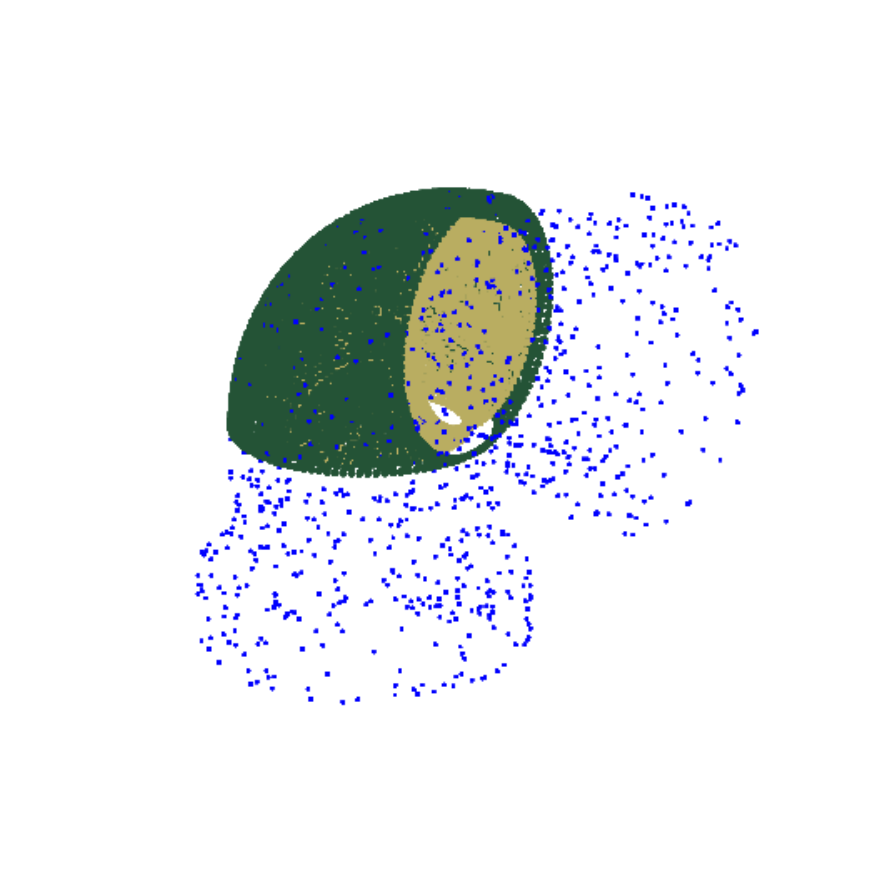}}
             \\
             Same plane
             &
             Same radius
             & 
             Same radius
             & 
             Same radius
             \\\hline
             \raisebox{-0.5\height}{\includegraphics[scale=0.25, trim={1cm 1cm 1cm 1cm}, clip]{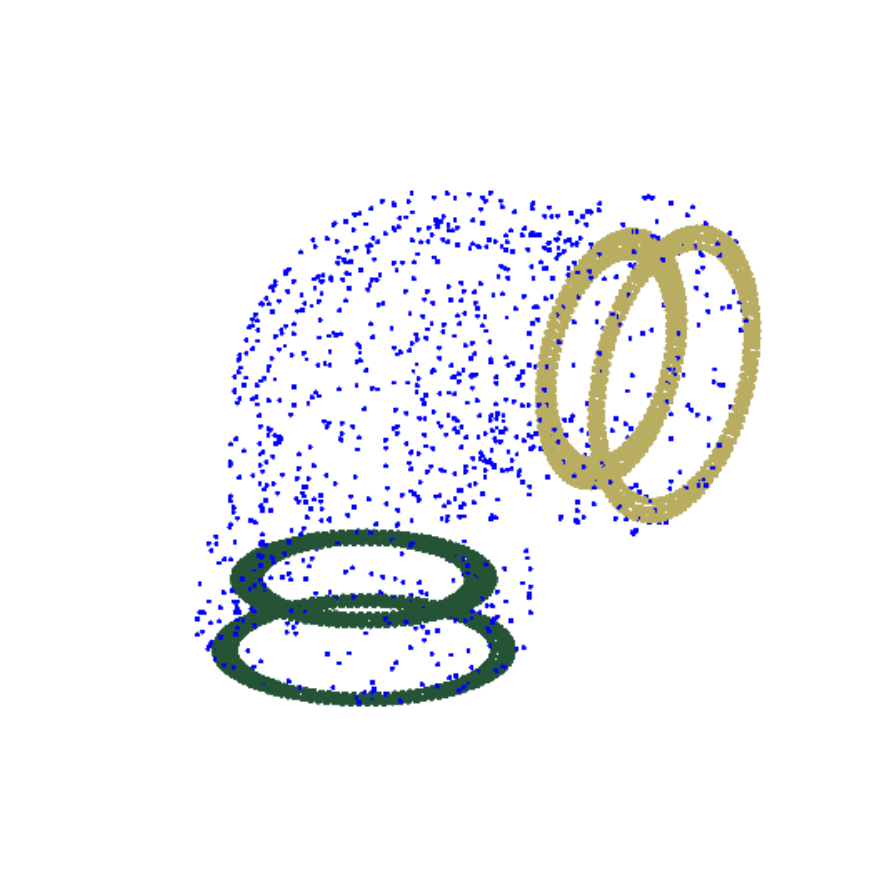}}
             &
             \raisebox{-0.5\height}{\includegraphics[scale=0.25, trim={1cm 1cm 1cm 1cm}, clip]{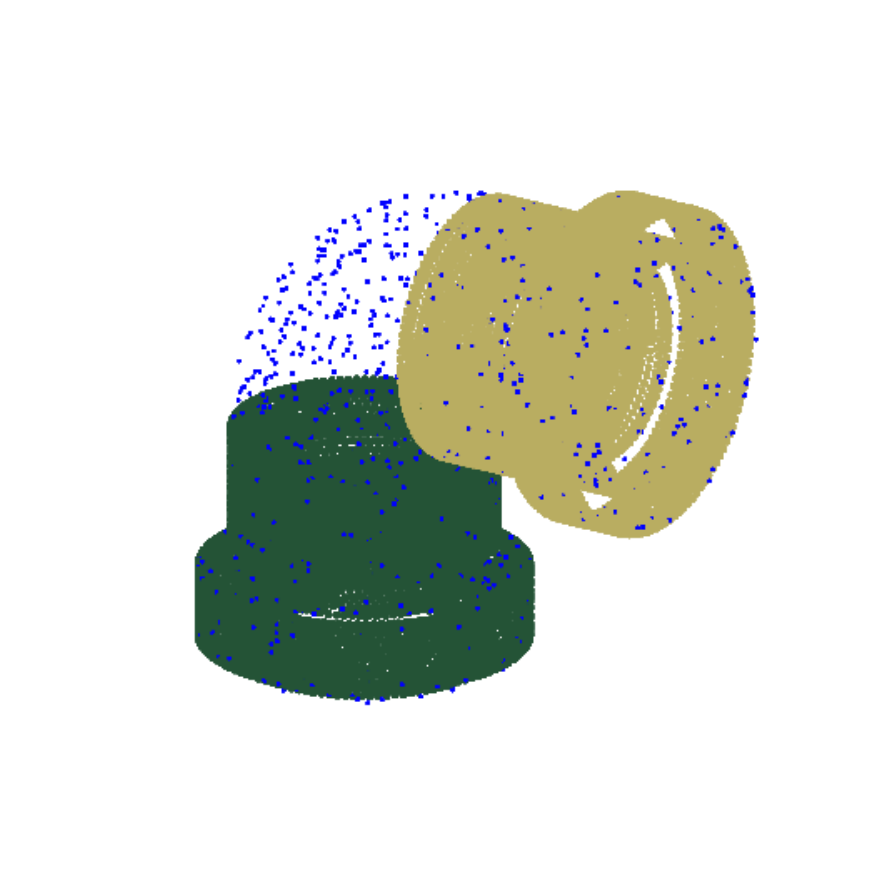}}
             &
             \raisebox{-0.5\height}{\includegraphics[scale=0.25, trim={1cm 1cm 1cm 1cm}, clip]{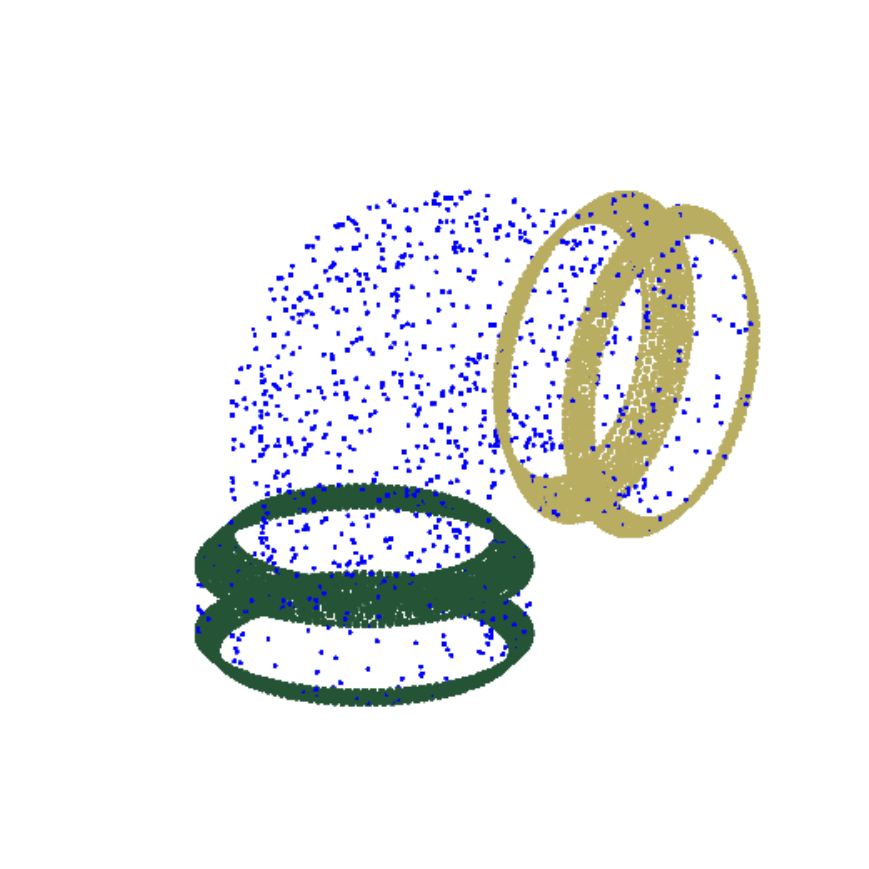}}
             &
             \raisebox{-0.5\height}{\includegraphics[scale=0.25, trim={1cm 1cm 1cm 1cm}, clip]{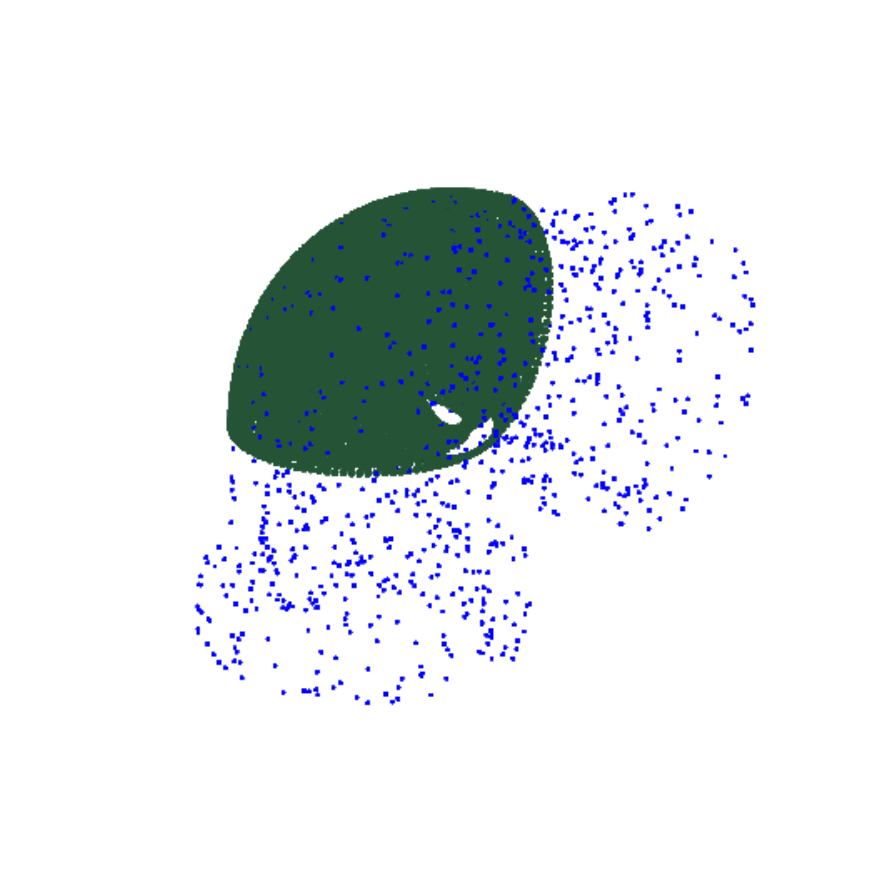}}
             \\
             Parallel planes 
             &
             Same rotational axis
             & 
             Same rotational axis
             & 
             Same rotational axis
             \\\hline
            \end{tabular}
    \end{center}
    \caption{A model from \cite{Koch:2019}: we show segments of the same primitive type exhibiting different similarities. Colours are used to visually represent primitives sharing the same property.
    \label{fig:tubo}}
\end{figure}

\newcolumntype{P}[1]{>{\centering\arraybackslash}p{#1}}
\begin{figure}[h]
    \begin{center}
        \footnotesize
        \begin{tabular}{cc}
        \begin{tabular}{|P{2.2cm}|P{2.2cm}|P{2.2cm}|}
            \hline
            \rowcolor{teal!25}\makecell{Model \\ and planes} &  Cylinders & Tori \\\hline
            \rowcolor{white!25}\raisebox{-0.5\height}{\includegraphics[scale=0.055, trim={0cm 0cm 0cm 0cm}, clip]{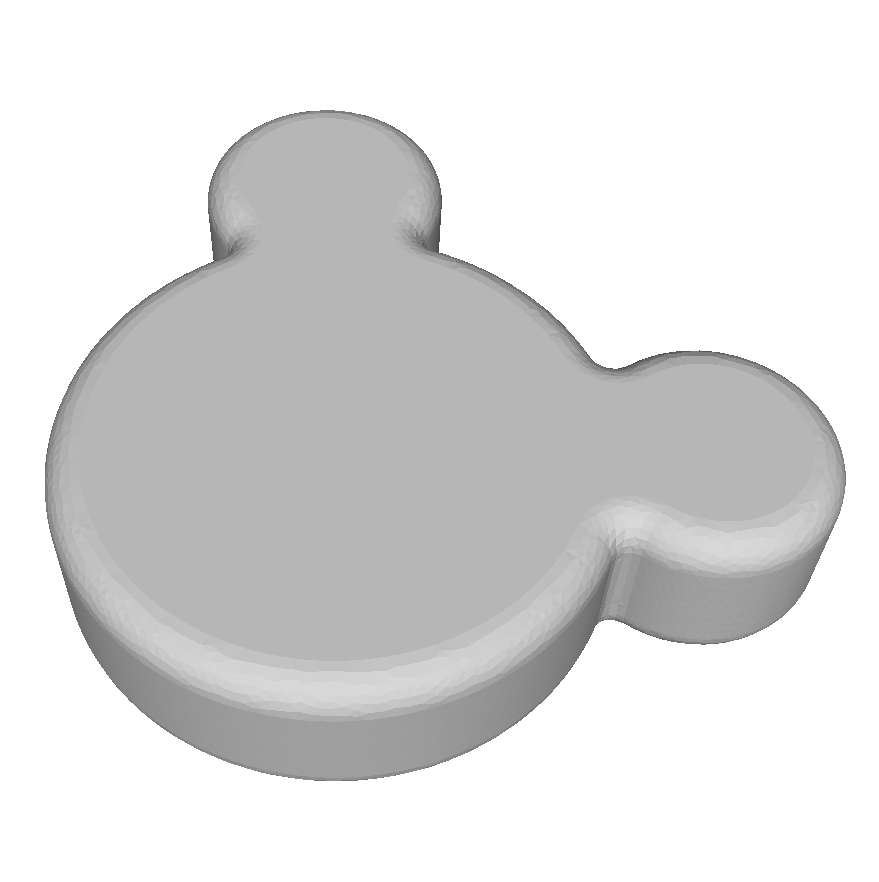}}
             &
             \raisebox{-0.5\height}{\includegraphics[scale=0.3, trim={1cm 2cm 1cm 1cm}, clip]{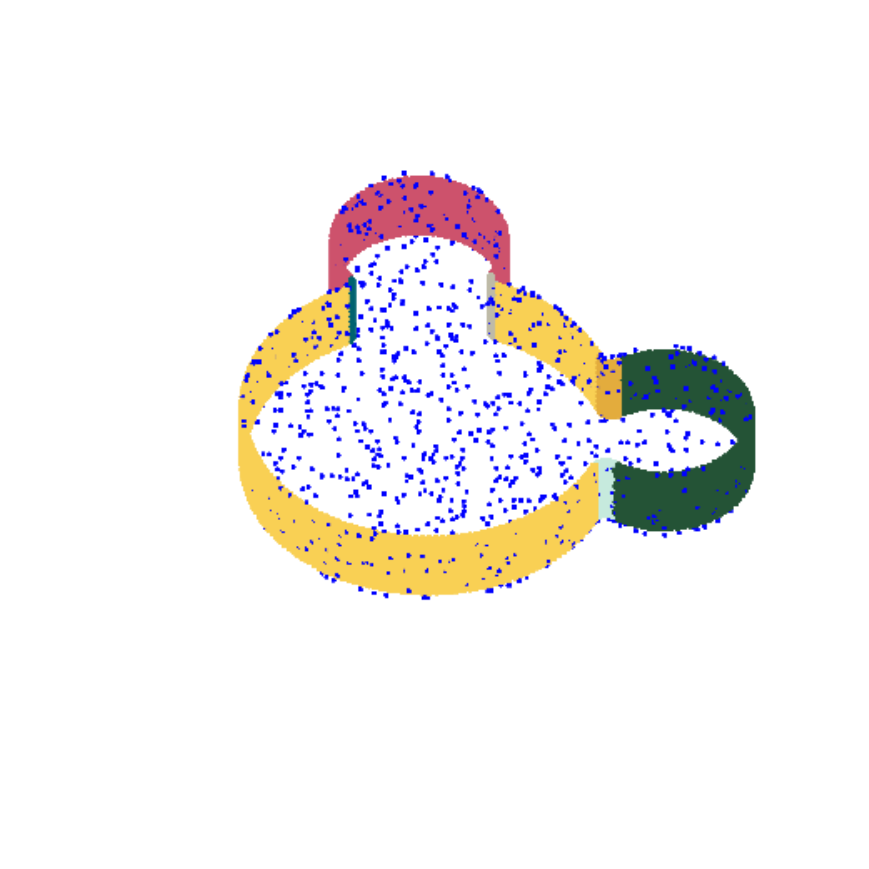}}
             &
             \raisebox{-0.5\height}{\includegraphics[scale=0.3, trim={1cm 2cm 1cm 1cm}, clip]{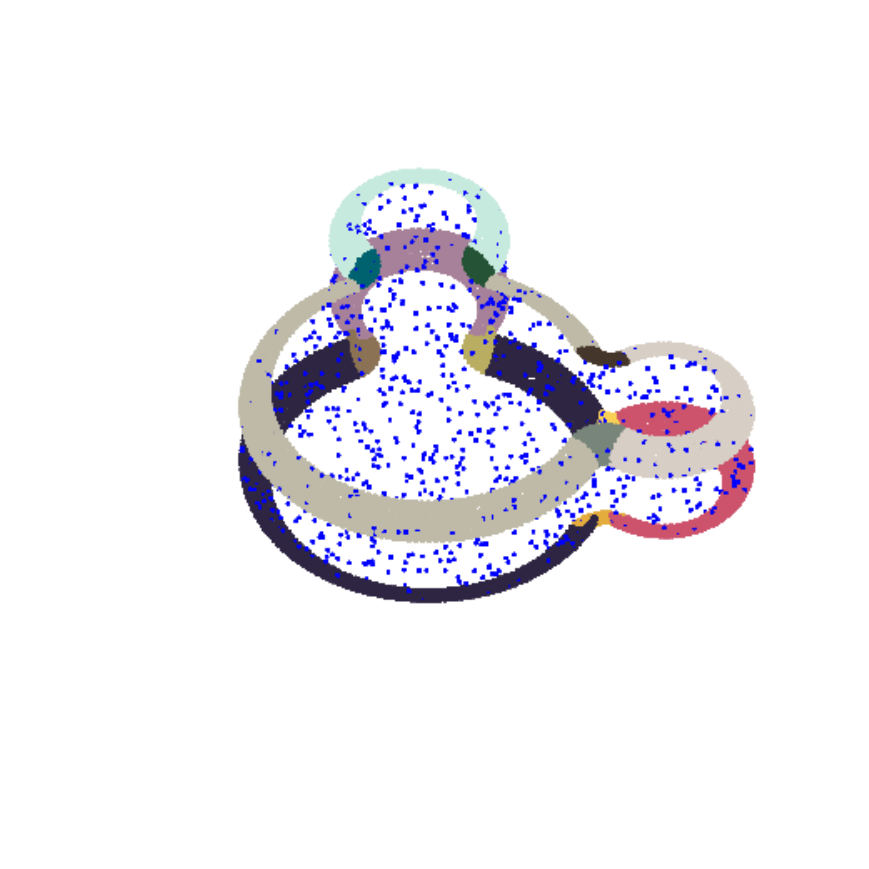}}
             \\
             Original model 
             &
             Same cylinder
             & 
             Same torus
             \\\hline
             \raisebox{-0.4\height}{\includegraphics[scale=0.3, trim={1cm 2cm 1cm 1cm}, clip]{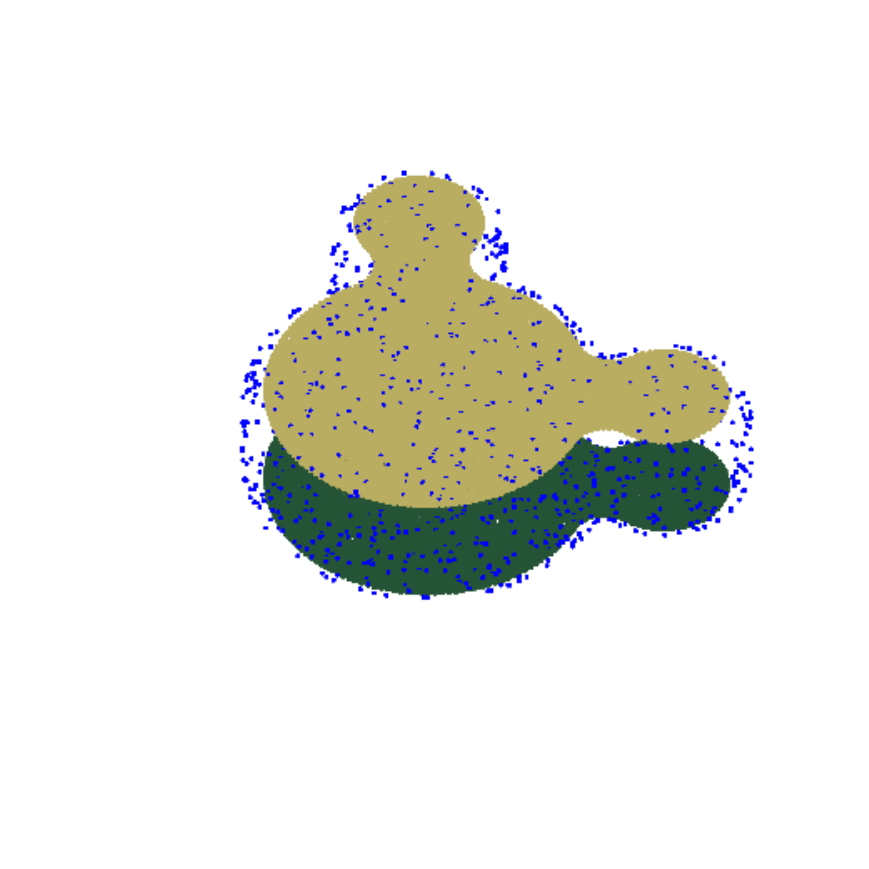}}
             &
             \raisebox{-0.4\height}{\includegraphics[scale=0.3, trim={1cm 2cm 1cm 1cm}, clip]{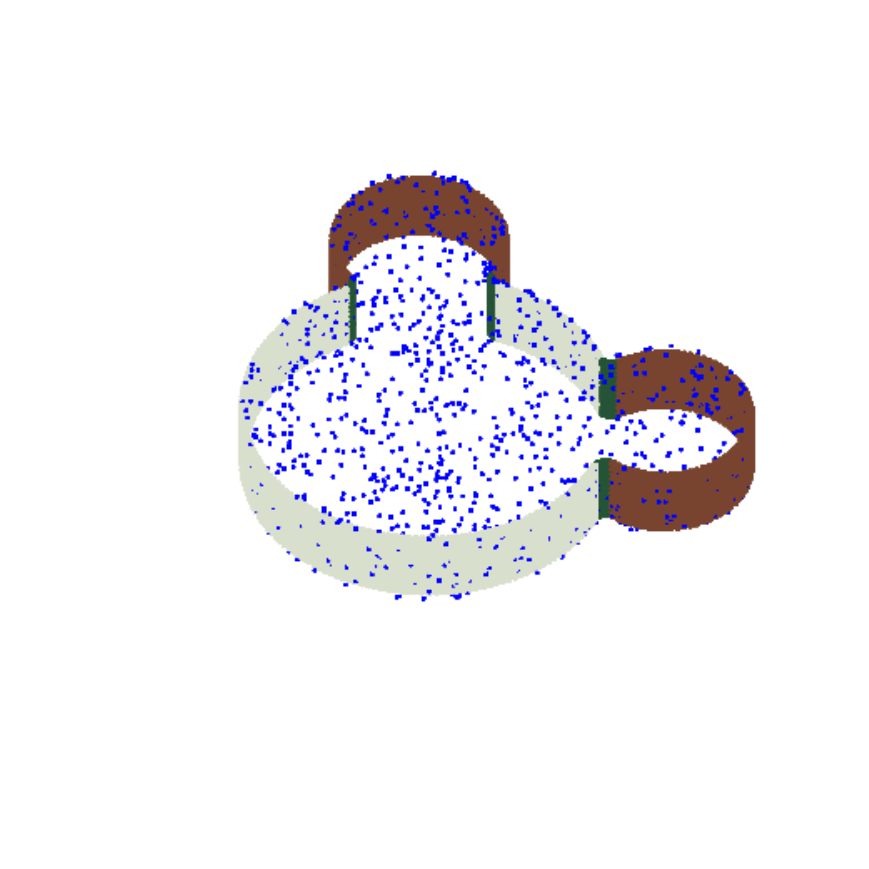}}
             &
             \raisebox{-0.4\height}{\includegraphics[scale=0.3, trim={1cm 2cm 1cm 1cm}, clip]{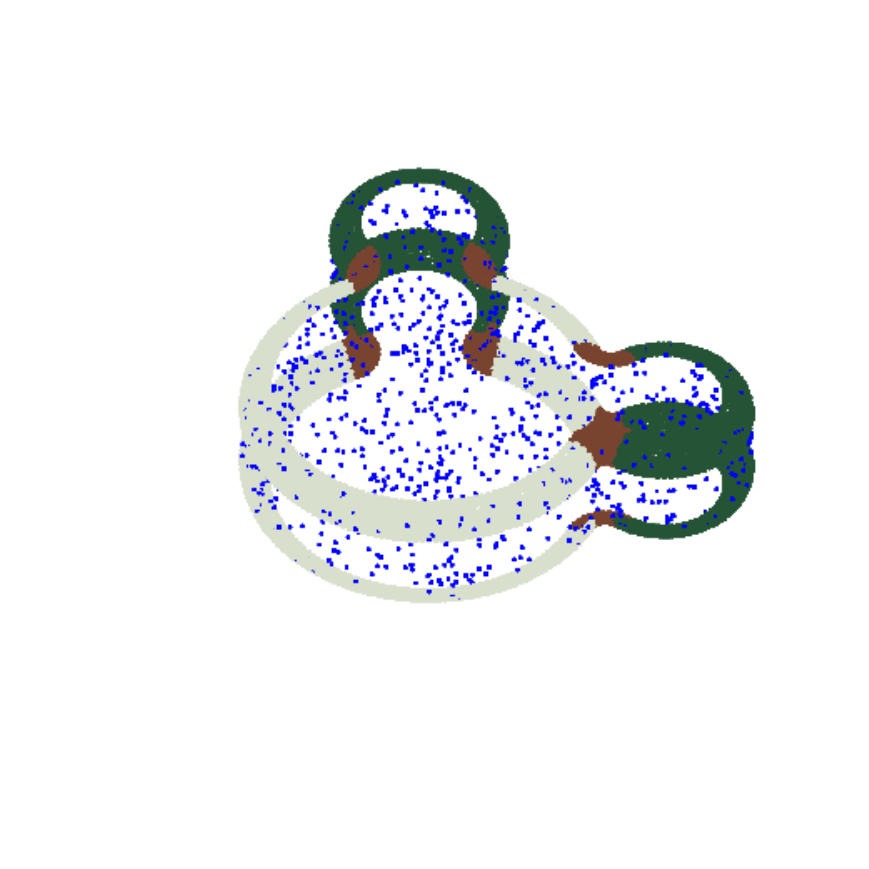}}
             \\
             Same plane
             &
             Same radius
             & 
             Same radius
             \\\hline
             \raisebox{-0.4\height}{\includegraphics[scale=0.3, trim={1cm 2cm 1cm 1cm}, clip]{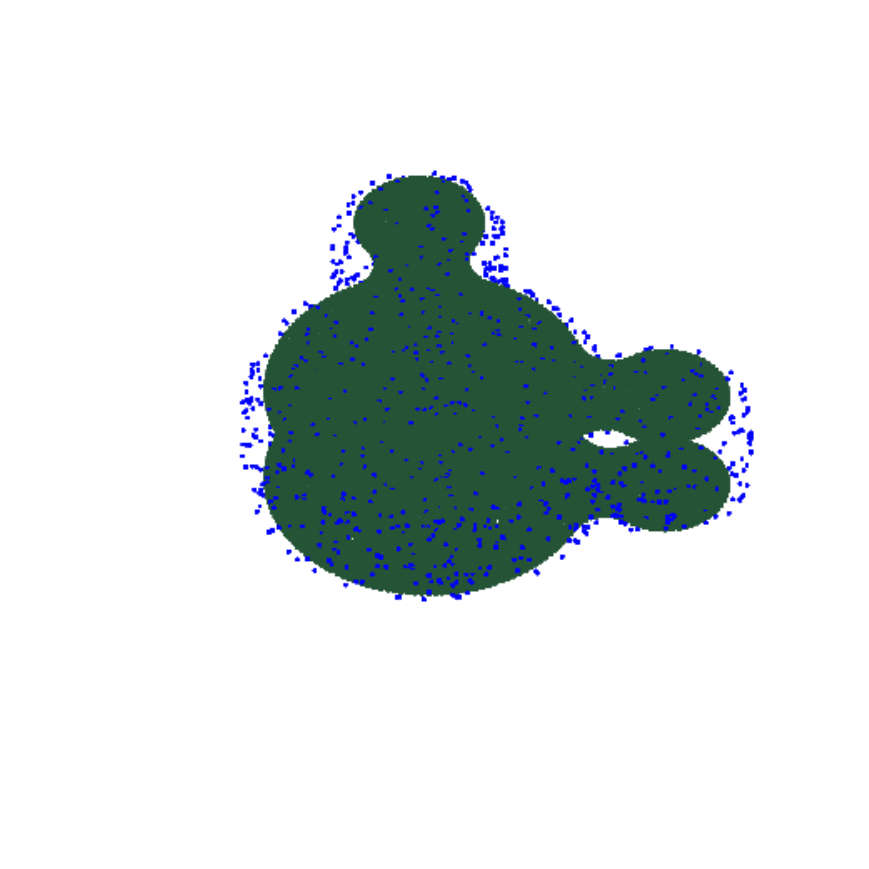}}
             &
             \raisebox{-0.4\height}{\includegraphics[scale=0.28, trim={1cm 2cm 1cm 1cm}, clip]{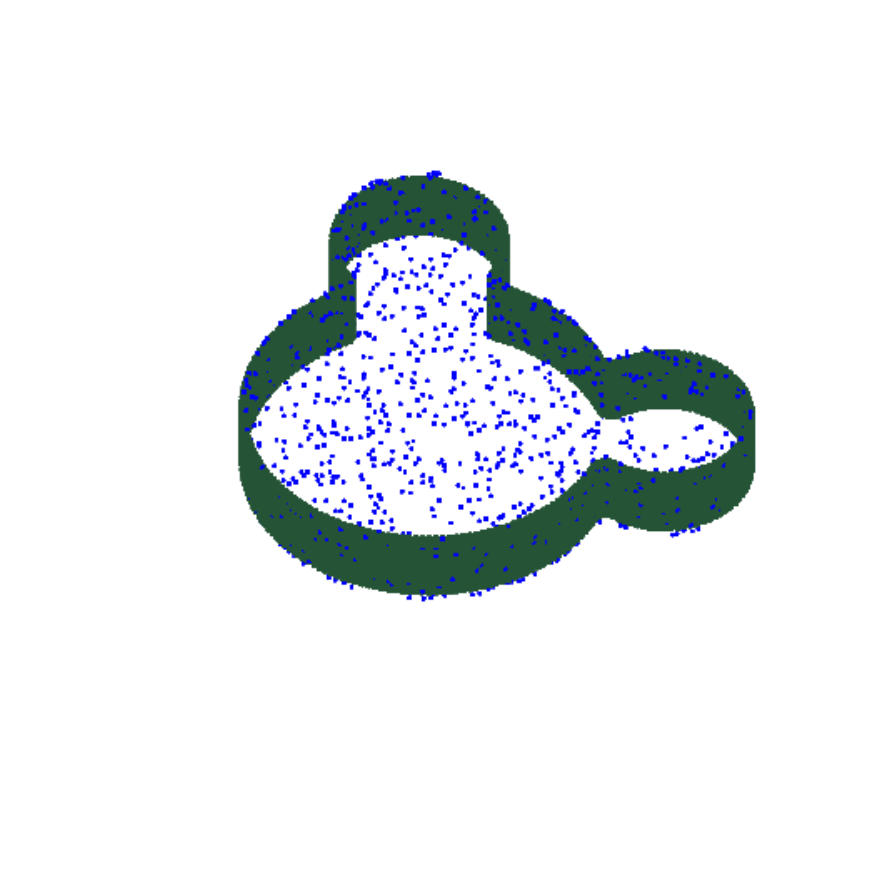}}
             &
             \raisebox{-0.4\height}{\includegraphics[scale=0.28, trim={1cm 2cm 1cm 1cm}, clip]{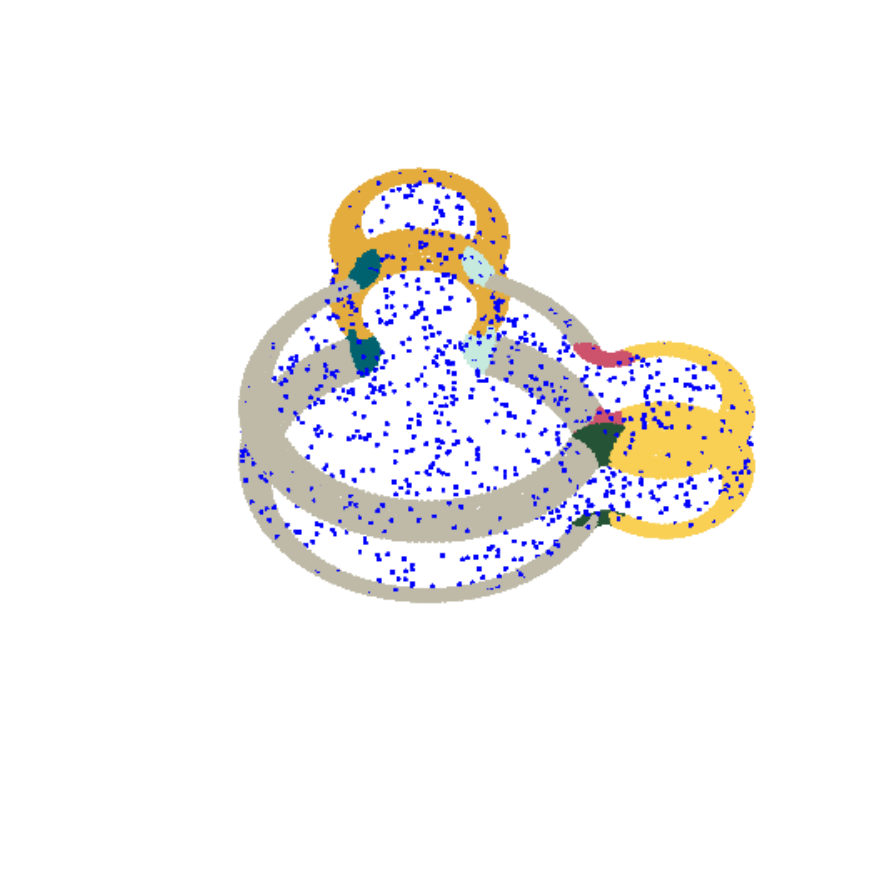}}
             \\
             Parallel planes 
             &
             \makecell{Parallel \\rotational axes}
             & 
             \makecell{Same \\rotational axis}
             \\\hline
            \end{tabular}
            &
            \begin{tabular}{|P{2.25cm}|P{2.25cm}|P{2.25cm}|}
            \hline
            \rowcolor{teal!25} Planes & Cylinders & \makecell{Model \\ and spheres}\\\hline
            \rowcolor{white!25}\raisebox{-0.4\height}{\includegraphics[scale=0.375, trim={1cm 2cm 1cm 1cm}, clip]{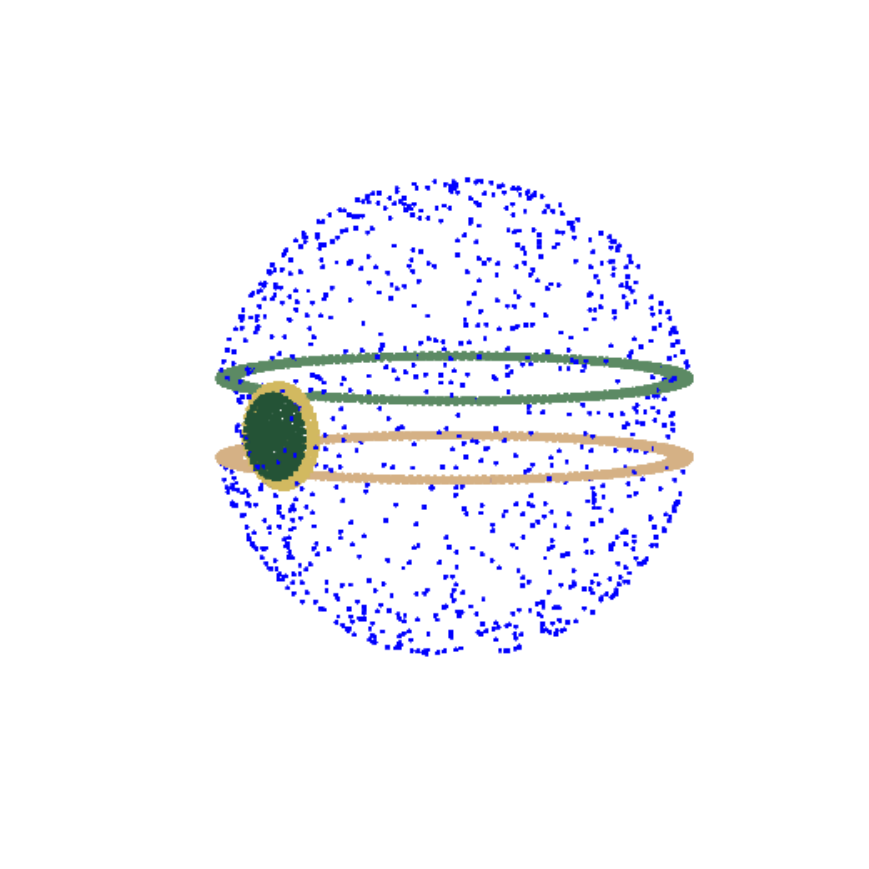}}
            &
            \raisebox{-0.4\height}{\includegraphics[scale=0.350, trim={1cm 2cm 1cm 1cm}, clip]{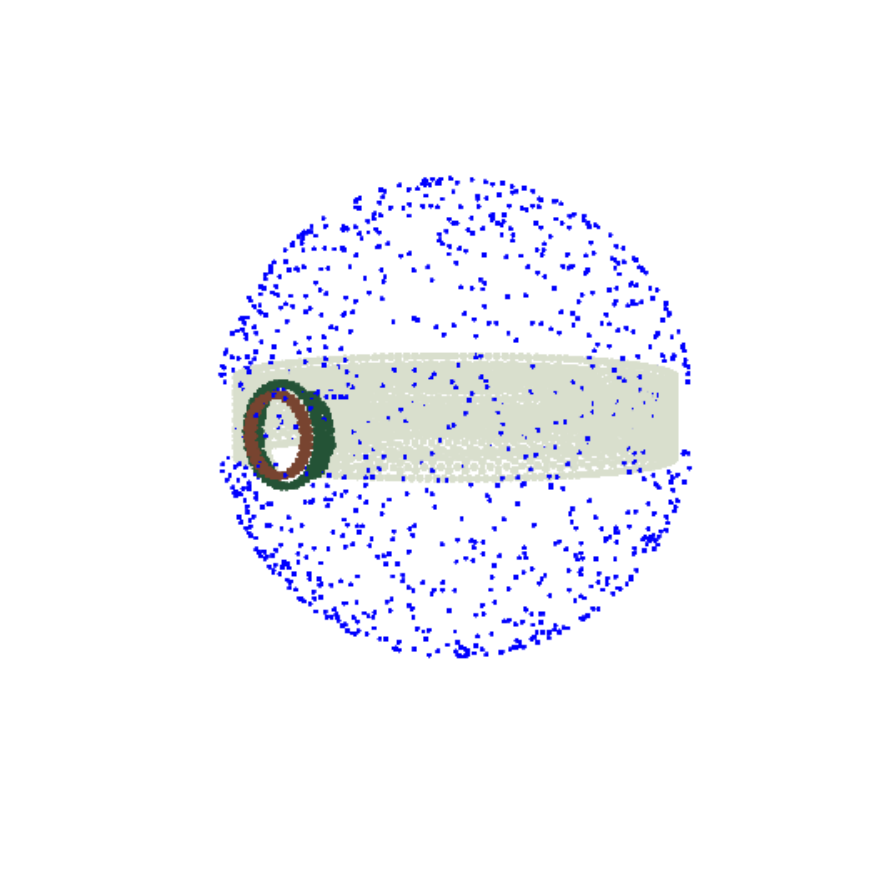}}
            &
            \raisebox{-0.4\height}{\includegraphics[scale=0.10, trim={1cm 1cm 1cm 0cm}, clip]{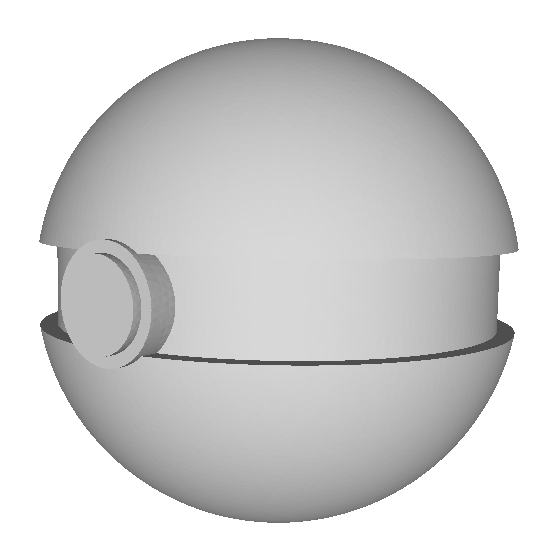}}
            \\
            Same plane 
            &
            Same cylinder
            &
            Original model
            \\\hline
            \raisebox{-0.4\height}{\includegraphics[scale=0.350, trim={1cm 2cm 1cm 1cm}, clip]{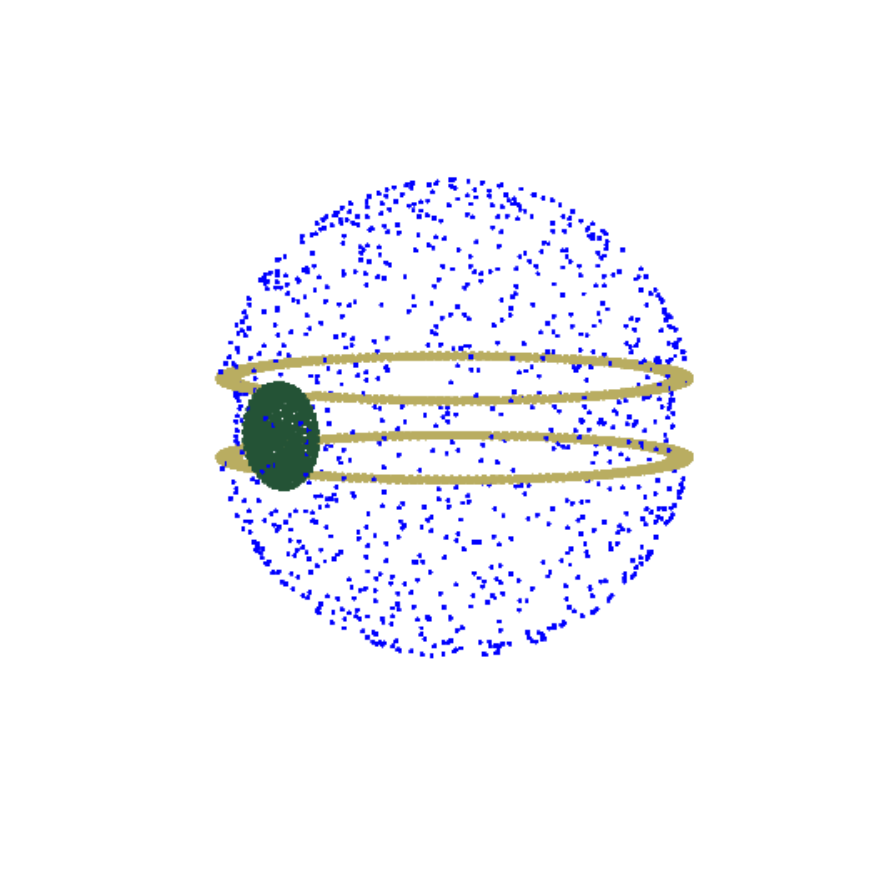}}
            &
            \raisebox{-0.4\height}{\includegraphics[scale=0.350, trim={1cm 2cm 1cm 1cm}, clip]{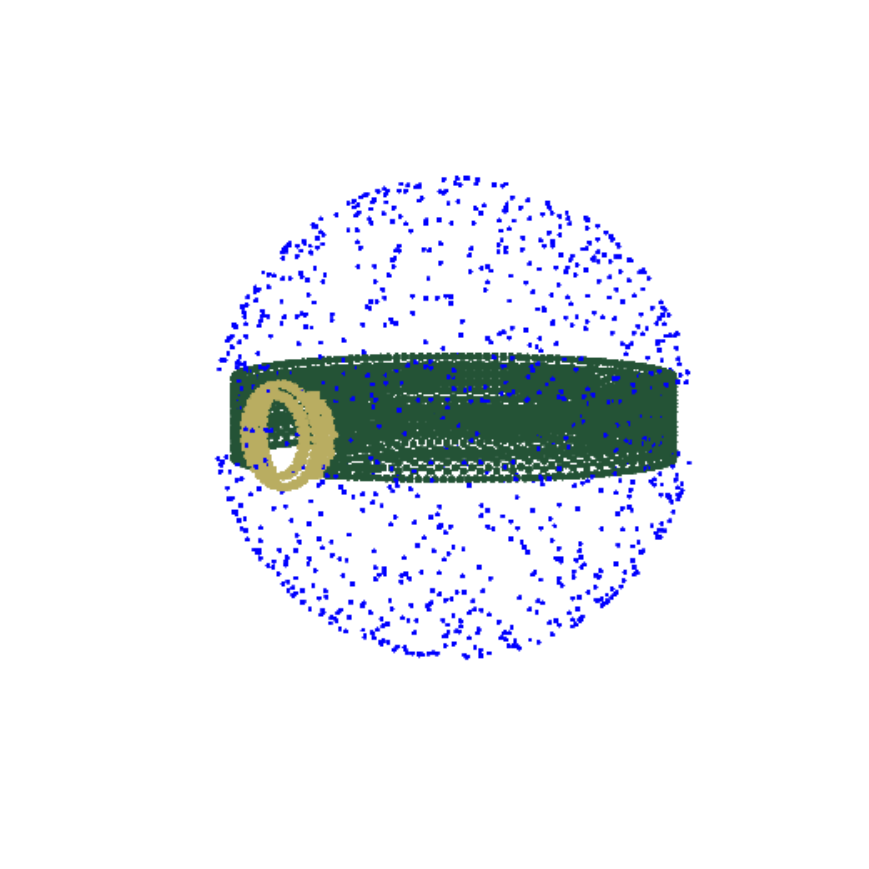}}
            &
            \raisebox{-0.4\height}{\includegraphics[scale=0.350, trim={1cm 2cm 1cm 1cm}, clip]{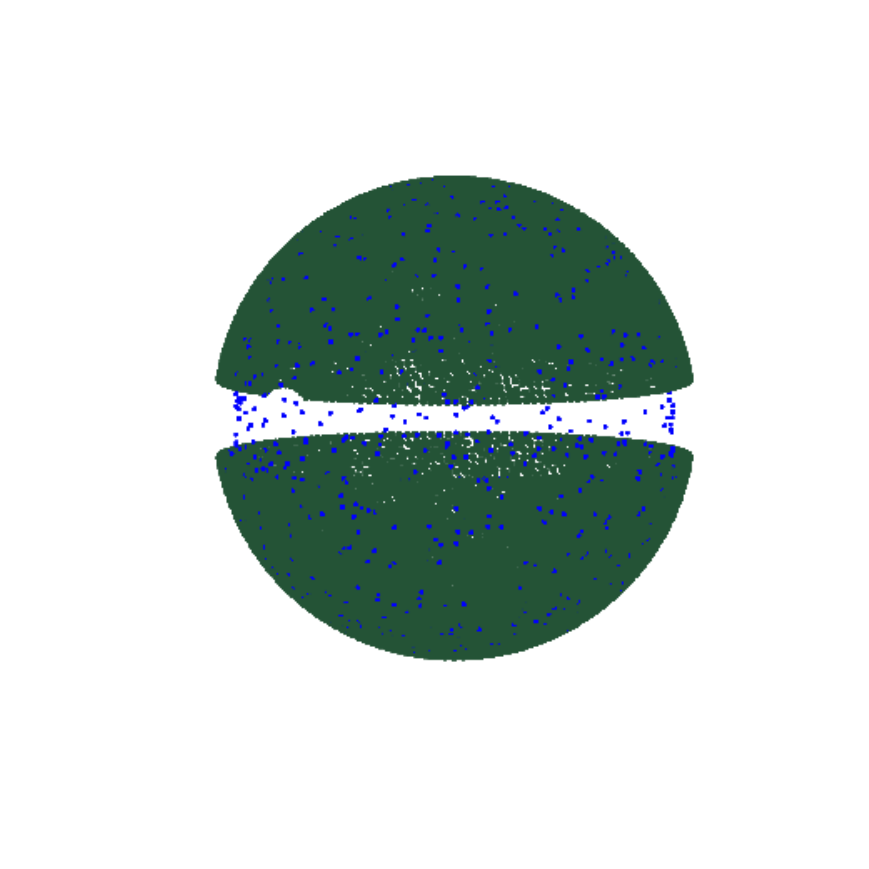}}
            \\
            Parallel planes
            &
            \makecell{Parallel \\rotational axes}
            &
            Same sphere
            \\\hline
        \end{tabular}
            \\
             & \\
            (a) & (b)\\
    \end{tabular}
    \end{center}
    \caption{Two models from \cite{Koch:2019}: segments exhibit different similarities.
    \label{fig:topolino_pokeball}}
\end{figure}

\begin{figure}[h!]
\footnotesize
    \begin{center}
        \begin{tabular}{| >{\centering} z{1em} |c|cc|cc|}
            \hline
            \rowcolor{teal!25} & Point cloud &  \multicolumn{2}{c}{ Planes} & \multicolumn{2}{c|}{ Cylinders} \\\hline
            \rotatebox[origin=c]{90}{(a) $\sigma=0.10$} & \raisebox{-0.5\height}{\includegraphics[scale=0.380, trim={1.3cm 1.1cm 1cm 0.8cm}, clip]{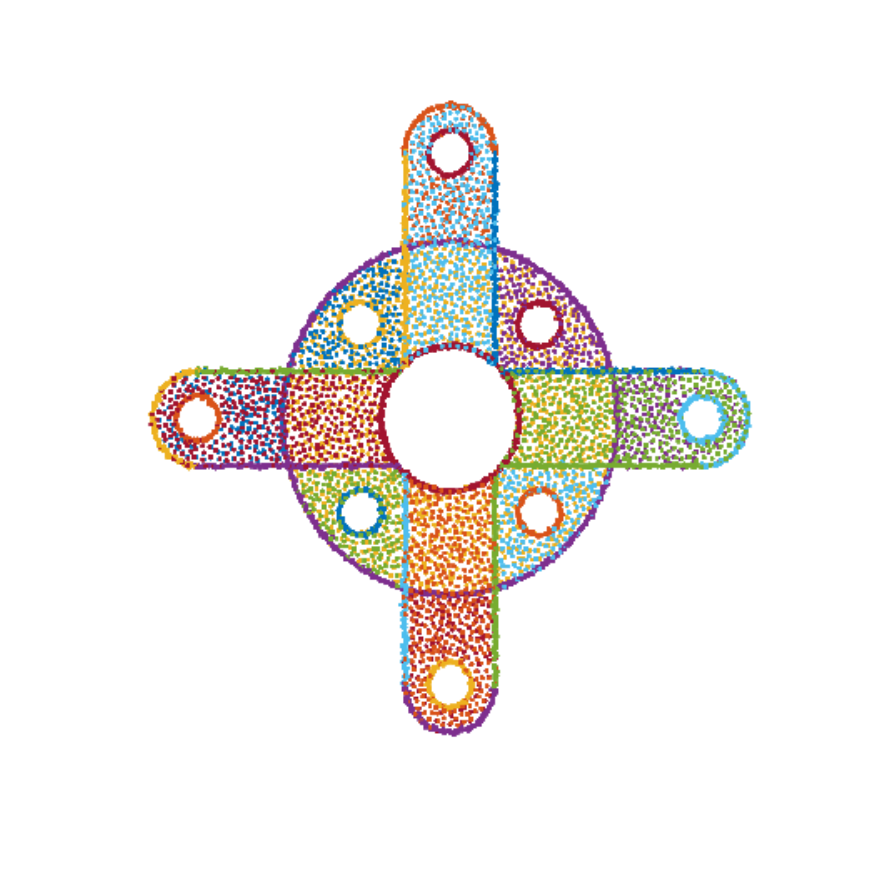}}
            &
            \raisebox{-0.5\height}{\includegraphics[scale=0.380, trim={1.3cm 1.1cm 1cm 1cm}, clip]{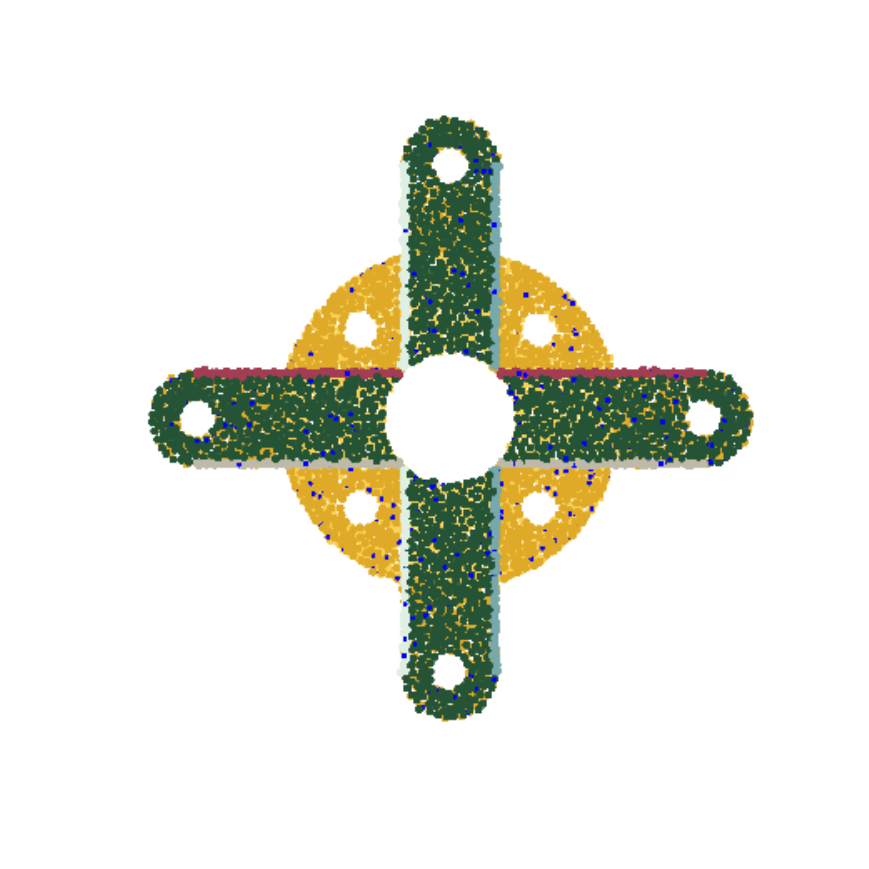}}
            &
            \raisebox{-0.5\height}{\includegraphics[scale=0.380, trim={1.3cm 1.1cm 1cm 1cm}, clip]{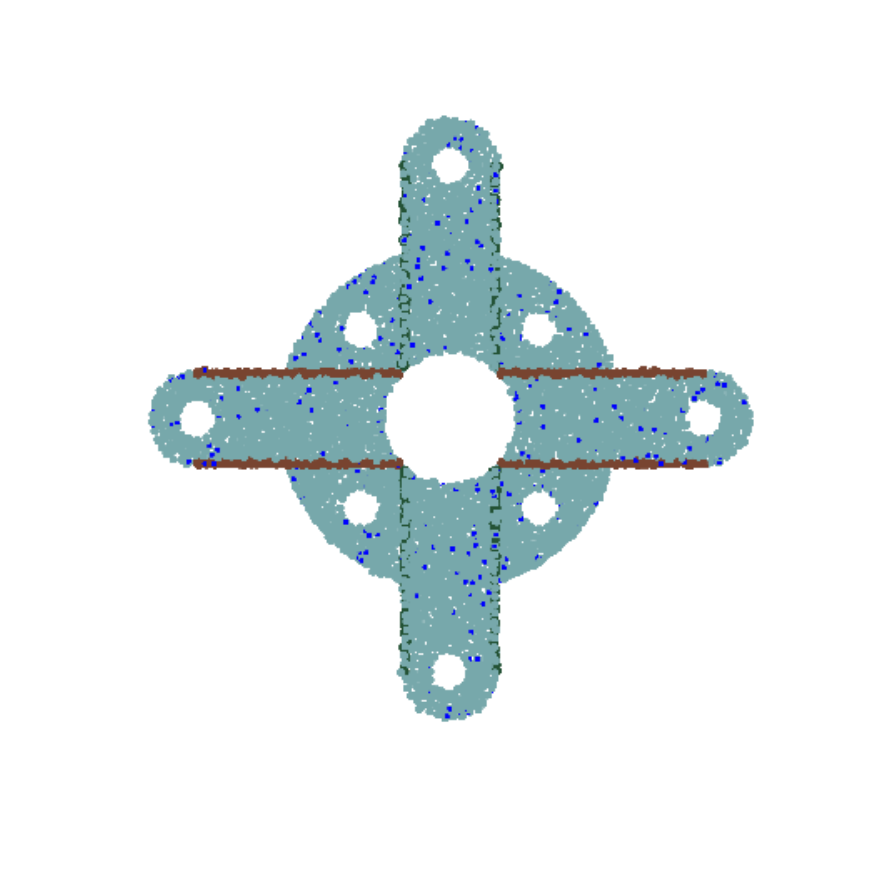}}
            &
            \raisebox{-0.5\height}{\includegraphics[scale=0.380, trim={1.3cm 1.1cm 1cm 1cm}, clip]{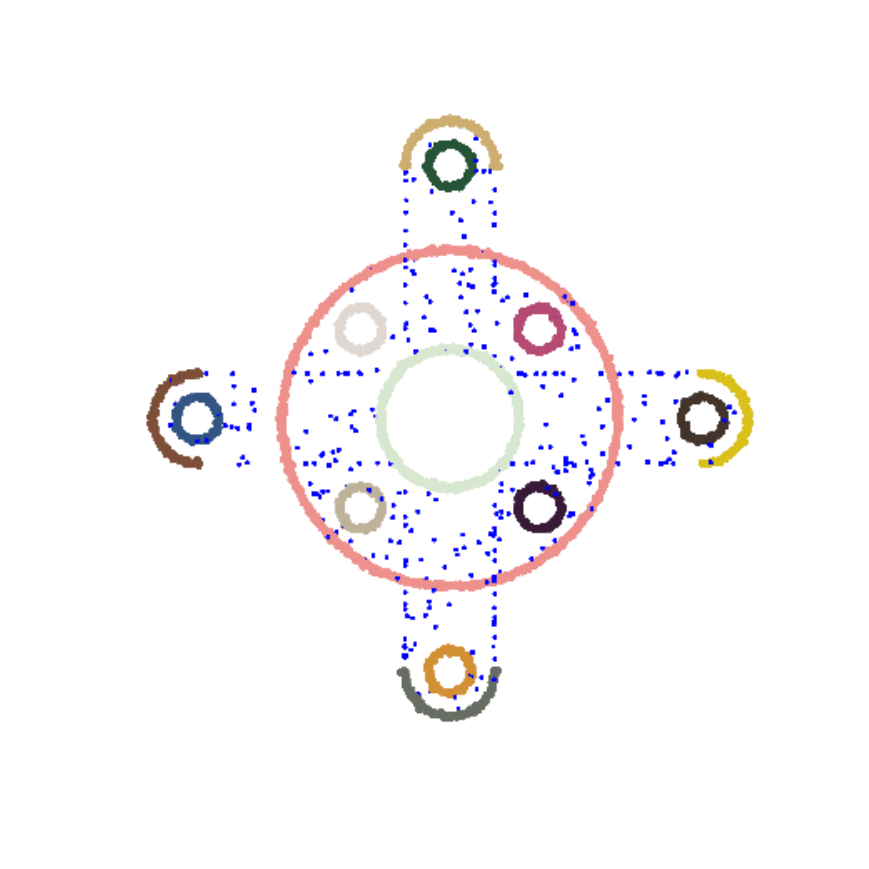}}
            &
            \raisebox{-0.5\height}{\includegraphics[scale=0.380, trim={1.3cm 1.1cm 1cm 1cm}, clip]{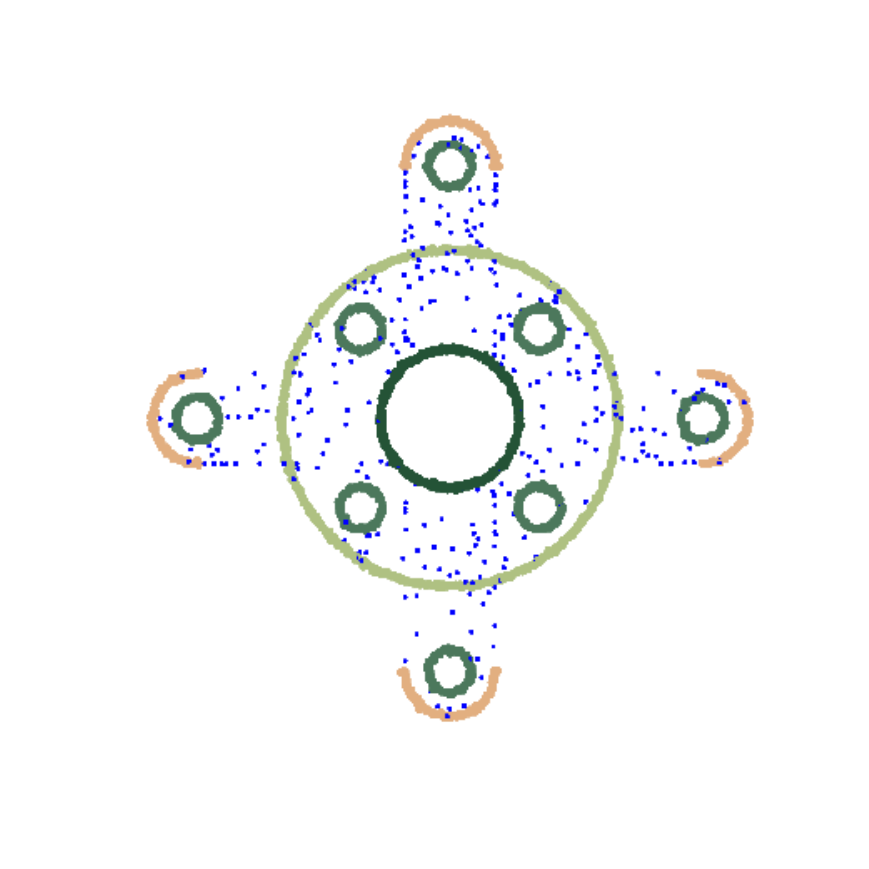}}
            \\
            &  & Same plane & Parallel planes & Same cylinder & Same radius 
            \\
            \hline
            \rotatebox[origin=c]{90}{(b) $\sigma=0.25$} & \raisebox{-0.5\height}{\includegraphics[scale=0.380, trim={1.3cm 1.1cm 1cm 0.8cm}, clip]{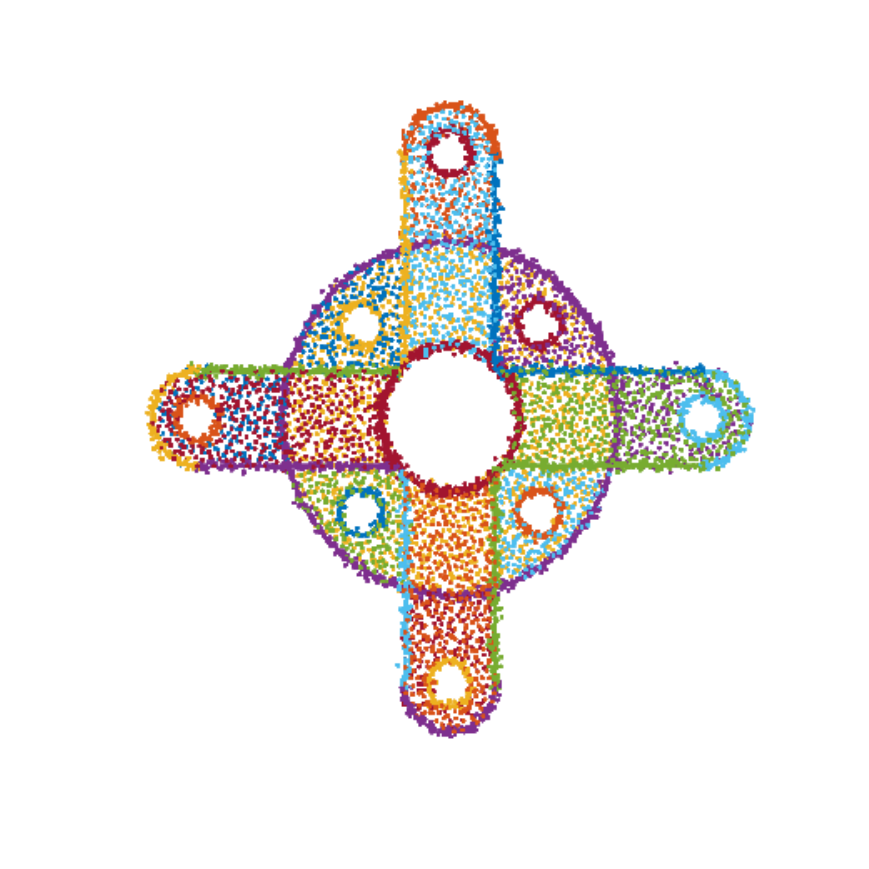}}
            &
            \raisebox{-0.5\height}{\includegraphics[scale=0.380, trim={1.3cm 1.1cm 1cm 1cm}, clip]{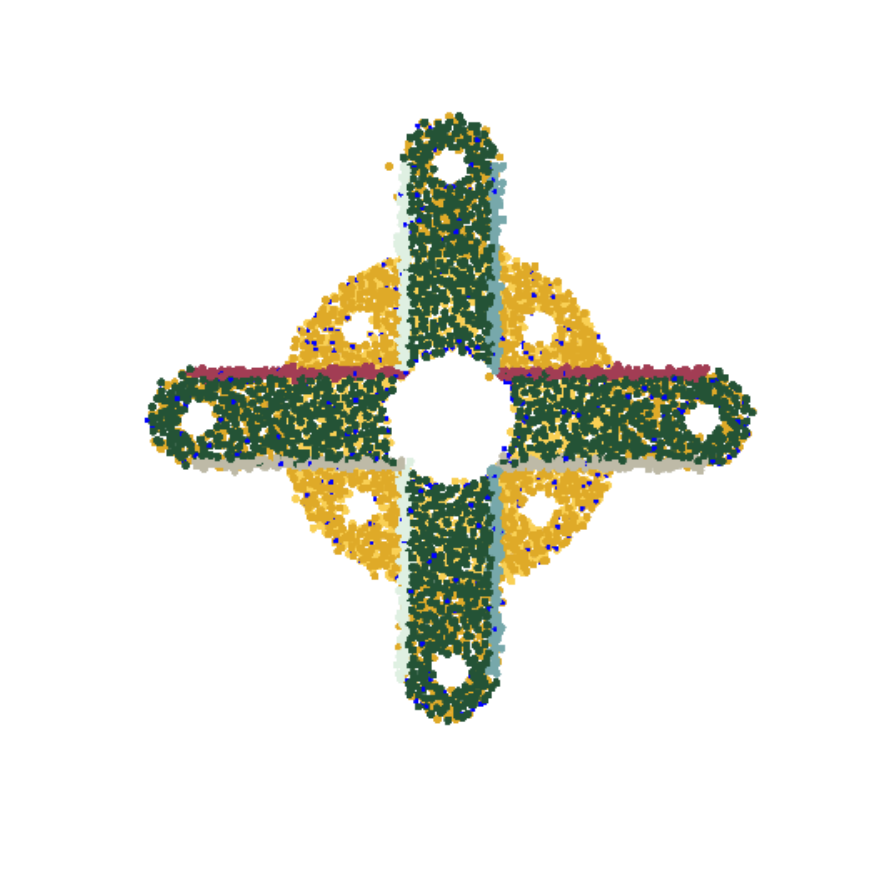}}
            &
            \raisebox{-0.5\height}{\includegraphics[scale=0.380, trim={1.3cm 1.1cm 1cm 1cm}, clip]{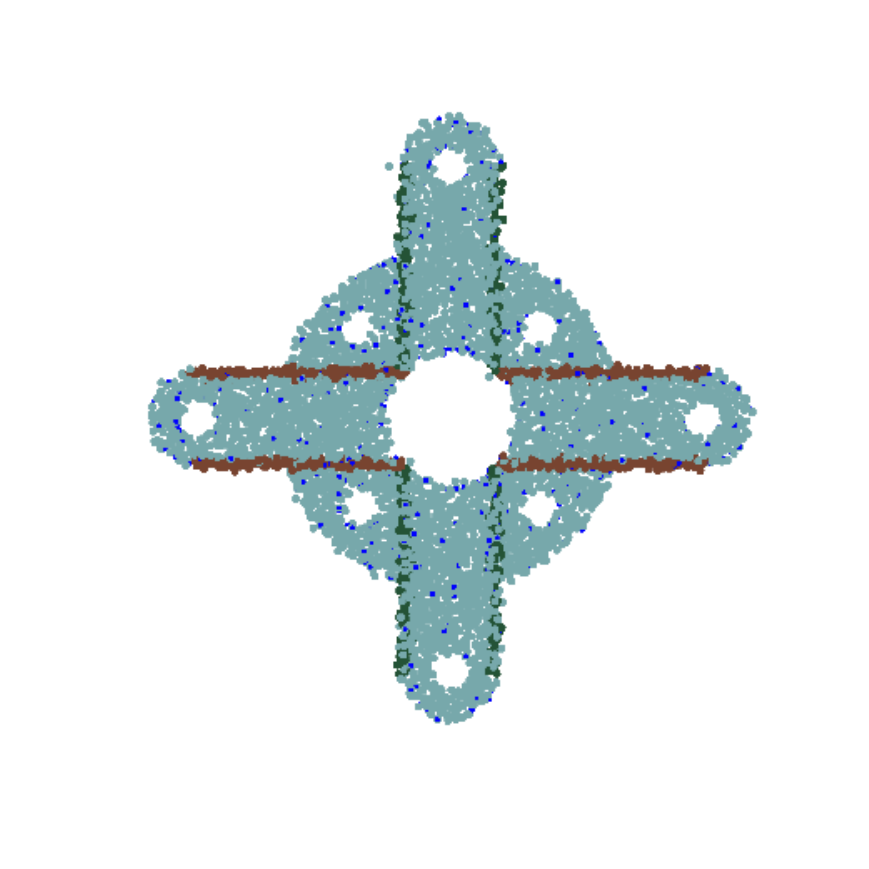}}
            &
            \raisebox{-0.5\height}{\includegraphics[scale=0.380, trim={1.3cm 1.1cm 1cm 1cm}, clip]{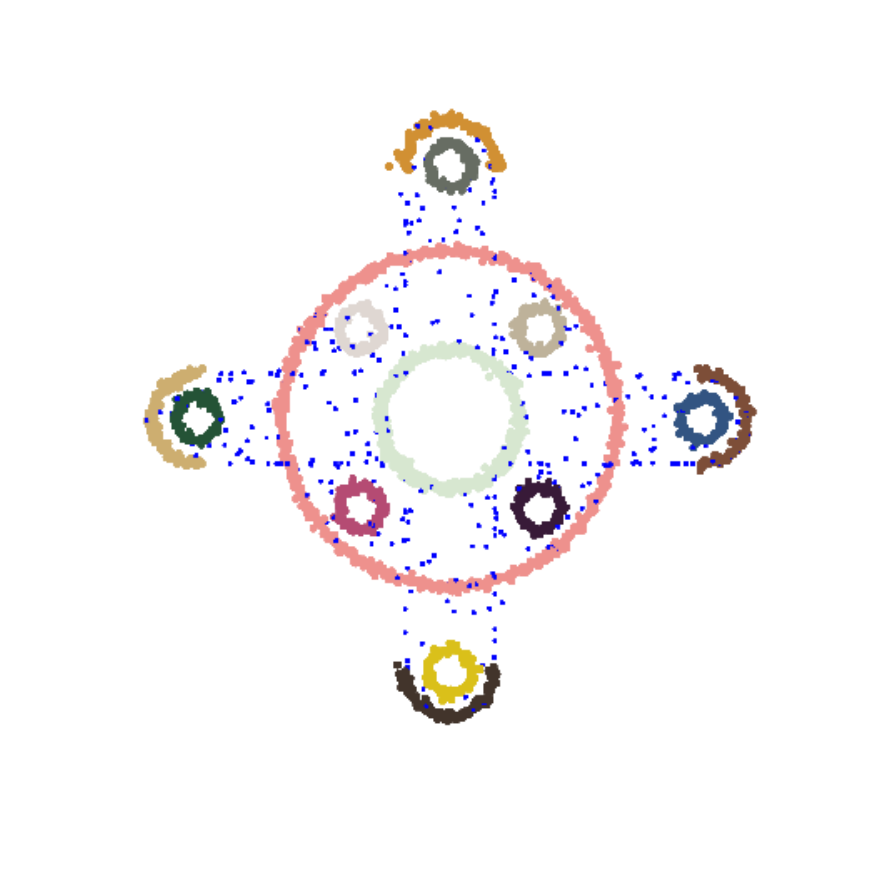}}
            &
            \raisebox{-0.5\height}{\includegraphics[scale=0.380, trim={1.3cm 1.1cm 1cm 1cm}, clip]{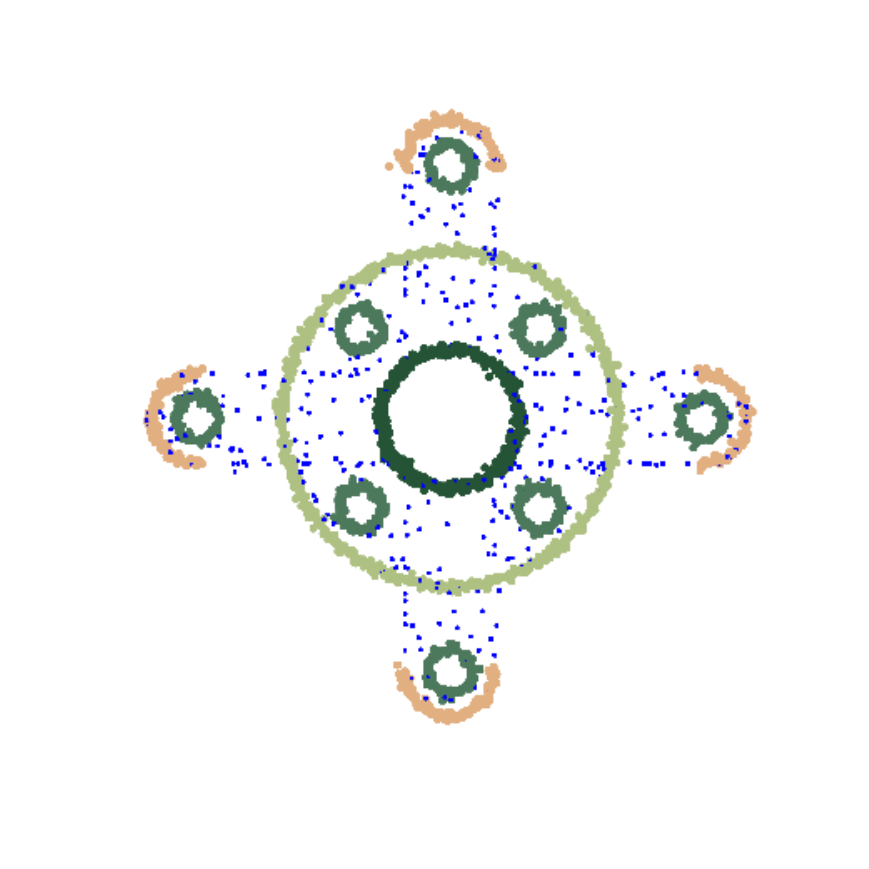}}
            \\
            & & Same plane & Parallel planes & Same cylinder & Same radius
            \\
            \hline
            \rotatebox[origin=c]{90}{(c) $\sigma=0.50$} & \raisebox{-0.5\height}{\includegraphics[scale=0.380, trim={1.3cm 1.1cm 1cm 1cm}, clip]{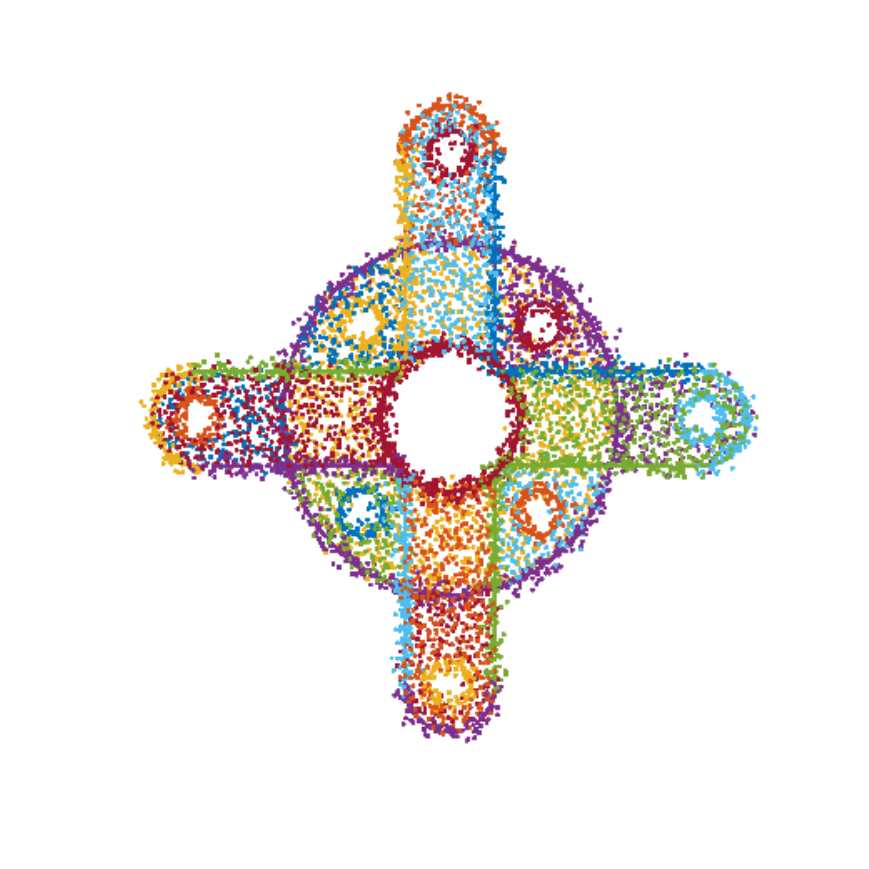}}
            &
            \raisebox{-0.5\height}{\includegraphics[scale=0.380, trim={1.3cm 1.1cm 1cm 0.8cm}, clip]{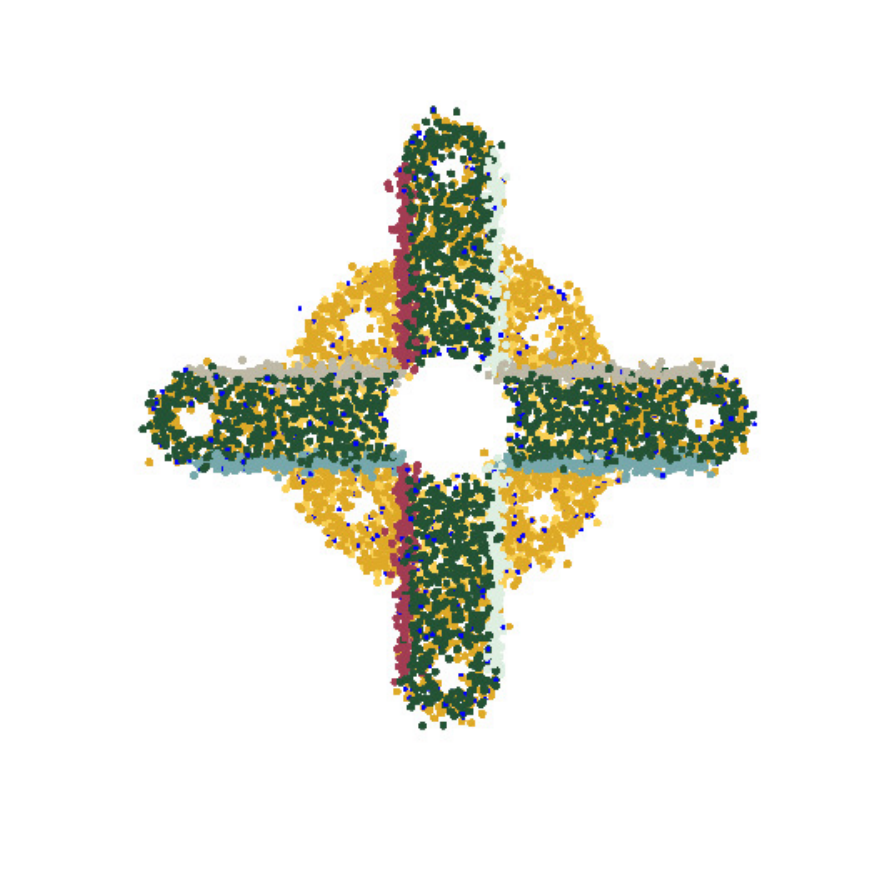}}
            &
            \raisebox{-0.5\height}{\includegraphics[scale=0.380, trim={1.3cm 1.1cm 1cm 1cm}, clip]{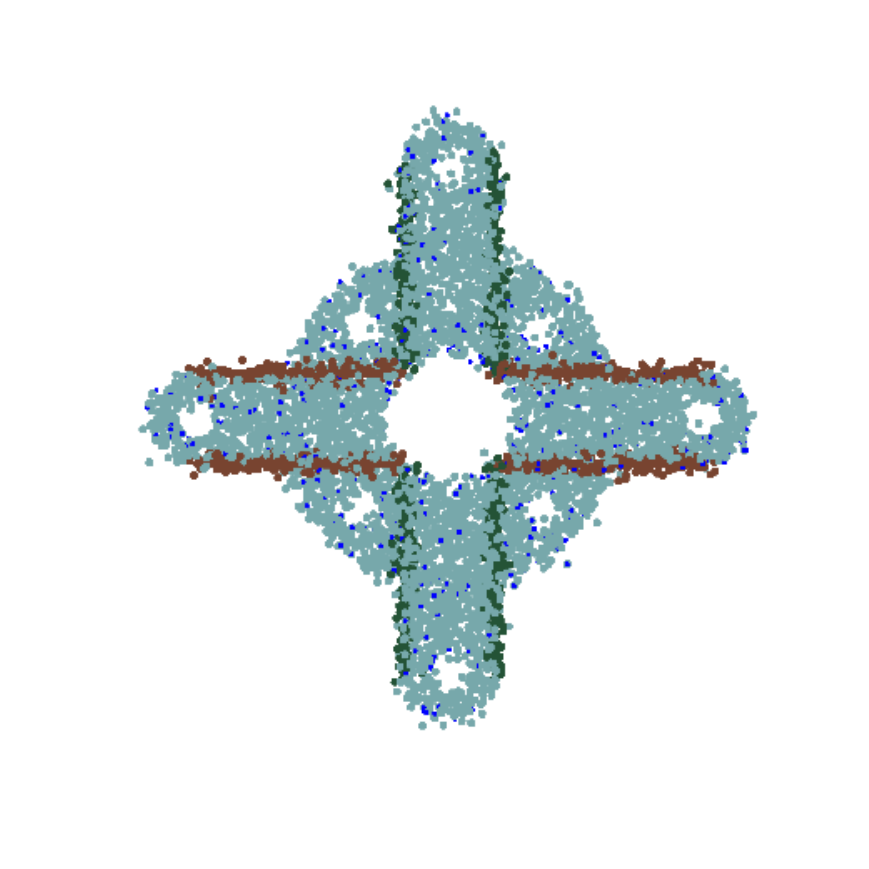}}
            &
            \raisebox{-0.5\height}{\includegraphics[scale=0.380, trim={1.3cm 1.1cm 1cm 1cm}, clip]{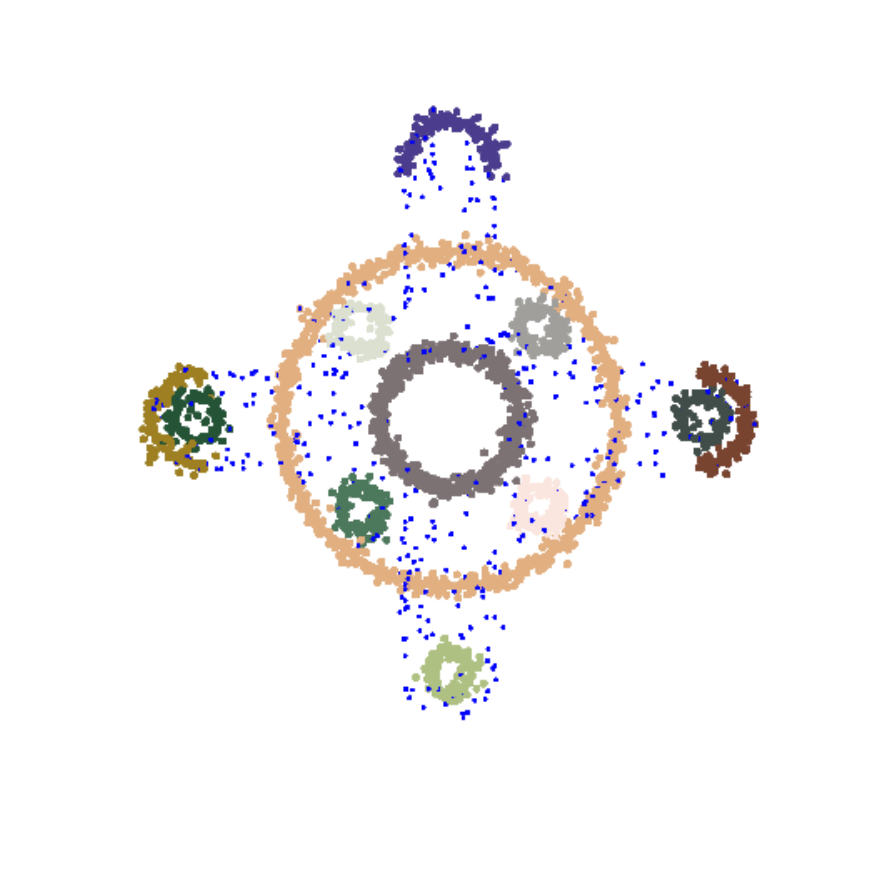}}
            &
            \raisebox{-0.5\height}{\includegraphics[scale=0.380, trim={1.3cm 1.1cm 1cm 1cm}, clip]{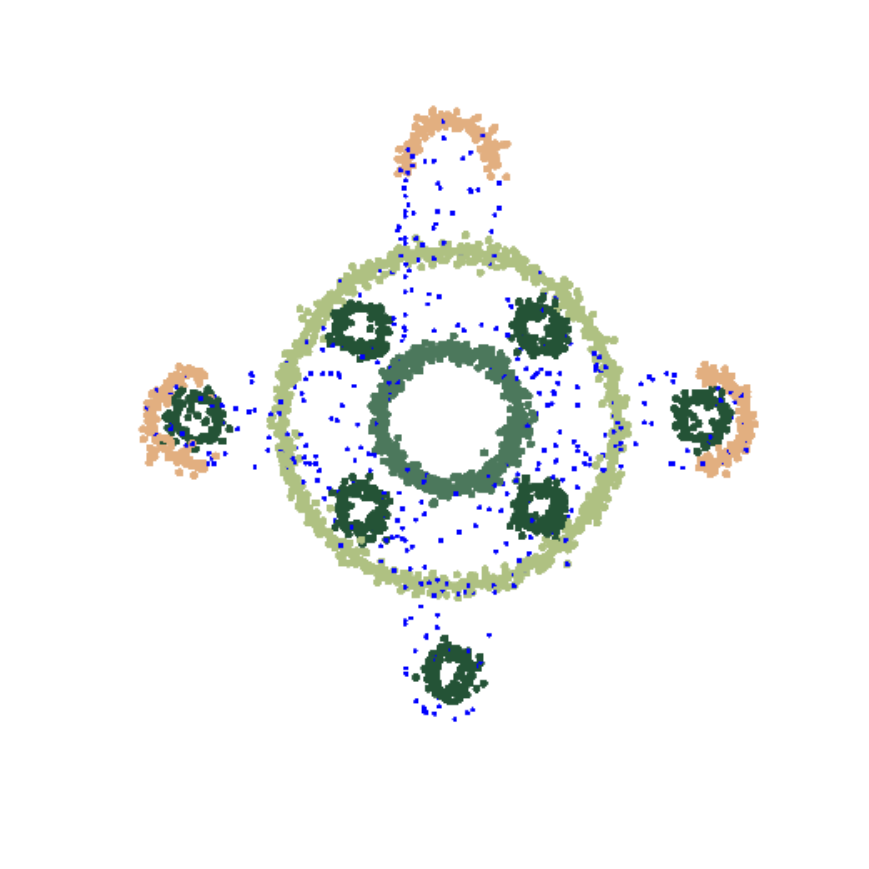}}
            \\
            & & Same plane & Parallel planes & Same cylinder & Same radius
            \\
            \hline
        \end{tabular}
     \end{center}
     \caption{A model from \cite{Koch:2019}: Gaussian noise is applied at three levels of intensity. \label{fig:noise}}
\end{figure}

\subsubsection{Performance over the Fit4CAD dataset (\cite{Fit4CAD})}\label{sec:Fit4CAD}
Fit4CAD is a benchmark created specifically for the evaluation and comparison of methods for fitting simple geometric primitives in point clouds representing CAD objects.
This dataset contains models affected by some point cloud artifacts (e.g., undersampling or missing data).

Table \ref{tab:class_metrics_Fit4CAD} reports the value of five common classification measures employed in \cite{Fit4CAD}: True Positive and Negative Rates (TPR and TNR), Positive and Negative Predicted Values (PPV and NPV), and accuracy (ACC). The values presented in the table are obtained by averaging the single measures over all models. The average accuracy ACC is always above $90\%$; the only cases where it is below $95\%$ correspond to queries involving tori, suggesting a reduced robustness for the corresponding geometric descriptors.
 The average TNR  is always above $98\%$, while the average TPR is always lower than the average TNR. These results shows that the method is great to spot true negatives, while it is more hesitant to cluster together segments. This is particularly an issue for spheres and tori. A possible solution to alleviate this problem would be to allow the use of user-defined thresholds for the dendrogram.
Average PPV and NPV show the degree of correctness of the method in indicating possible positives or negatives. Again, the method has a slightly lower performance for tori. 

\begin{table}[h!]
    \begin{center}
        \caption{Average classification performances for different correlation queries on the Fit4CAD dataset. \label{tab:class_metrics_Fit4CAD}}
        \scriptsize
        \begin{tabular}{|ccccccc|}
            \hline
            \rowcolor{blue!12}Primitive & Query & PPV & TPR & TNR & NPV & ACC \\\hline\Tstrut\Bstrut
            \multirow{2}{*}{Plane}  & Same plane  & 0.998 & 0.980 & 1.000 & 0.999 & 0.999 \\
            & Parallel planes & 0.975 & 0.962 & 0.992 & 0.994 & 0.987 \\[0.1cm]
            \cline{1-7}\Tstrut\Bstrut
            \multirow{4}{*}{Cylinder} & Same cylinder  & 0.995 & 0.980 & 0.999 & 0.999 & 0.998 \\
            & Same radius & 0.974 & 0.970 & 0.995 & 0.996 & 0.991 \\
            & Parallel rotational axes & 0.977 & 0.976 & 0.991 & 0.980 & 0.984 \\
            & Same rotational axis & 0.992 & 0.985 & 0.995 & 0.998 & 0.995 \\[0.1cm]
            \cline{1-7}\Tstrut\Bstrut
            \multirow{5}{*}{Cone} & Same cone  & 1.000 & 0.944 & 1.000 & 0.994 & 0.994 \\
            & Same aperture & 1.000 & 0.922 & 1.000 & 0.981 & 0.985 \\
            & Same apex & 1.000 & 0.944 & 1.000 & 0.994 & 0.994 \\
            & Parallel rotational axes & 1.000 & 0.944 & 1.000 & 0.986 & 0.988 \\
            & Same rotational axis & 1.000 & 0.833 & 1.000 & 0.942 & 0.952 \\
            \cline{1-7}\Tstrut\Bstrut
            \multirow{3}{*}{Sphere} & Same sphere & 1.000 & 0.667 & 1.000 & 0.976 & 0.976 \\
            & Same radius & 1.000 & 0.667 & 1.000 & 0.976 & 0.976 \\
            & Same center & 1.000 & 0.667 & 1.000 & 0.976 & 0.976 \\[0.1cm]
            \cline{1-7}\Tstrut\Bstrut
            \multirow{5}{*}{Torus} & Same torus & 1.000 & 0.583 & 1.000 & 0.958 & 0.958 \\
            & Same radii & 1.000 & 0.583 & 1.000 & 0.958 & 0.958 \\
            & Same center & 0.889 & 0.600 & 0.985 & 0.955 & 0.943 \\
            & Parallel rotational axes & 0.875 & 0.583 & 0.993 & 0.896 & 0.901 \\
            & Same rotational axis & 0.875 & 0.583 & 0.993 & 0.896 & 0.901 \\
            \hline
        \end{tabular}
    \end{center}
\end{table}

\subsection{Comparison with the method (\cite{Raffo:2020})}\label{sec:comparison_Raffo}
We compared our method with \cite{Raffo:2020} because it adopts a  pipeline similar to ours and, just as importantly, its implementation is available online.

The method \cite{Raffo:2020} consists of a combination of approximate implicitization -- which reduces to a least squares minimization in its discrete formulation \cite{Barrowclough:2012} -- and hierarchical clustering. Unlike our method, it uses the coefficients of implicit representations as geometric descriptor; although computationally efficient, this choice results in problems of instability when point cloud artifacts are present, as pointed out by the authors. Moreover, \cite{Raffo:2020} only checks which segments lie on the same surface, but it does not identify the segment type (e.g., cylinder vs. cone), nor  to formulate more complex queries.  

In Table \ref{tab:comparison}, this comparison is drawn with respect to the five classification measures from the previous section. We can notice a generally better performance of our approach, especially in the case of noisy or perturbed data; indeed, polynomial estimations used in \cite{Raffo:2020} have been proved to be particularly sensitive to data perturbation, while the HT paradigm is popularly known for its robustness. On the other hand, \cite{Raffo:2020} has a lower computational complexity, making it preferable for an initial inspection when the input is clean or when the user is not interested in other correlation queries. 

\begin{table}[h]
 \centering
\caption{Classification performance: comparison between our method and the approach proposed in \cite{Raffo:2020}.
\label{tab:comparison}}
 \scriptsize
\begin{tabular}{|c|cccccc|}
\hline
 \rowcolor{blue!12}Model & Method & PPV & TPR & TNR & NPV & ACC \\\hline
 Fig \ref{fig:tubo} & \begin{tabular}{c}
 Ours
 \\
 \cite{Raffo:2020}
 \end{tabular}
 &
 \begin{tabular}{c}
 $1.000$
 \\
 $0.611$
 \end{tabular}
 & 
 \begin{tabular}{c}
 $1.000$
 \\
 $1.000$
 \end{tabular}
 & 
 \begin{tabular}{c}
 $1.000$
 \\
 $0.948$
 \end{tabular}
 & 
 \begin{tabular}{c}
 $1.000$
 \\
 $1.000$
 \end{tabular}
 & 
 \begin{tabular}{c}
 $1.000$
 \\
 $0.951$
 \end{tabular}
 \\\hline
 Fig \ref{fig:topolino_pokeball}(a) & \begin{tabular}{c}
 Ours
 \\
 \cite{Raffo:2020}
 \end{tabular}
 &
 \begin{tabular}{c}
 $1.000$
 \\
 $0.739$
 \end{tabular}
 & 
 \begin{tabular}{c}
 $1.000$
 \\
 $0.957$
 \end{tabular}
 & 
 \begin{tabular}{c}
 $1.000$
 \\
 $0.937$
 \end{tabular}
 & 
 \begin{tabular}{c}
 $1.000$
 \\
 $0.996$
 \end{tabular}
 & 
 \begin{tabular}{c}
 $1.000$
 \\
 $0.936$
 \end{tabular}
 \\\hline
 Fig \ref{fig:topolino_pokeball}(b)& \begin{tabular}{c}
 Ours
 \\
 \cite{Raffo:2020}
 \end{tabular}
 &
 \begin{tabular}{c}
 $0.875$
 \\
 $0.875$
 \end{tabular}
 & 
 \begin{tabular}{c}
 $1.000$
 \\
 $1.000$
 \end{tabular}
 & 
 \begin{tabular}{c}
 $0.964$
 \\
 $0.964$
 \end{tabular}
 & 
 \begin{tabular}{c}
 $1.000$
 \\
 $1.000$
 \end{tabular}
 & 
 \begin{tabular}{c}
 $0.969$
 \\
 $0.969$
 \end{tabular}
 \\\hline
 Fig \ref{fig:noise}(a) & \begin{tabular}{c}
 Ours
 \\
 \cite{Raffo:2020}
 \end{tabular}
 &
 \begin{tabular}{c}
 $1.000$
 \\
 $0.079$
 \end{tabular}
 & 
 \begin{tabular}{c}
 $1.000$
 \\
 $0.988$
 \end{tabular}
 & 
 \begin{tabular}{c}
 $1.000$
 \\
 $0.094$
 \end{tabular}
 & 
 \begin{tabular}{c}
 $1.000$
 \\
 $0.952$
 \end{tabular}
 & 
 \begin{tabular}{c}
 $1.000$
 \\
 $0.160$
 \end{tabular}
 \\\hline
 Fig \ref{fig:noise}(b) & \begin{tabular}{c}
 Ours
 \\
 \cite{Raffo:2020}
 \end{tabular}
 &
 \begin{tabular}{c}
 $1.000$
 \\
 $0.076$
 \end{tabular}
 & 
 \begin{tabular}{c}
 $1.000$
 \\
 $1.000$
 \end{tabular}
 & 
 \begin{tabular}{c}
 $1.000$
 \\
 $0.000$
 \end{tabular}
 & 
 \begin{tabular}{c}
 $1.000$
 \\
 $0.000$
 \end{tabular}
 & 
 \begin{tabular}{c}
 $1.000$
 \\
 $0.076$
 \end{tabular}
 \\\hline
 Fig \ref{fig:noise}(c) & \begin{tabular}{c}
 Ours
 \\
 \cite{Raffo:2020}
 \end{tabular}
 &
 \begin{tabular}{c}
 $0.905$
 \\
 $0.076$
 \end{tabular}
 & 
 \begin{tabular}{c}
 $0.905$
 \\
 $1.000$
 \end{tabular}
 & 
 \begin{tabular}{c}
 $1.000$
 \\
 $0.000$
 \end{tabular}
 & 
 \begin{tabular}{c}
 $0.998$
 \\
 $0.000$
 \end{tabular}
 & 
 \begin{tabular}{c}
 $0.998$
 \\
 $0.076$
 \end{tabular}
 \\\hline
 Fit4CAD & \begin{tabular}{c}
 Ours
 \\
 \cite{Raffo:2020}
 \end{tabular}
 &
 \begin{tabular}{c}
 $0.995$
 \\
 $0.791$
 \end{tabular}
 & 
 \begin{tabular}{c}
 $0.997$
 \\
 $0.992$
 \end{tabular}
 & 
 \begin{tabular}{c}
 $0.999$
 \\
 $0.946$
 \end{tabular}
 & 
 \begin{tabular}{c}
 $0.999$
 \\
 $0.998$
 \end{tabular}
 & 
 \begin{tabular}{c}
 $0.999$
 \\
 $0.950$
 \end{tabular}
 \\\hline

\end{tabular}
\end{table}

\subsection{Tests on point clouds segmented by different methods}\label{sec:input_methods}
To evaluate the performance of our method, we applied it to point clouds segmented with different techniques. The first set of tests includes segments obtained with the learning technique presented in \cite{ParseNet}. The second set of tests considers segments obtained by applying RANSAC on industrial scans.

\subsubsection{Tests on point clouds segmented by a learning approach}
\begin{figure}[t!]
\begin{center}
        \footnotesize
        \begin{tabular}{|P{2.5cm}|P{2.5cm}|P{2.5cm}|P{2.5cm}|}
            \hline
            \rowcolor{teal!25}\makecell{Segmentation} &  \multicolumn{3}{|c|}{Primitives} \\\hline
            \includegraphics[scale=0.3, trim={1.5cm 1.0cm 0.75cm 0.5cm}, clip]{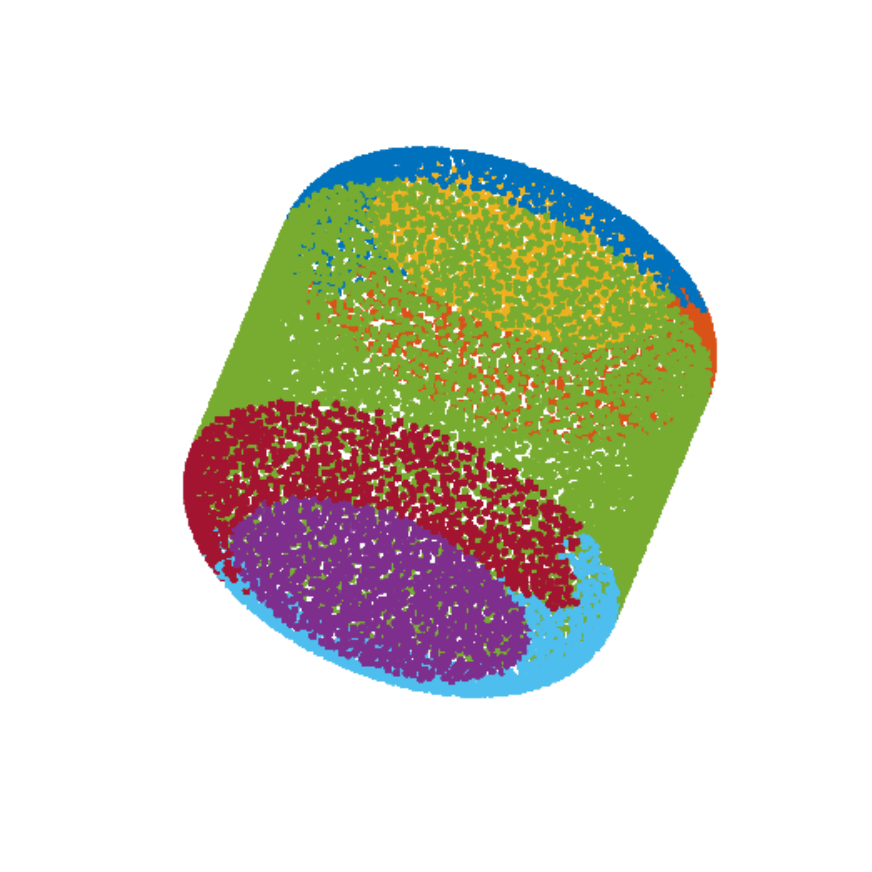}
             &
             \includegraphics[scale=0.3, trim={1.5cm 1.0cm 0.75cm 0.5cm}, clip]{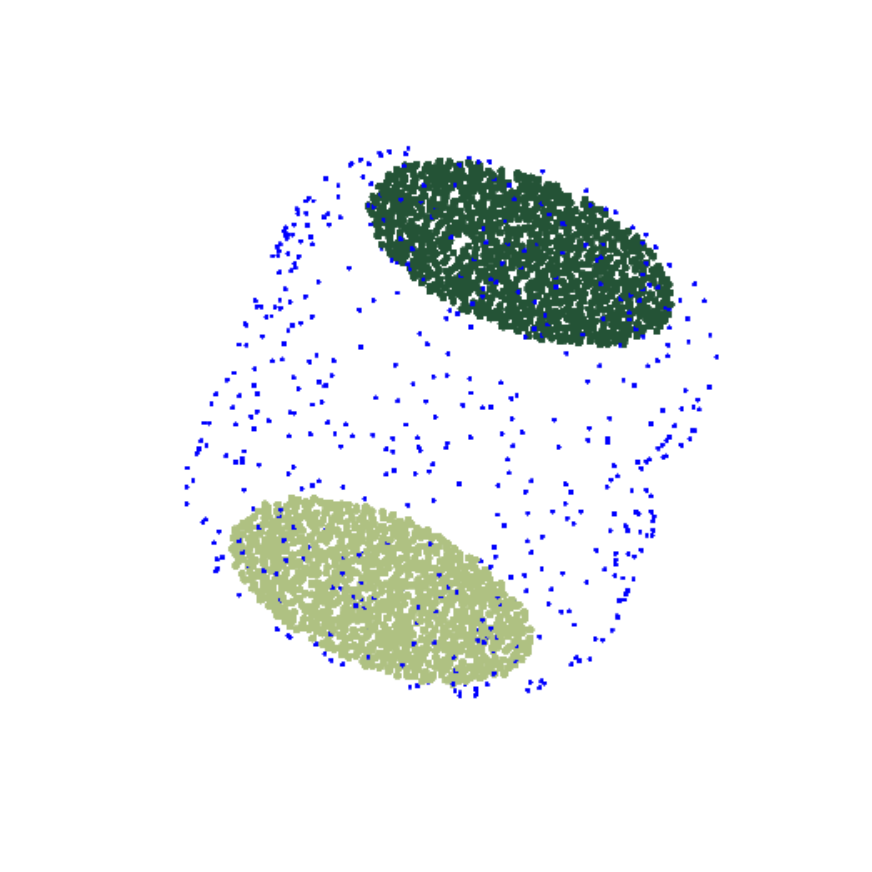}
             &
             \includegraphics[scale=0.3, trim={1.5cm 1.0cm 0.75cm 0.5cm}, clip]{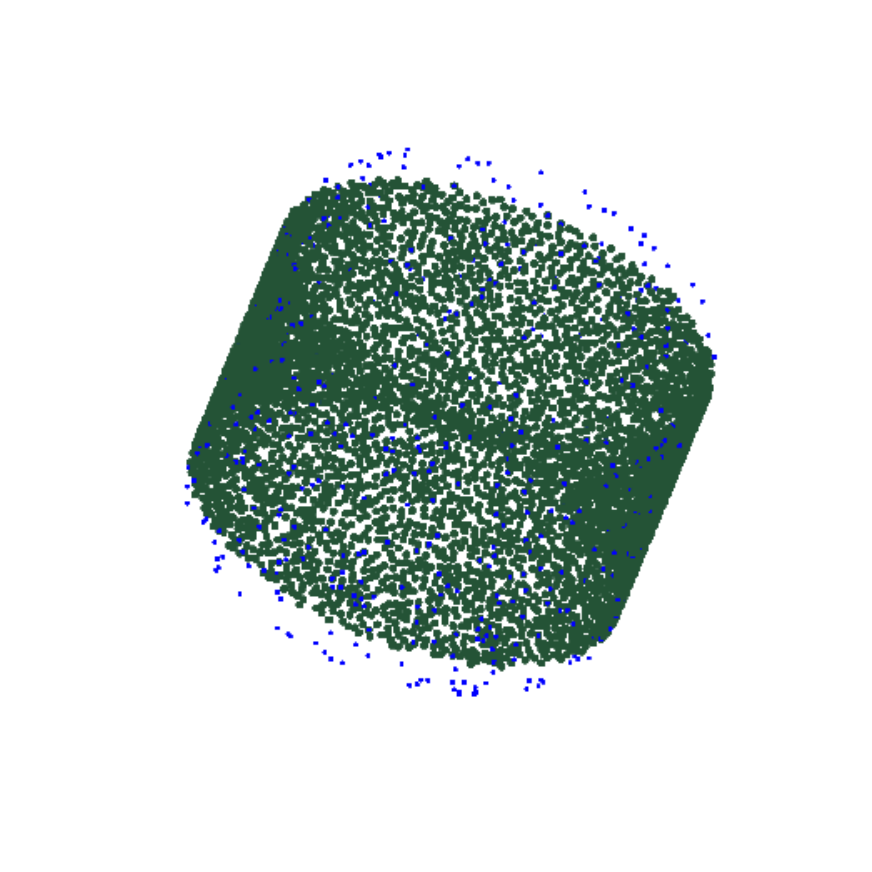}
             &
             \includegraphics[scale=0.3, trim={1.5cm 1.0cm 0.75cm 0.5cm}, clip]{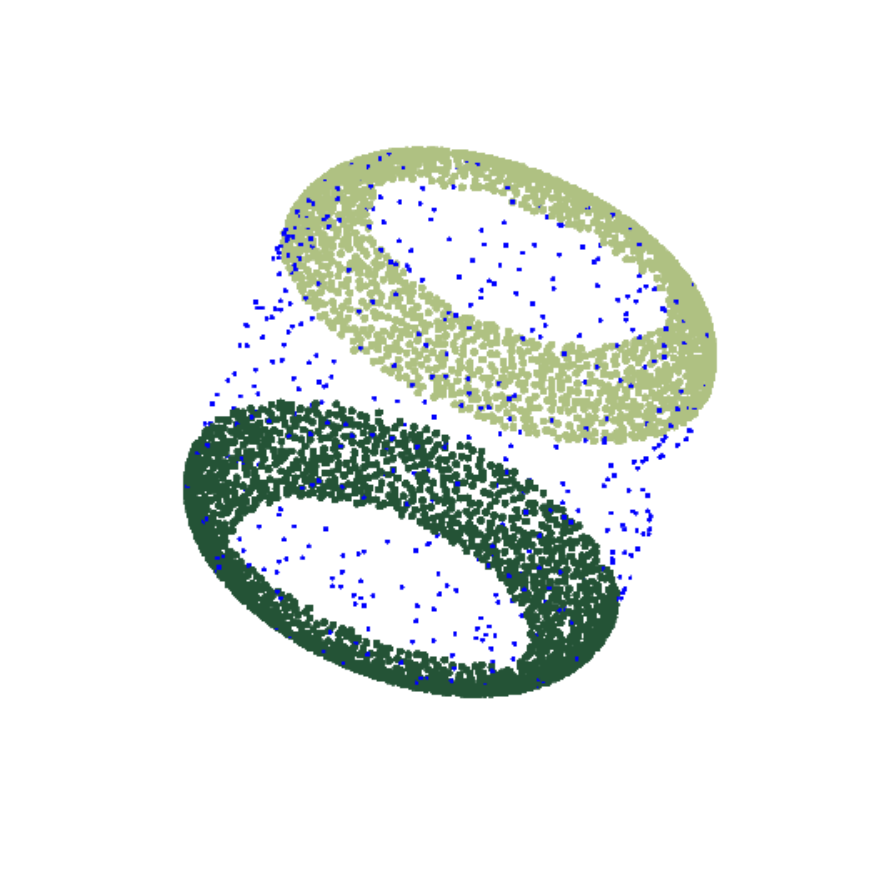}
             \\
             (a) & Planes & Cylinders & Tori \\
             \hline
             \includegraphics[scale=0.3, trim={1.5cm 2.5cm 2.5cm 1.0cm}, clip]{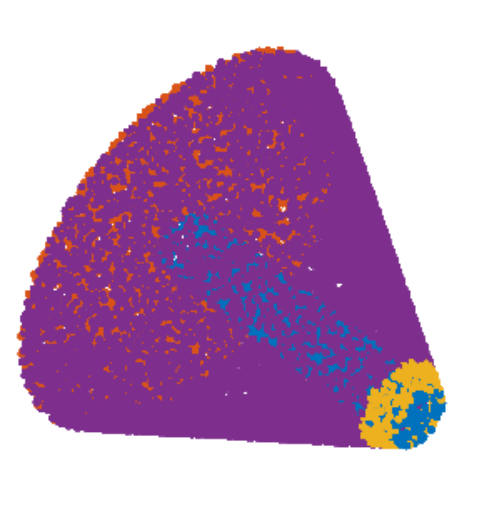}
             &
             \includegraphics[scale=0.3, trim={1.5cm 2.5cm 2.5cm 1.0cm}, clip]{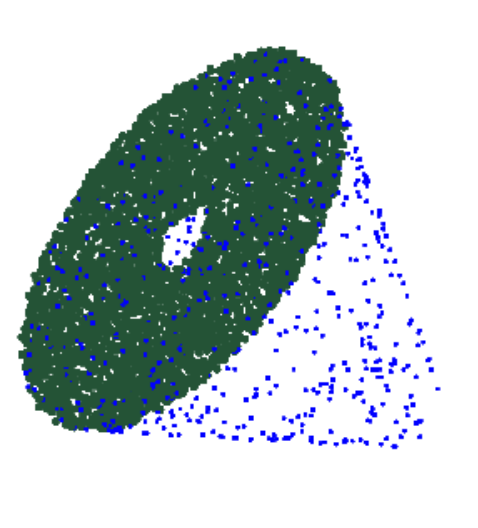}
             &
             \includegraphics[scale=0.3, trim={1.5cm 2.5cm 2.5cm 1.0cm}, clip]{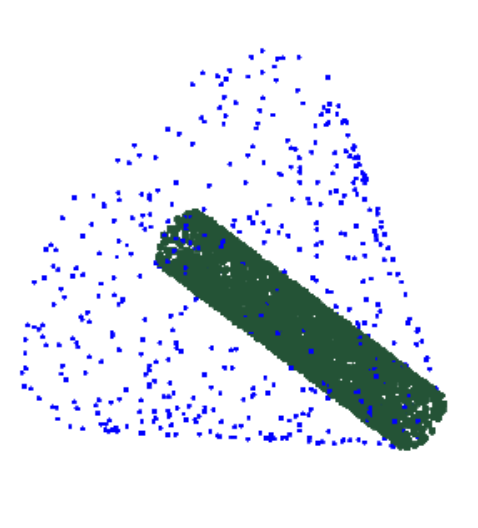}
             &
             \includegraphics[scale=0.3, trim={1.5cm 2.5cm 2.5cm 1.0cm}, clip]{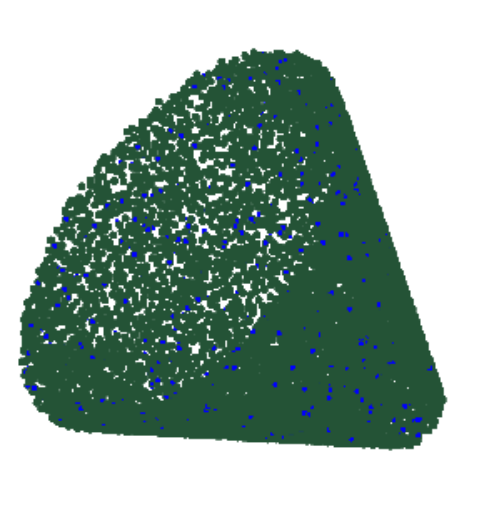}
             \\
             (b) & Planes & Cylinders & Cones \\
             \hline
             \includegraphics[scale=0.3, trim={0.5cm 1.5cm 0.75cm 0.5cm}, clip]{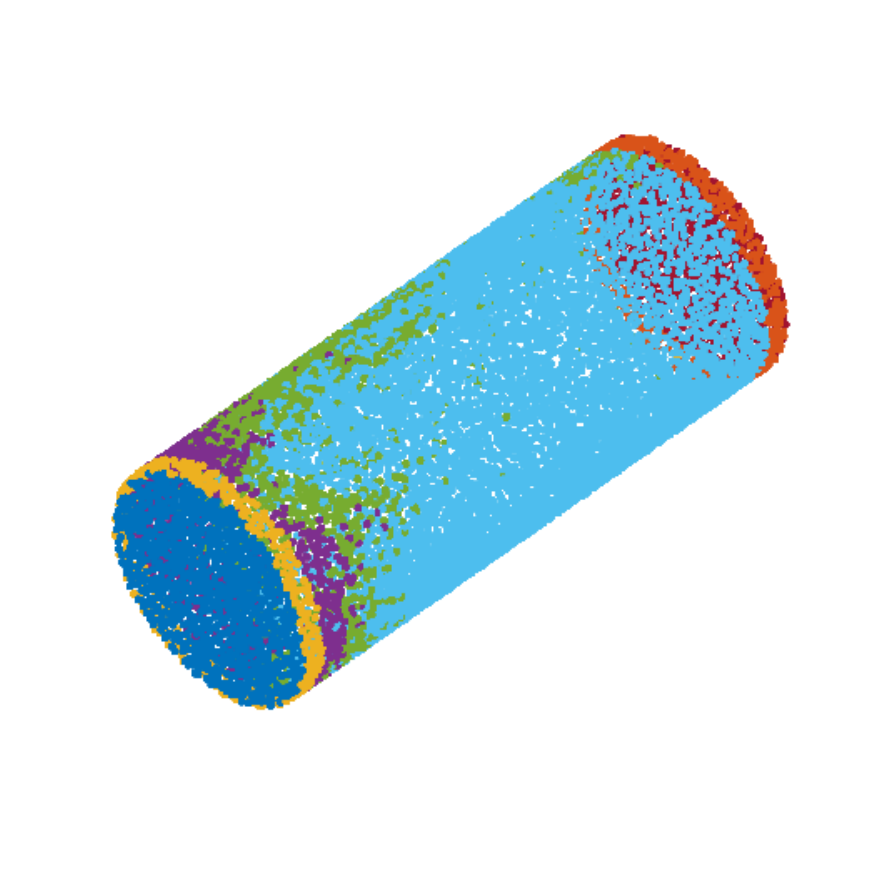}
             &
             \includegraphics[scale=0.3, trim={0.5cm 1.5cm 0.75cm 0.5cm}, clip]{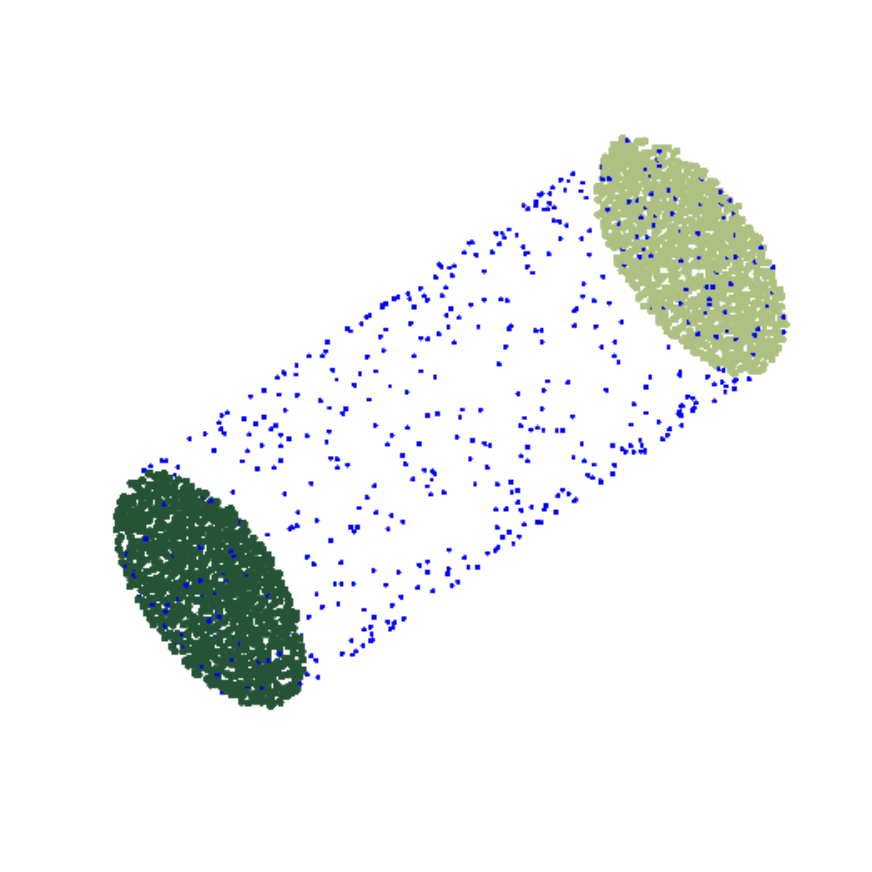}
             &
             \includegraphics[scale=0.3, trim={0.5cm 1.5cm 0.75cm 0.5cm}, clip]{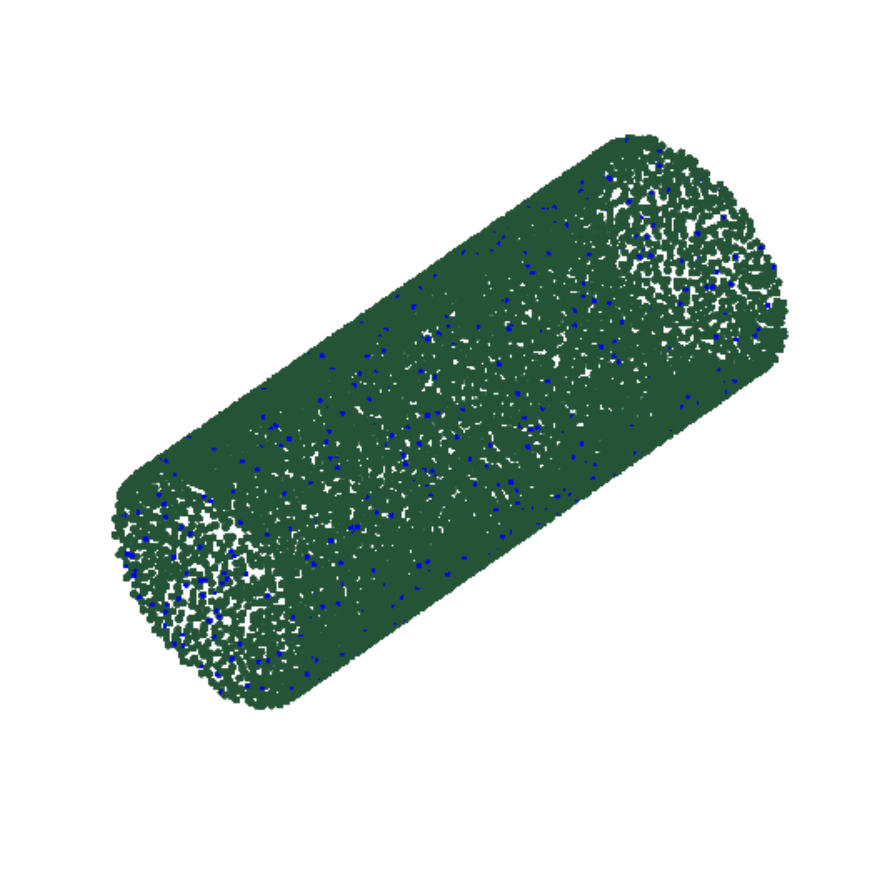}
             &
             \\
             (c) & Planes & Cylinders &  \\
             \hline
             \includegraphics[scale=0.3, trim={1.5cm 1.0cm 0.75cm 0.5cm}, clip]{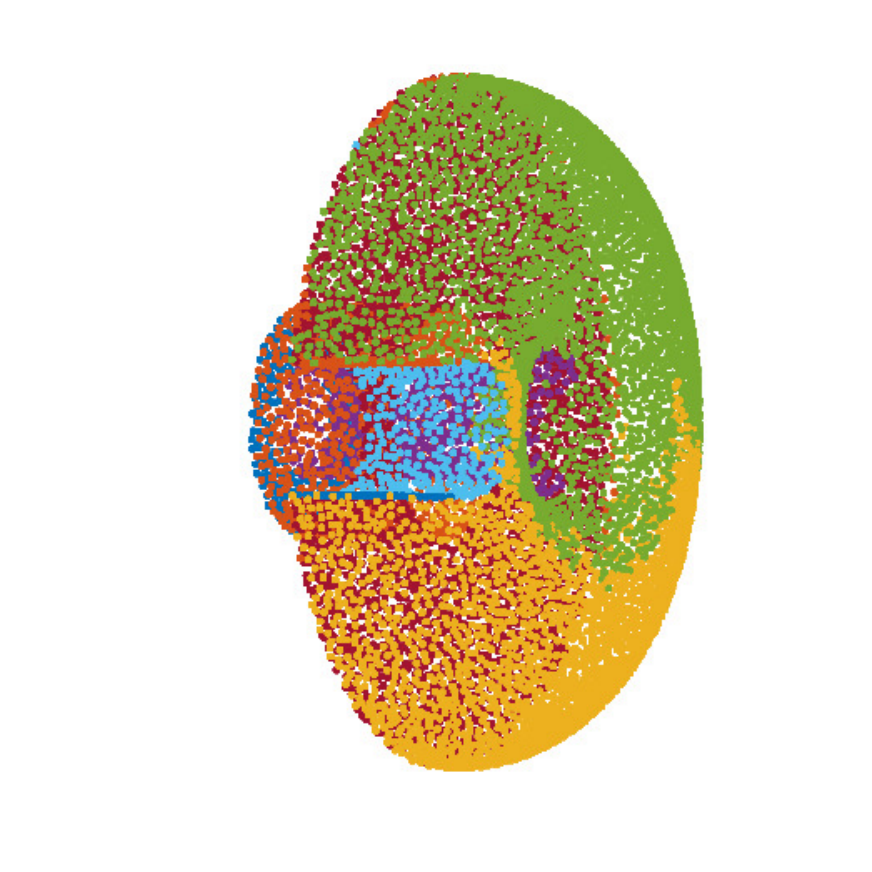}
             &
             \includegraphics[scale=0.3, trim={1.5cm 1.0cm 0.75cm 0.5cm}, clip]{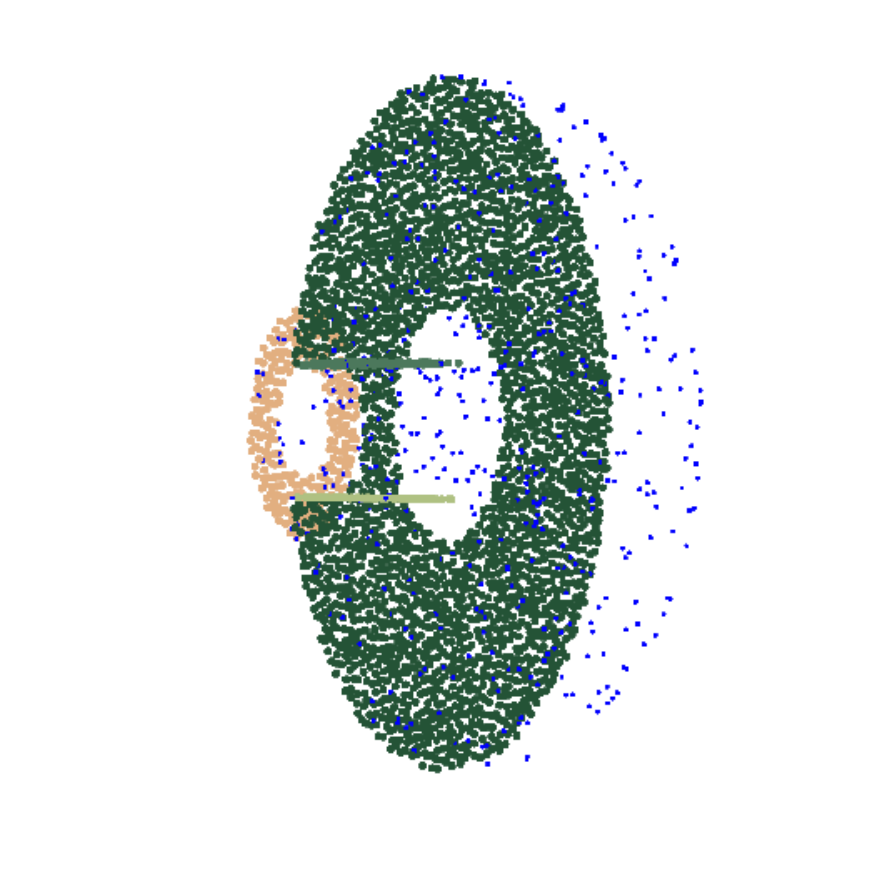}
             &
             \includegraphics[scale=0.3, trim={1.5cm 1.0cm 0.75cm 0.5cm}, clip]{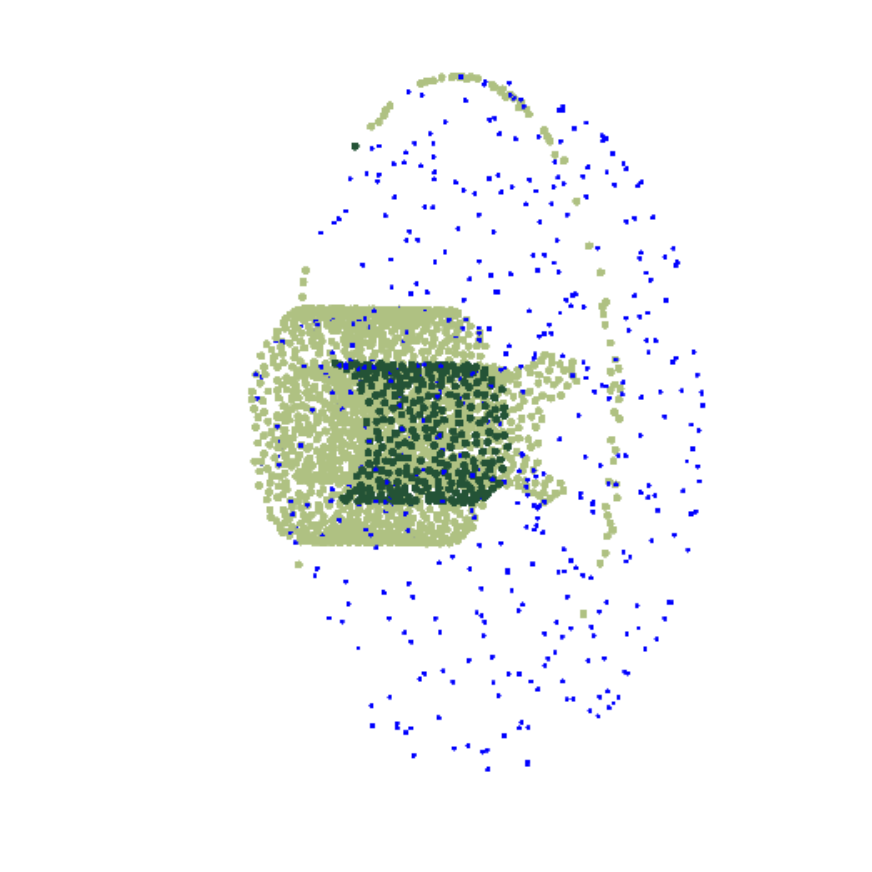}
             &
             \includegraphics[scale=0.3, trim={1.5cm 1.0cm 0.75cm 0.5cm}, clip]{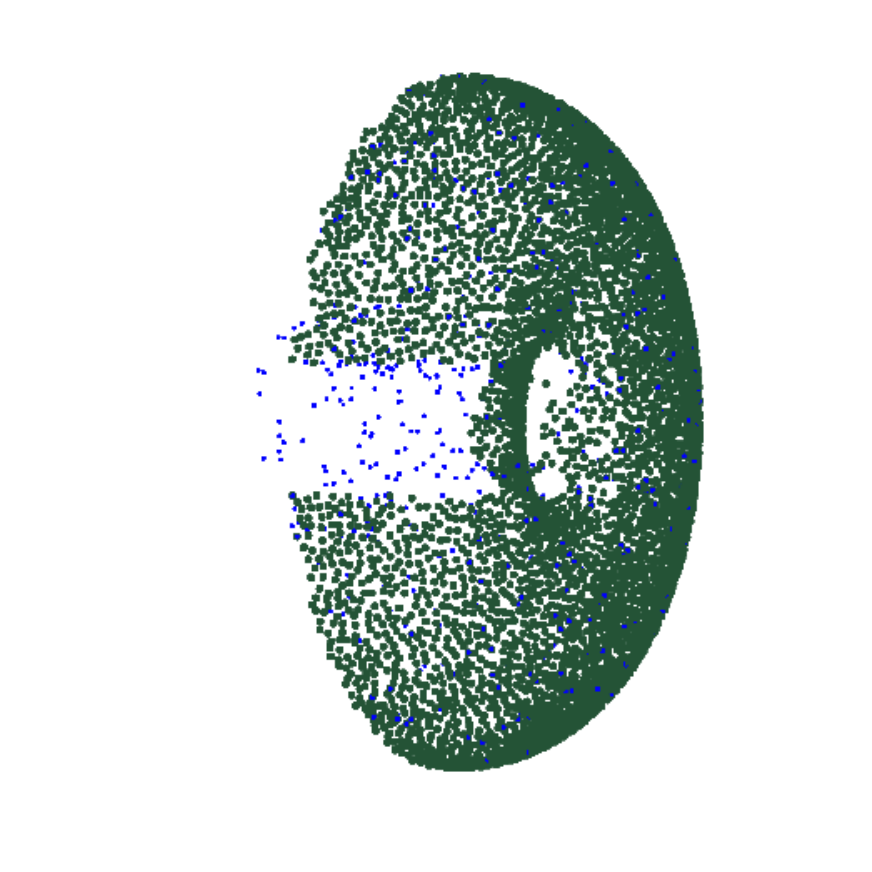}
             \\
             (d) & Planes & Cylinders & Tori \\
             \hline
             \includegraphics[scale=0.3, trim={1.5cm 1.0cm 0.75cm 0.5cm}, clip]{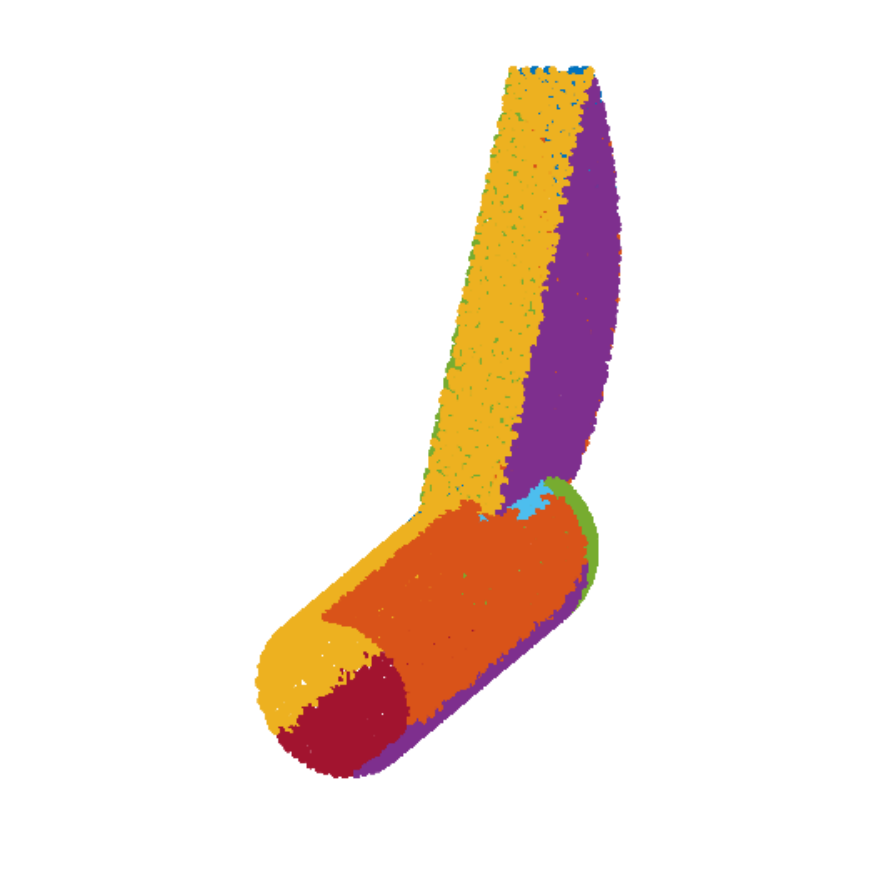}
             &
             \includegraphics[scale=0.3, trim={1.5cm 1.0cm 0.75cm 0.5cm}, clip]{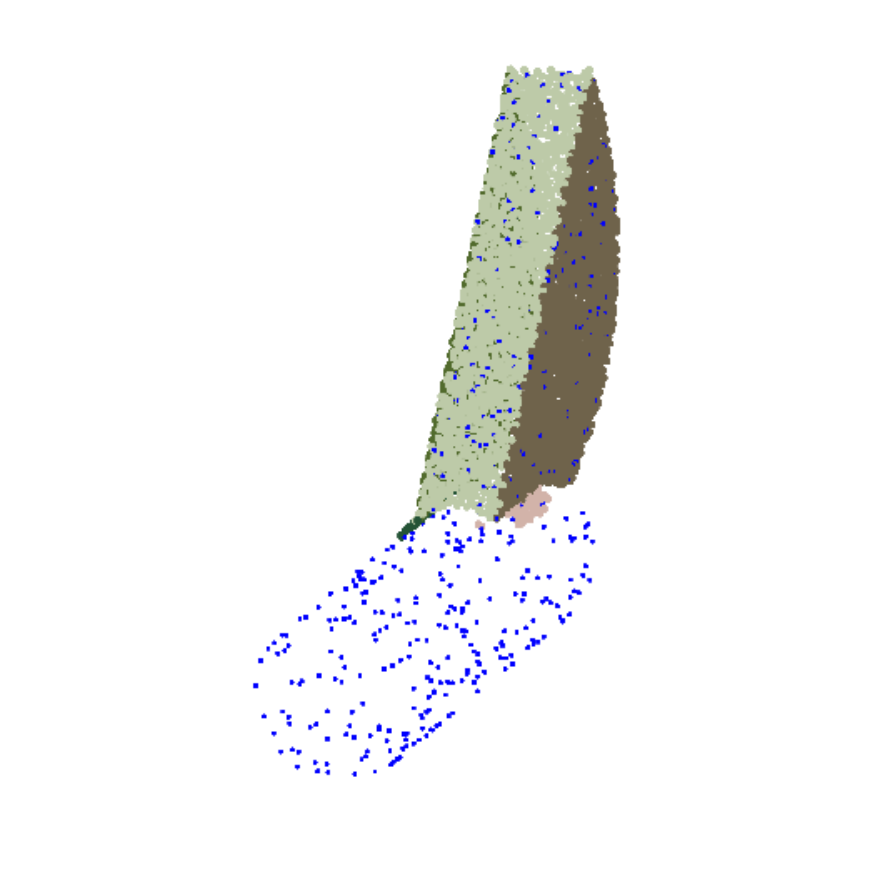}
             &
             \includegraphics[scale=0.3, trim={1.5cm 1.0cm 0.75cm 0.5cm}, clip]{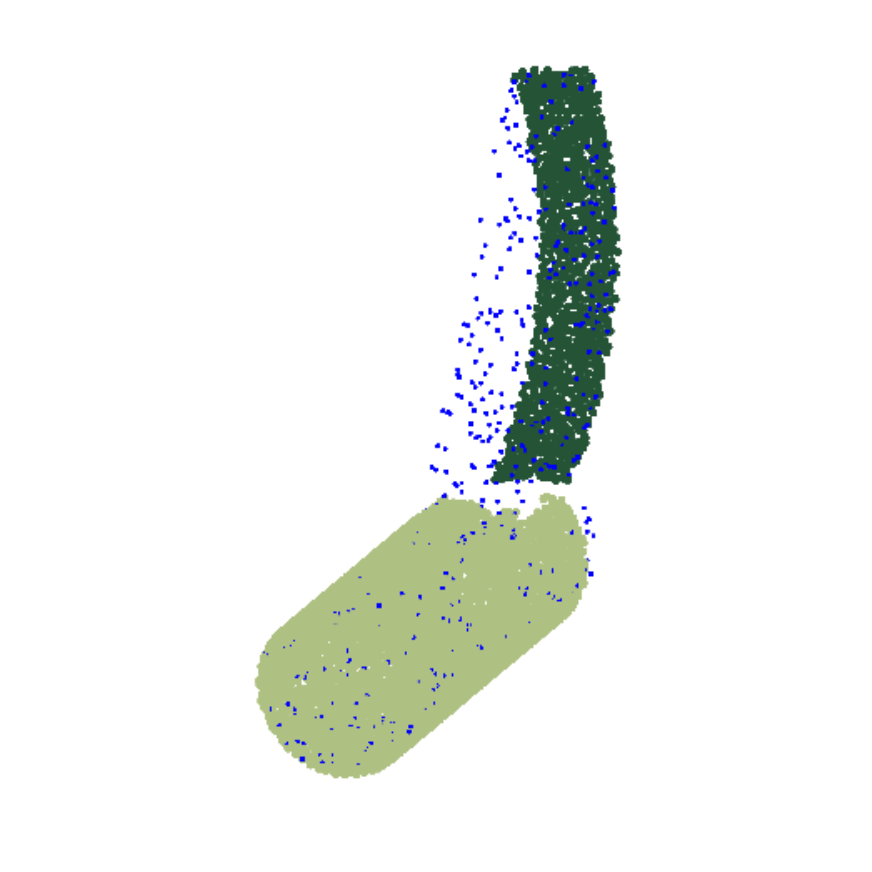}
             &
             \includegraphics[scale=0.3, trim={1.5cm 1.0cm 0.75cm 0.5cm}, clip]{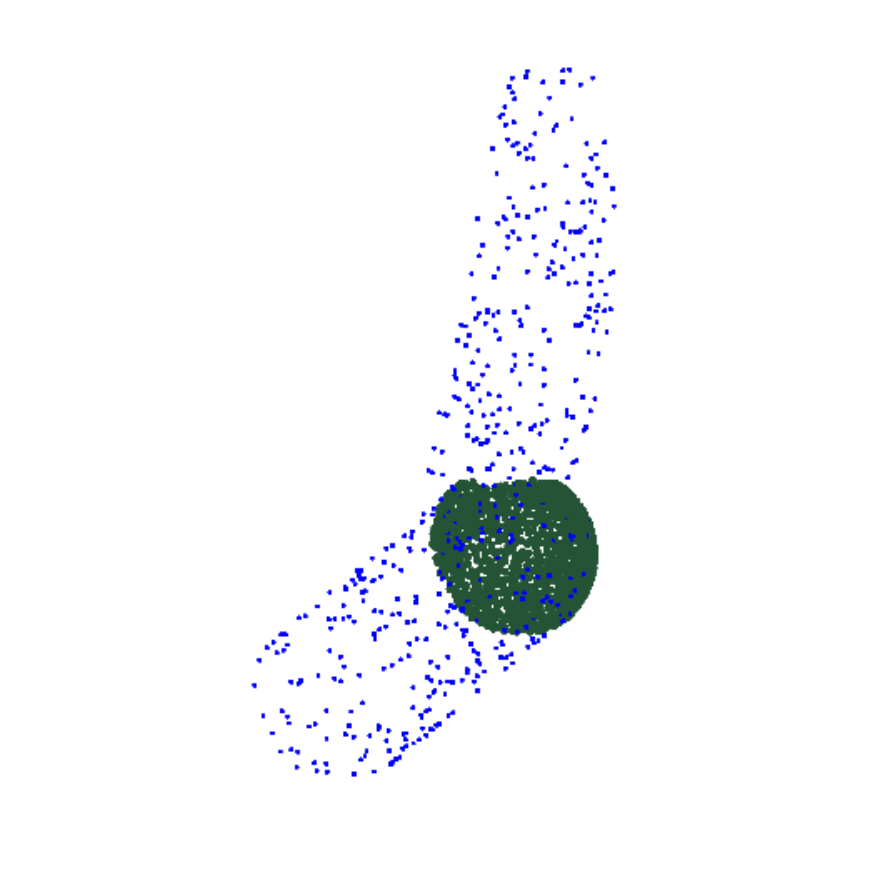}
             \\
             (e) & Planes & Cylinders & Spheres \\
             \hline
            \end{tabular}
    \end{center}
    \caption{Examples of segmentations from \cite{ParseNet}: the original segmentations (first column) are post-processed by our method to overcome the problem of oversegmentation. 
    \label{fig:parsenet}}
\end{figure}

We tested our method on some of the segmented point clouds provided by a learning approach \cite{ParseNet} that are composed of $10,000$ points. In this case, our technique is useful as post processing to overcome the problem of oversegmentation and may confirm or modify the classification of the primitive. Indeed, as shown in Figure \ref{fig:parsenet}, the combination of the geometric descriptors identification and clustering procedure permits to aggregate segments belonging to the same primitive. Specifically, Figure \ref{fig:parsenet}(a) presents a point cloud divided in 7 segments, where four of them can be grouped two by two since they belong to the same torus. A similar grouping of segments belonging to the same torus is provided in Figure \ref{fig:parsenet}(d). Figure \ref{fig:parsenet}(b) shows a model divided in 4 segments and composed of a plane, a cylinder and a cone split in two parts, of which one very small. Our method groups these two pieces since they belong to the same primitive. In Figure \ref{fig:parsenet}(c) the point cloud is divided in 7 segments, where five of them belong to the same cylinder. Finally, in Figure \ref{fig:parsenet}(e) a point cloud made of 12 segments is reduced to 9 by grouping  four pieces of the same cylinder. The mean of the MFE over all segments is: (a)  $0.0097$, (b) $0.0102$, (c) $0.0036$, (d) $0.0048$, (e)  $0.0061$.

\subsubsection{Tests on industrial scans segmented by the RANSAC technique}
\begin{figure}[h!]
\footnotesize
    \begin{center}
        \begin{tabular}{|c|c|c|c|}
            \hline
            \rowcolor{teal!25}\multicolumn{1}{|c|}{Point cloud and planes} &  Cylinders &  Cones & Spheres \\\hline
            \includegraphics[scale=0.45, trim={1cm 2cm 1cm 1.5cm}, clip]{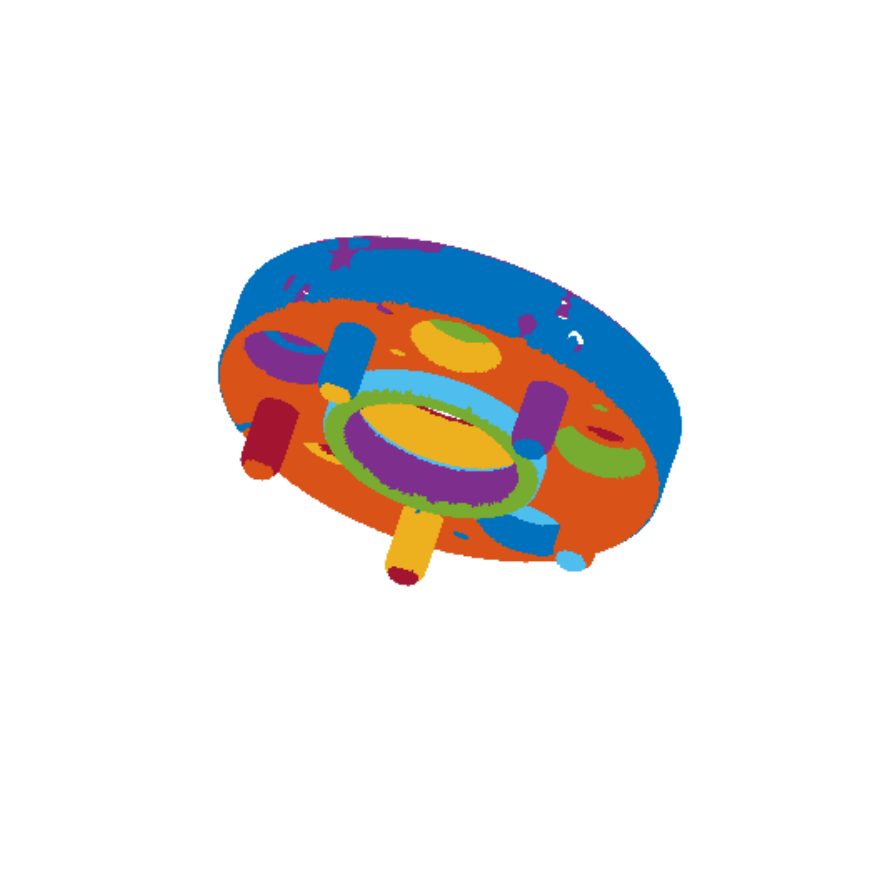}
            &
            \includegraphics[scale=0.375, trim={1cm 1.5cm 1cm 1.2cm}, clip]{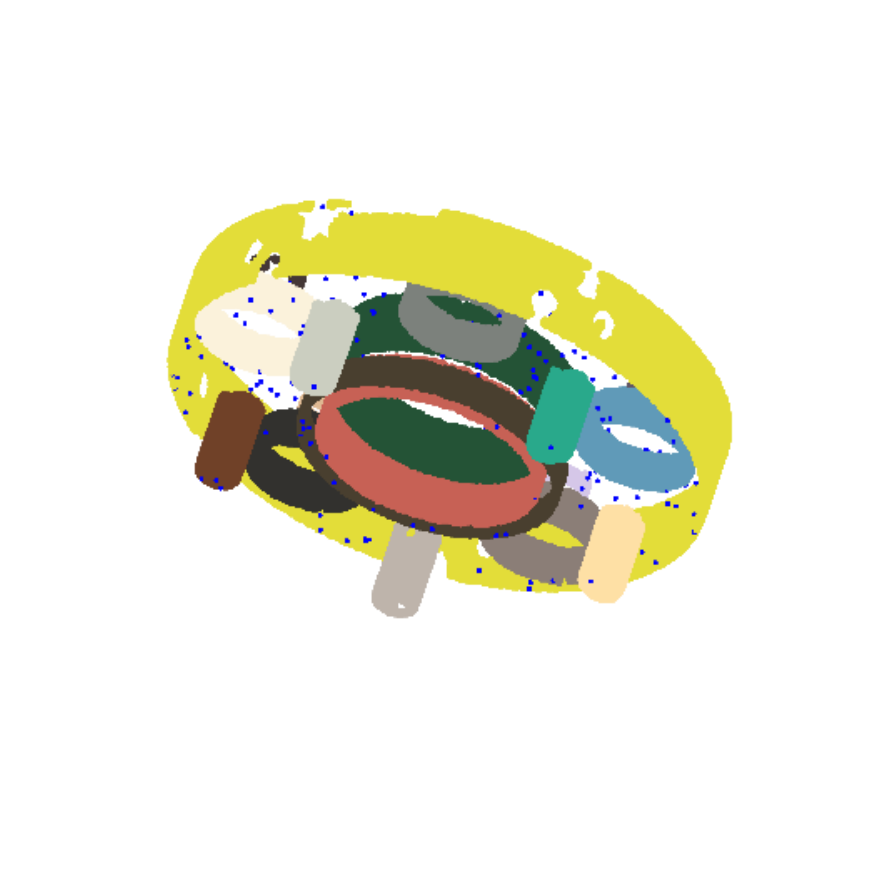}
            &
            \includegraphics[scale=0.375, trim={1cm 1.5cm 1cm 1.25cm}, clip]{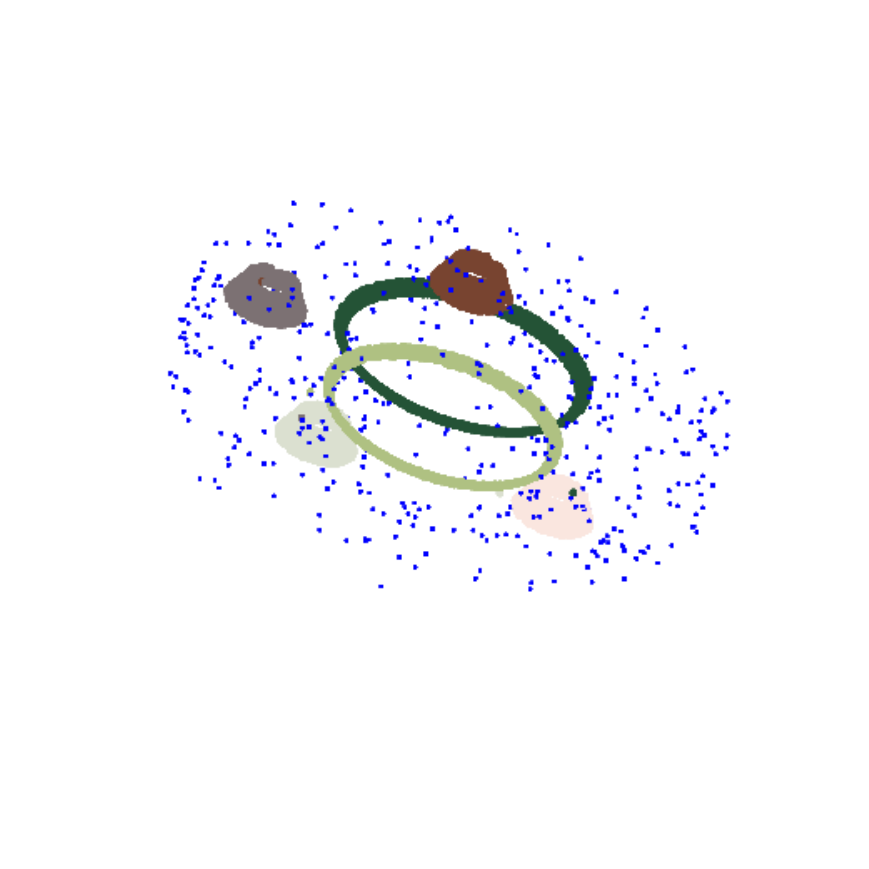}
            &
            \includegraphics[scale=0.375, trim={1cm 1.5cm 1cm 1.25cm}, clip]{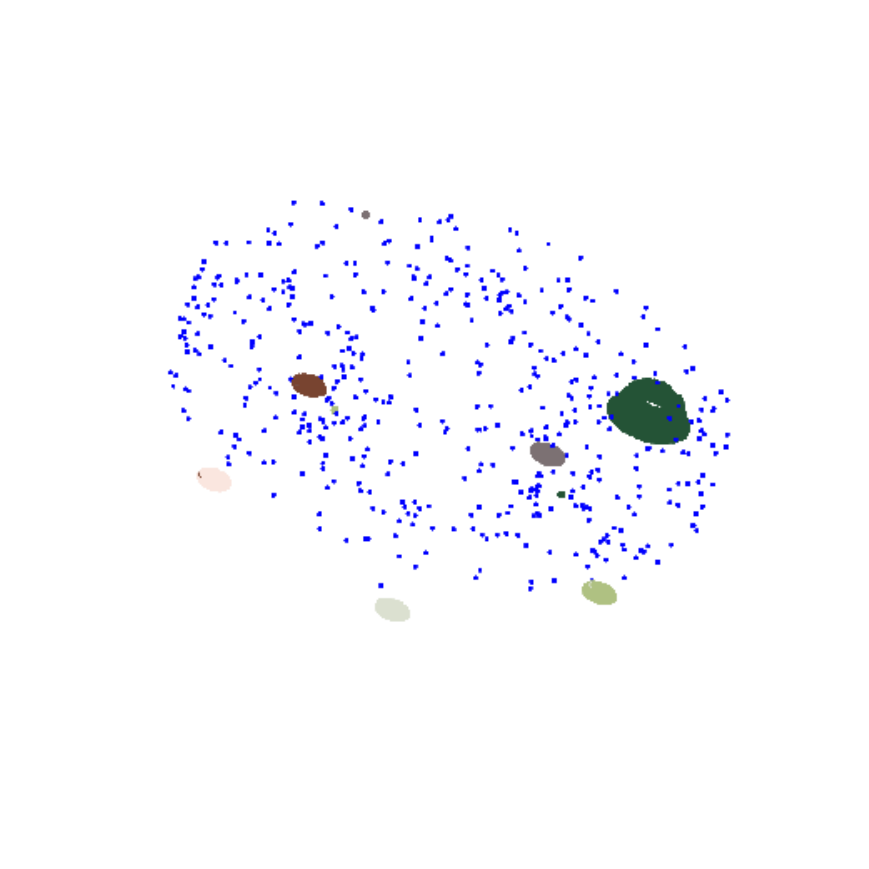}
            \\
            Original model
            &
            Same cylinder
            & Same cone
            & Same sphere
            \\\hline
             \includegraphics[scale=0.375, trim={1cm 1.5cm 1cm 1.25cm}, clip]{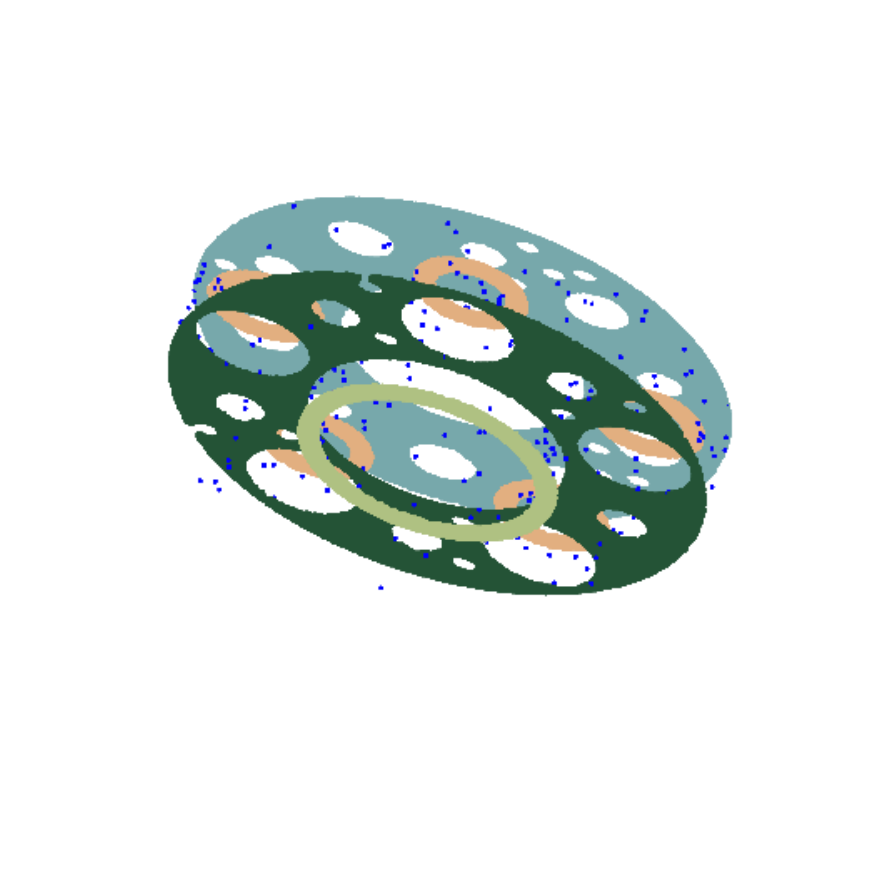}
             &
            \includegraphics[scale=0.375, trim={1cm 1.5cm 1cm 1.25cm}, clip]{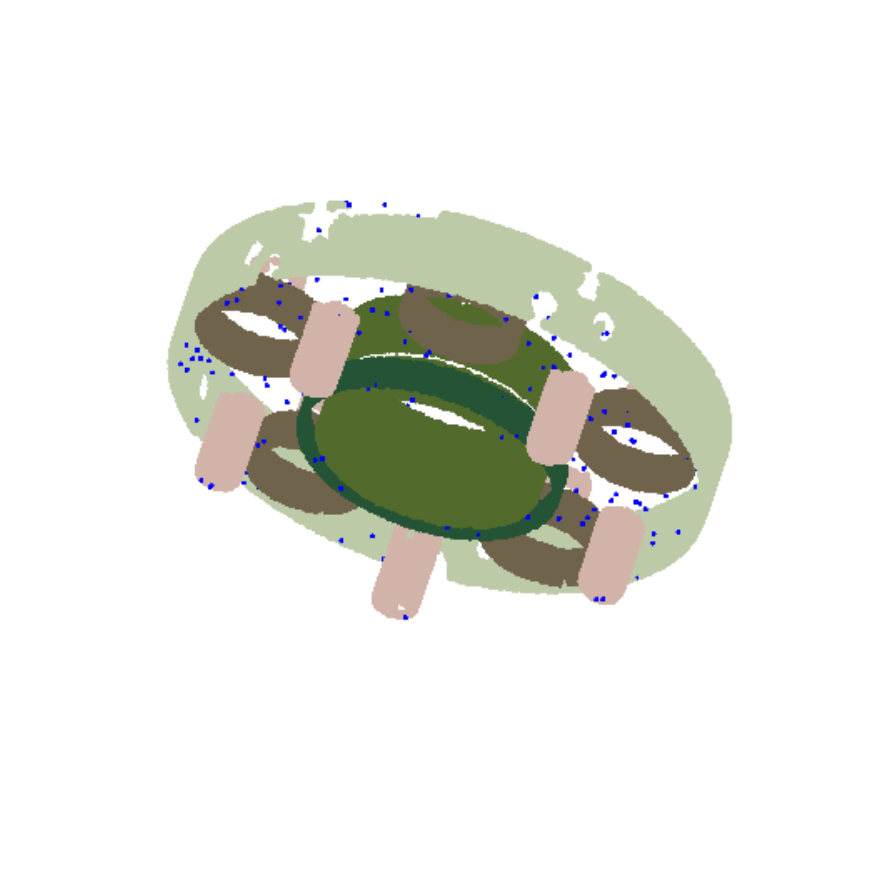}
            &
            \includegraphics[scale=0.375, trim={1cm 1.5cm 1cm 1.25cm}, clip]{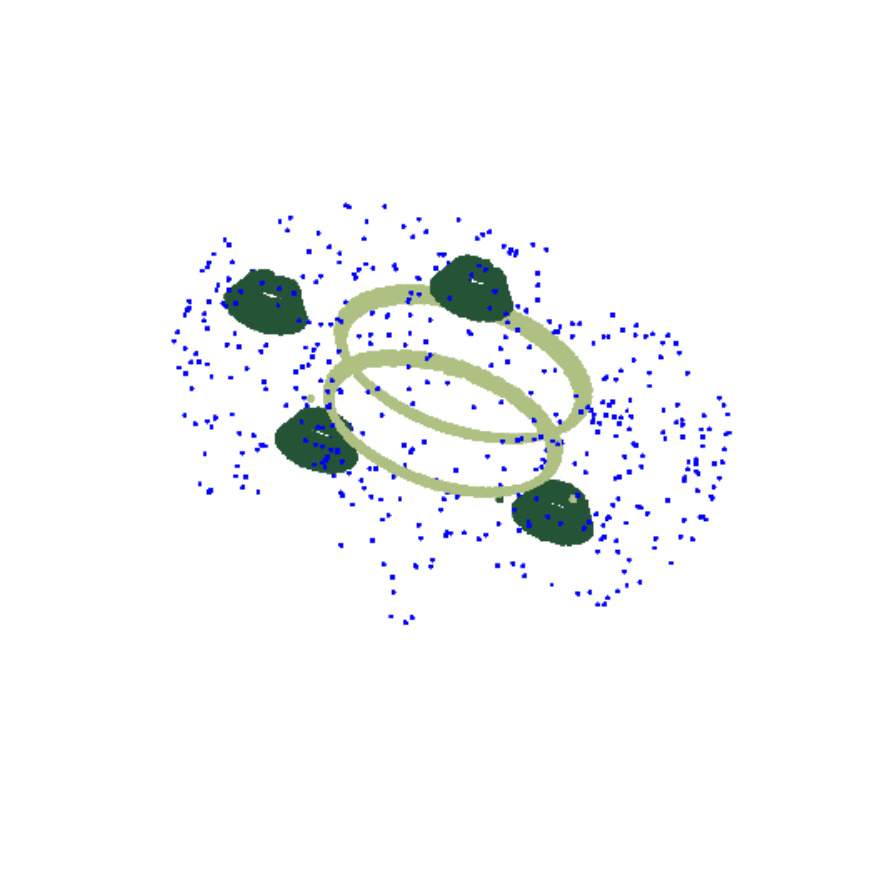}
            &
            \includegraphics[scale=0.375, trim={1cm 1.5cm 1cm 1.25cm}, clip]{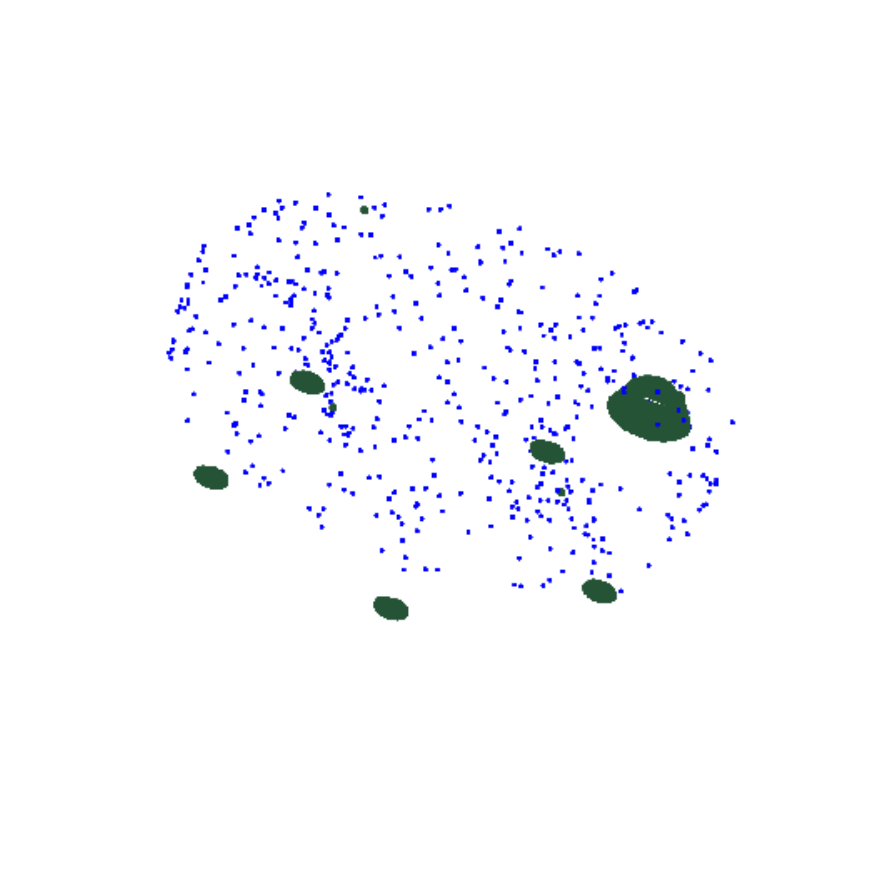}
            \\
            Same plane
            &
            Same radius
            &
            Same radius
            &
            Same radii
            \\\hline
            \includegraphics[scale=0.375, trim={1cm 1.5cm 1cm 1.25cm}, clip]{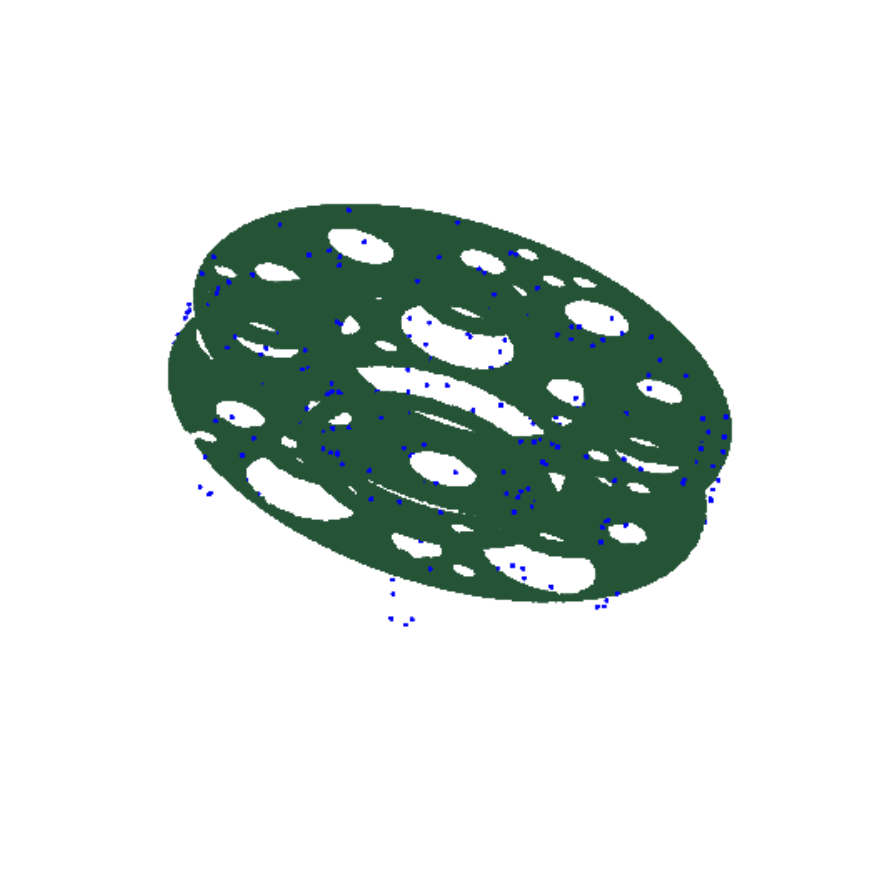}
             &
            \includegraphics[scale=0.375, trim={1cm 1.5cm 1cm 1.25cm}, clip]{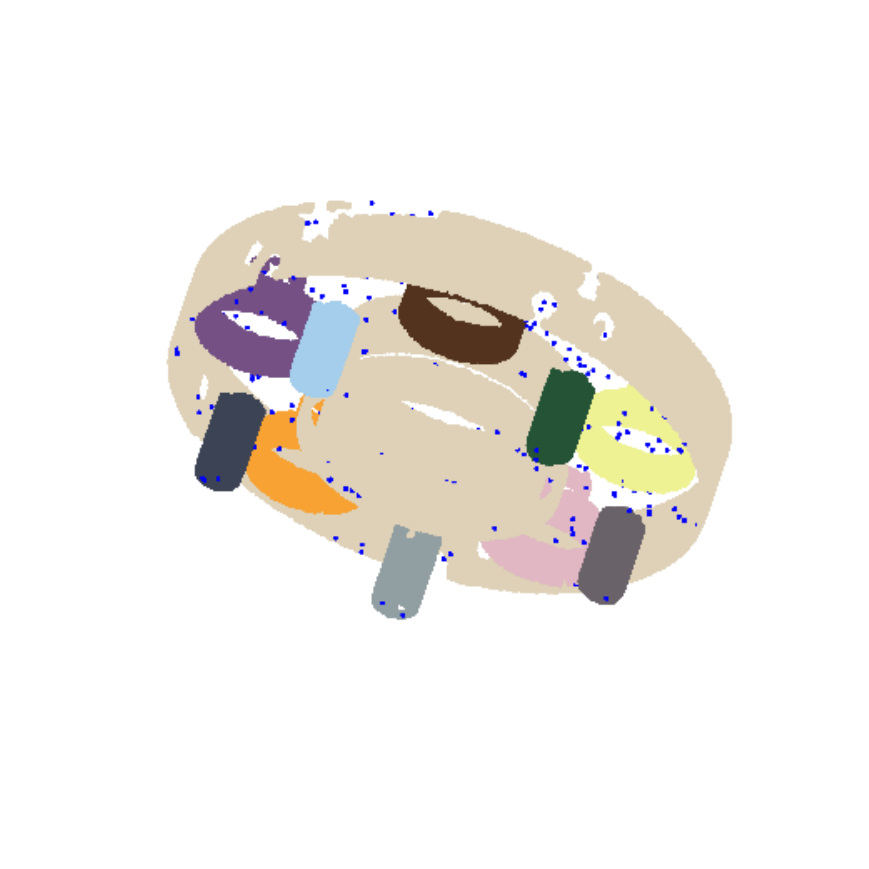}
            &
            \includegraphics[scale=0.375, trim={1cm 1.5cm 1cm 1.25cm}, clip]{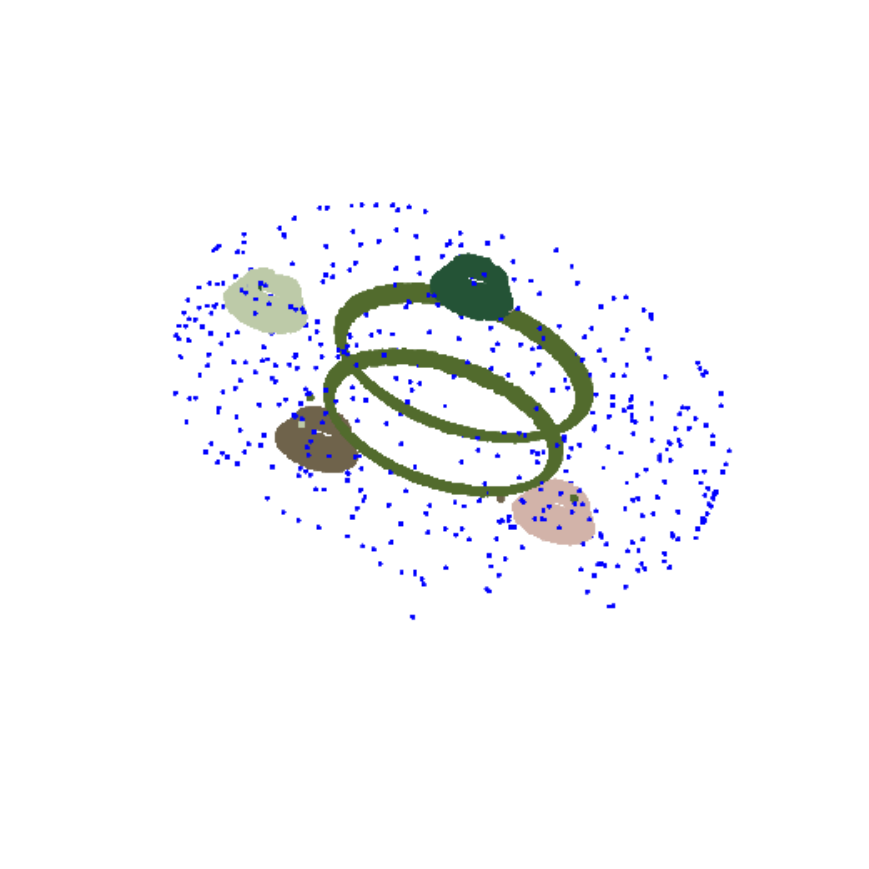}
            &
            \includegraphics[scale=0.375, trim={1cm 1.5cm 1cm 1.25cm}, clip]{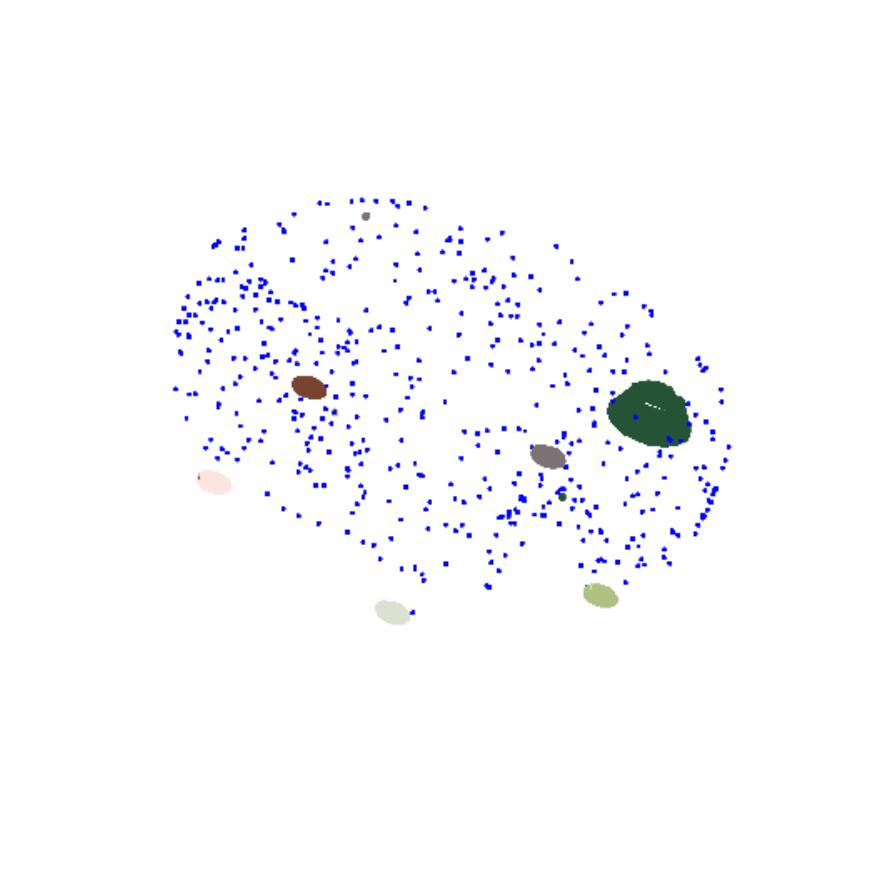}
            \\
            Parallel planes
            &
            \makecell{Same \\rotational axis}
            &
            \makecell{Same \\rotational axis}
            &
            Same center
            \\\hline
        \end{tabular}
    \end{center}
    \caption{Examples of queries for planar, cylindrical, conical and spherical segments obtained from a RANSAC segmentation of a scanned industrial object as provided in \cite{Li2011}. \label{fig:fig14_globfit}}
\end{figure}

We tested the method on two of the industrial scanned objects used in \cite{Li2011}. Unfortunately, no ground truth is available for these objects, and we will just report the average MFE. As a starting point we used the same input as \cite{Li2011}, i.e., RANSAC segmentations that were made available online by the authors in their GitHub page. The simplest example, shown in Figure \ref{fig:fig14_globfit}, contains $282,534$ points and $39$ segments. The main axes of planes, cylinders and cones are found to be parallel. Besides parallelism, this model presents also cylinders, cones and spheres with the same radius. Note that only one segment is wrongly labeled as a sphere instead of a truncated cone. Note that same consideration about axis aligned cylinders are provided in Figure 14 of \cite{Li2011}, but we are also able to find other types of correlations. For this point cloud, the mean of the MFE over all segments is $0.0091$.

\begin{figure*}[h!]
    \begin{center}
        \footnotesize
        \begin{tabular}{| >{\centering} z{2em} |c|c|c|}
            \hline
            \rotatebox[origin=c]{90}{\makecell{Model \\ and planes}}&
            \raisebox{-0.45\height}{\includegraphics[scale=0.5, trim={2.0cm 2.0cm 1.5cm 2.0cm}, clip]{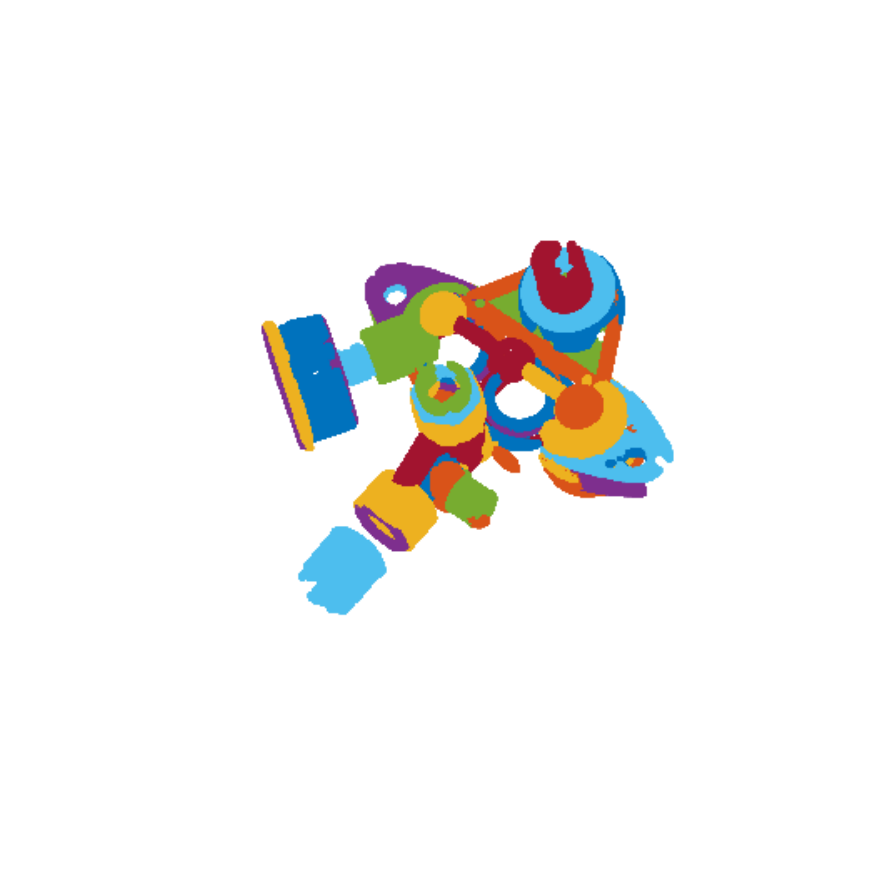}}
            &
            \raisebox{-0.45\height}{\includegraphics[scale=0.5, trim={2.0cm 1.5cm 1.5cm 1.5cm}, clip]{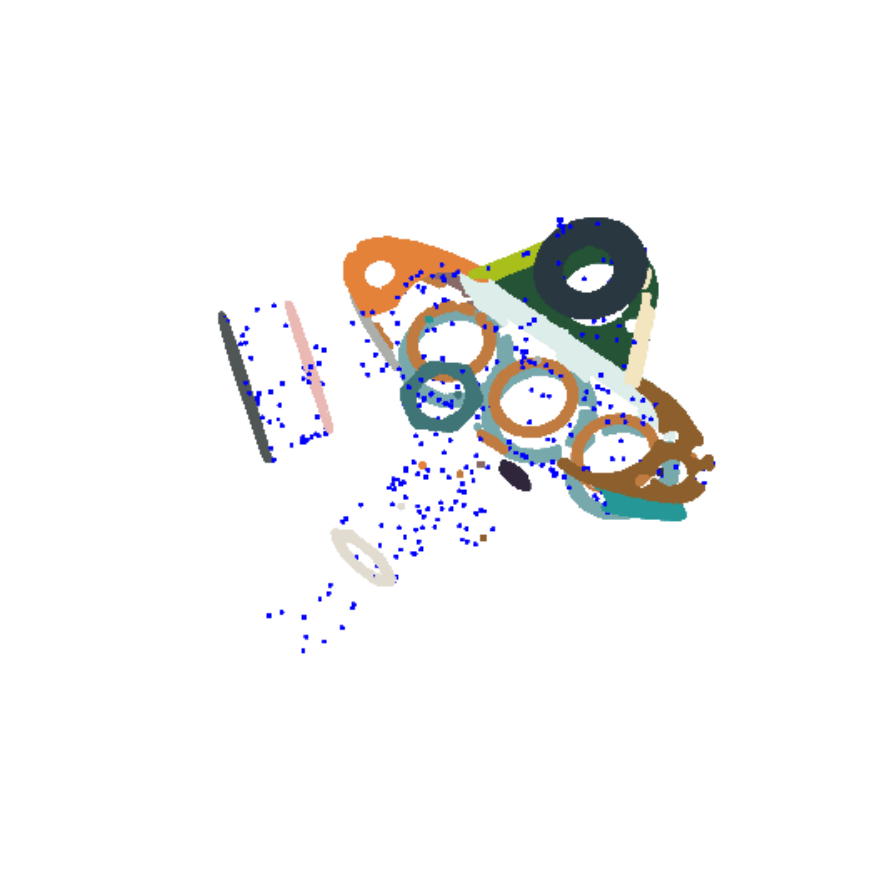}}
            &
            \raisebox{-0.45\height}{\includegraphics[scale=0.5, trim={2.0cm 1.5cm 1.5cm 1.5cm}, clip]{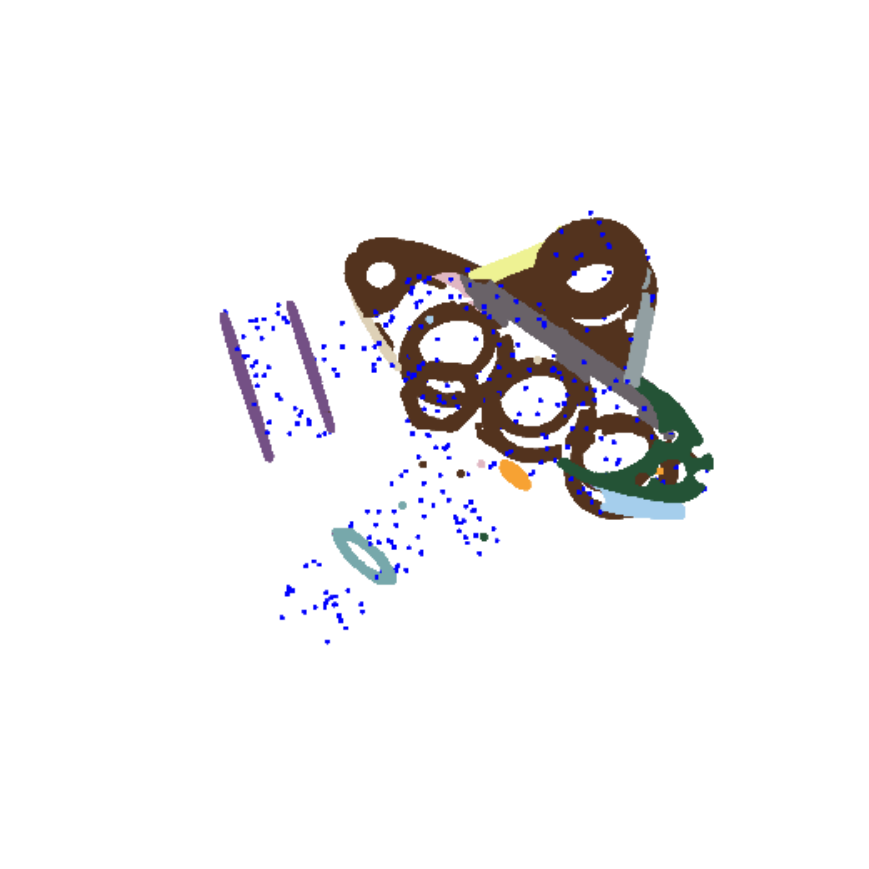}}
            \\
            & Original model & Same plane & Parallel planes
            \\\hline
            \rotatebox[origin=c]{90}{Cylinders}
            &
            \raisebox{-0.4\height}{\includegraphics[scale=0.45, trim={2.0cm 1.5cm 1.5cm 1.5cm}, clip]{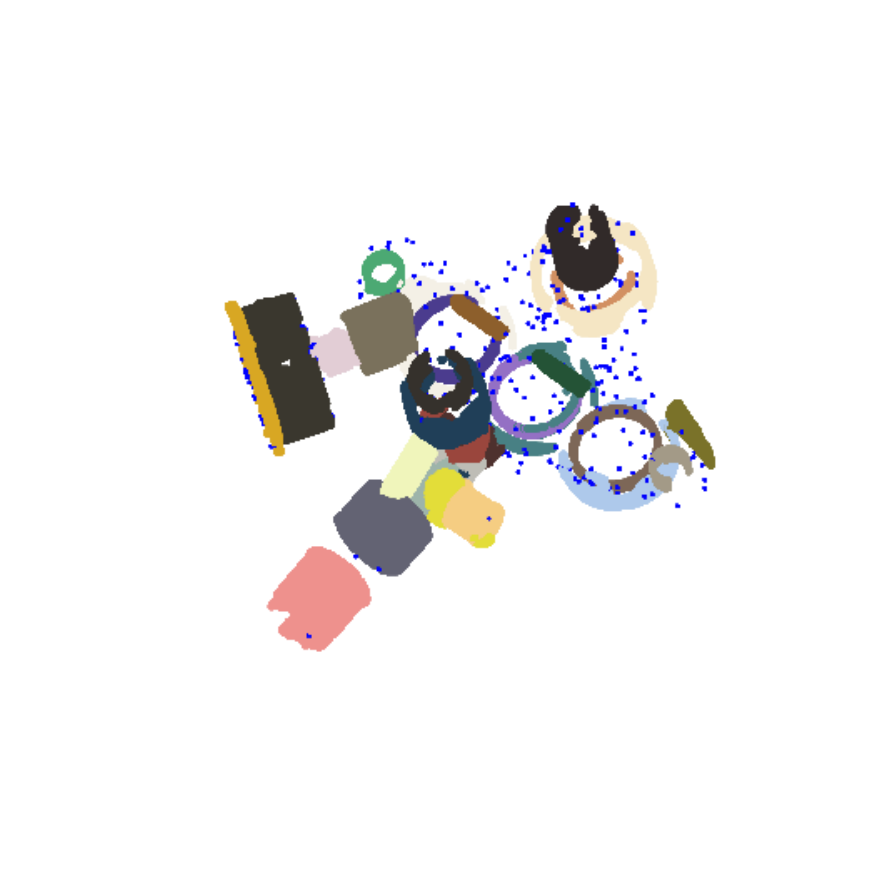}}
            &
            \raisebox{-0.4\height}{\includegraphics[scale=0.45, trim={2.0cm 1.5cm 1.5cm 1.5cm}, clip]{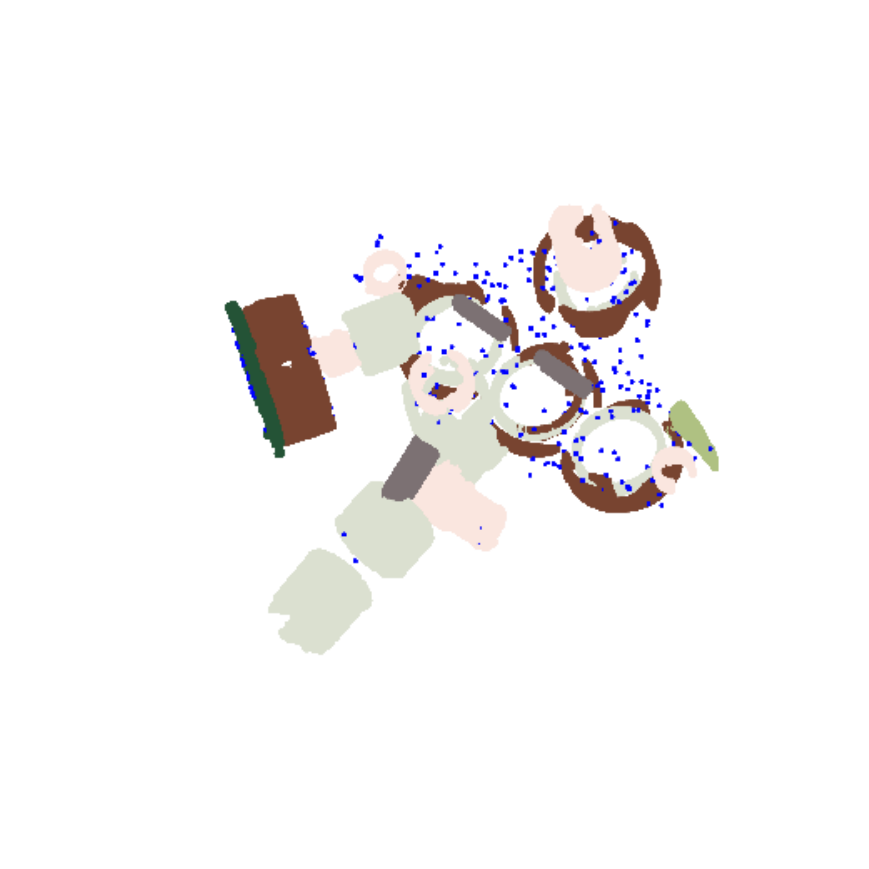}}
            & 
            \raisebox{-0.4\height}{\includegraphics[scale=0.45, trim={2.0cm 1.5cm 1.5cm 1.5cm}, clip]{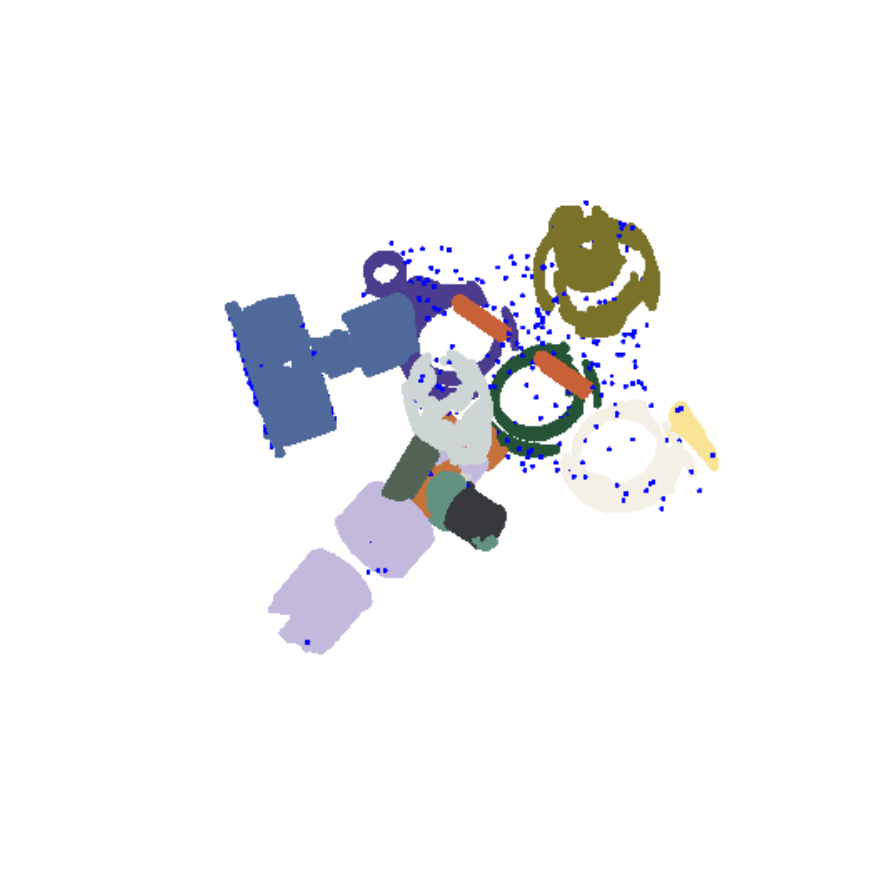}}
            \\
            &
            Same cylinder
            &
            Same radius
            &
            \makecell{Same \\rotational axis}
            \\\hline
            \rotatebox[origin=c]{90}{Spheres}
            &
            \raisebox{-0.4\height}{\includegraphics[scale=0.45, trim={2.0cm 1.5cm 1.5cm 1.5cm}, clip]{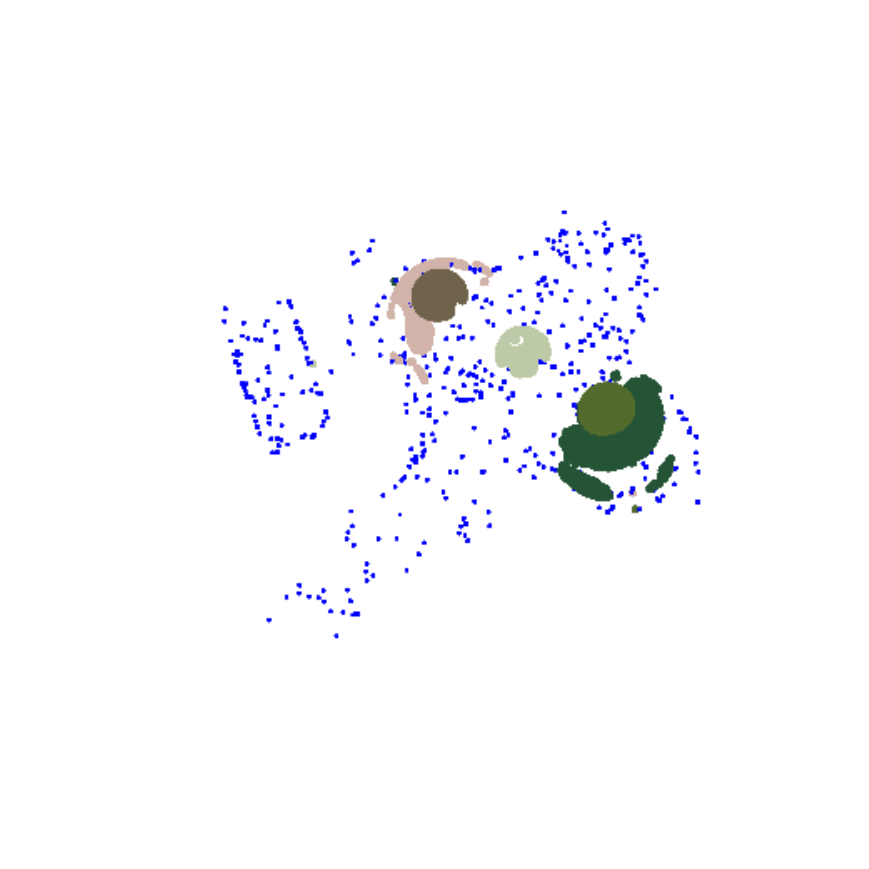}}
            &
            \raisebox{-0.4\height}{\includegraphics[scale=0.45, trim={2.0cm 1.5cm 1.5cm 1.5cm}, clip]{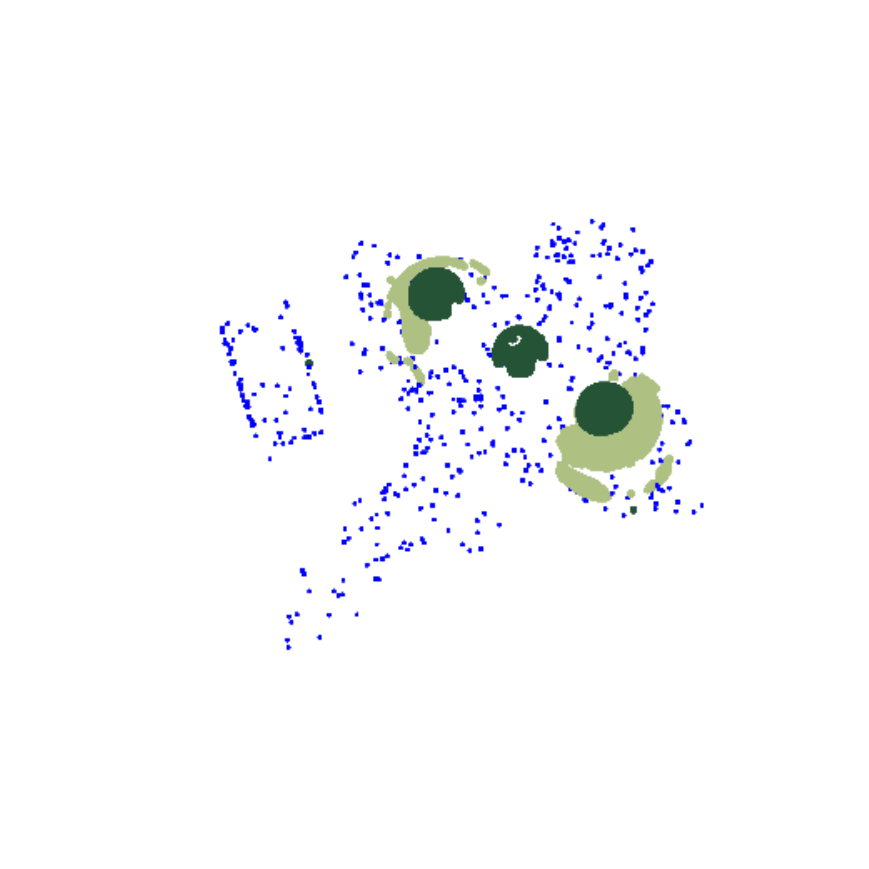}}
            & 
            \raisebox{-0.4\height}{\includegraphics[scale=0.45, trim={2.0cm 1.5cm 1.5cm 1.5cm}, clip]{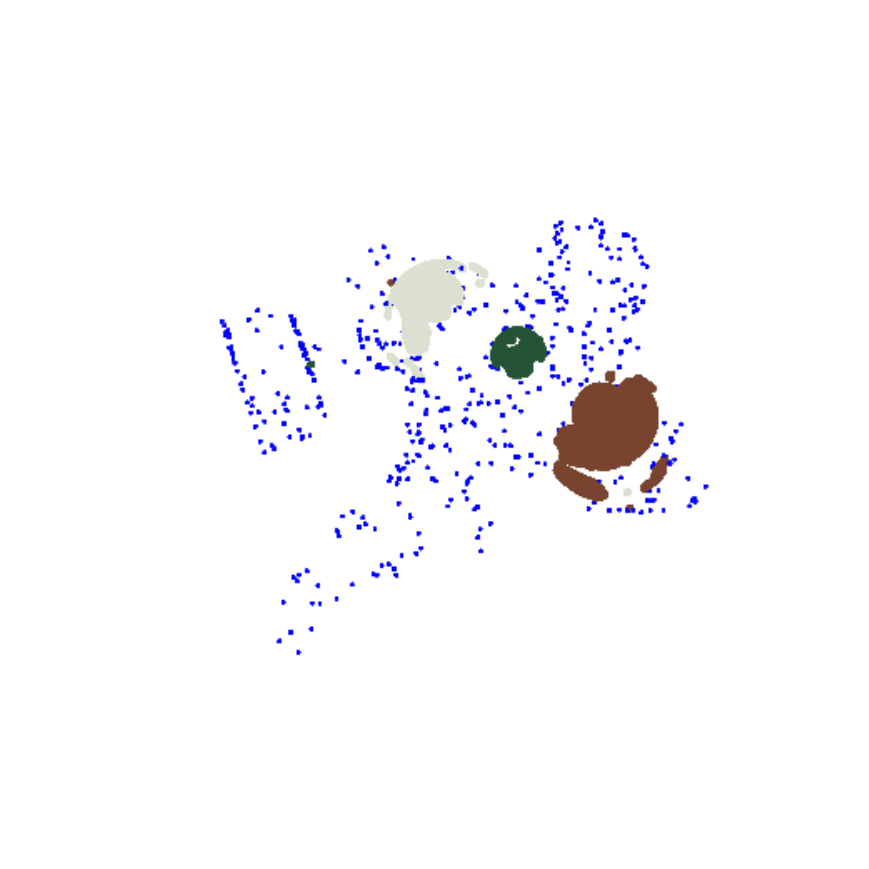}}
            \\
            &
            Same sphere
            &
            Same radius
            &
            Same center
            \\\hline
        \end{tabular}
    \end{center}
    \caption{Examples of queries for planar, cylindrical and spherical segments obtained from the RANSAC segmentation over a challenging point cloud acquired from an industrial object as provided in \cite{Li2011}. \label{fig:fig13_globfit}}
\end{figure*}

The point set in Figure \ref{fig:fig13_globfit} contains $529,006$ points for a total of $51$ segments, and corresponds to a machined part. Not only is our approach able to find segments lying on the same plane or cylinder, but also to identify cylinders having same radii or axes and spheres characterized by the same centers. The mean of the MFE over all segments is $0.0120$.

\section{Conclusions}
In this paper we have proposed a new method for fitting, recognizing and clustering  geometric primitives in segmented 3D point clouds. Our approach is based on the Hough Transform. In  particular, our method is able to use the HT also for non-trivial primitives, reducing its computational complexity.
For each segment we are able to provide a parametric representation and a number of parameters (centres, axes, vertices, etc.) that uniquely define it, thus feeding the clustering algorithm with a reliable segment description. 
As the method is applied to segmented point clouds,  the better the pre-segmentation technique, the better the results obtained. In any case, it concurs to improve the model in case of over-segmentation. Since the method is based on an aggregation technique, in the case of under-segmentation, it would be necessary to integrate it with an adaptive splitting strategy.

Thanks to the devised segment pre-processing technique, we are able to fit geometric primitives in a standard form (thus limiting the number of parameters needed by the HT) as well as to provide an estimate of the true solution (thus limiting the search for the optimal solution to a specific region which, in turn, allows us to solve the problem of unboundedness of the parameter space).
Our experiments confirm the robustness of the HT-based method to deal with various types of point cloud artifacts.

\begin{figure}[h!]
    \centering
    \begin{tabular}{cc}
     \includegraphics[width=3.5cm]{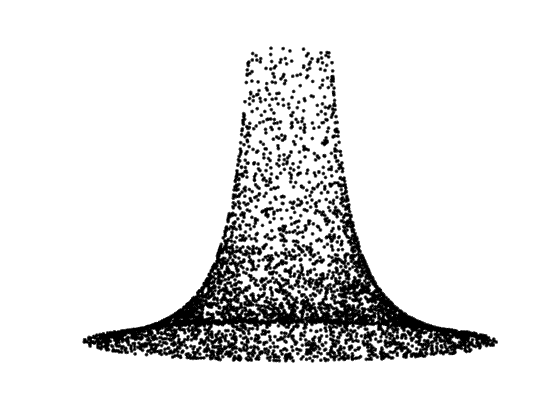} &
     \includegraphics[width=3.5cm]{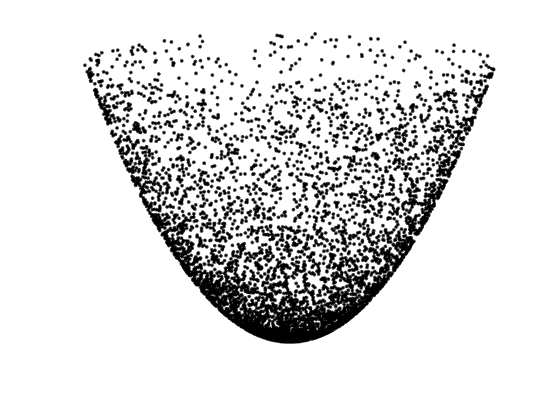} \\
    \end{tabular}
    \caption{Examples of surfaces of revolution.}
    \label{fig:surfRev}
\end{figure}

Our method currently deals with basic geometric primitives (planes, cylinders cones spheres and tori); however, even if we did not exhibit concrete examples, it is already able to deal with cylinders with an elliptic generatrix and ellipsoids, and it can be extended to surfaces of revolution,  as those shown in 
Figure \ref{fig:surfRev}.

\begin{figure}[h!]
    \centering
    \begin{tabular}{cccc}
     \includegraphics[width=3.2cm]{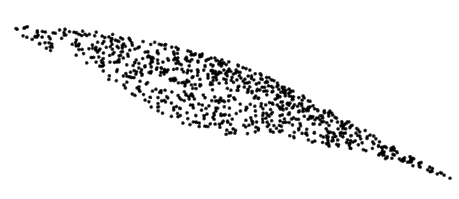} &
     \includegraphics[width=2.5cm]{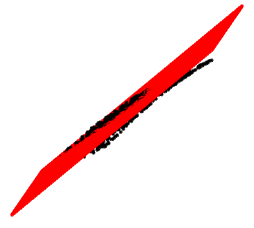} &
     \includegraphics[width=2.5cm]{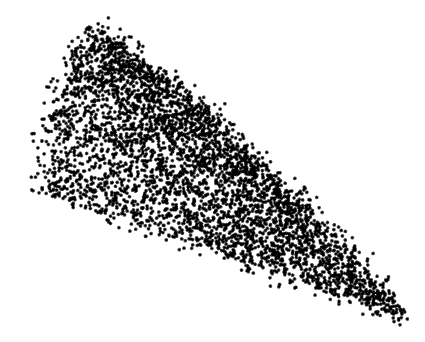} &
     \includegraphics[width=2.5cm]{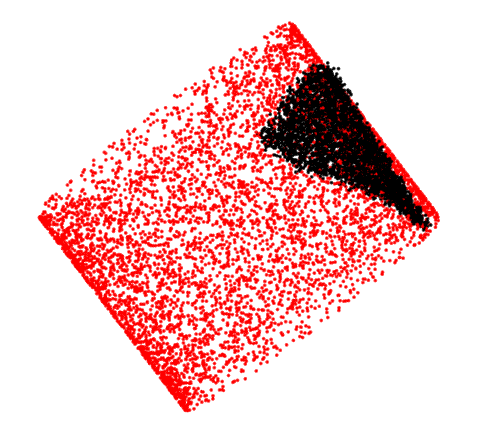}\\
     (a)& (b)& (c) & (d)\\
    \end{tabular}
    \caption{Examples of misclassified small pieces of primitives: the small cylindrical segment in (a) is confused with a plane in (b); the small conical segmenent in (c) is confused with a cylinder in (d).}
    \label{fig:smallPiece}
\end{figure}

Regarding the limitations of the method, when segments are small or lie on curved surfaces with a large radius, ambiguities on the primitive classification or instability in the estimation of the rotation or translation in the pre-processing step can arise. Figure \ref{fig:smallPiece} (a,c) shows two examples of possible misclassifications of small pieces of primitives: in (b) the part of a cylinder with a large radius recognized is fitted with a plane and in (d) a part of a cone affected by noise is fitted with a cylinder.

This is also the case, for instance, of segments classified as cylinders rather than tori,  when one radius of a torus is considerably larger than the other and therefore the segment is almost flat in one direction. 
Similar ambiguities arise between planes and curved surfaces, in particular when data are quite perturbed. Finally, even a strong under-sampling can become a source of ambiguity; for instance, when dealing with curved surfaces, it is necessary to have enough samples to correctly infer the curvature radius. 

\section*{Acknowledgements}
This work has been supported by the CNR research activities DIT.AD004.100, DIT.AD021.080.001 and DIT.AD021.125.

\bibliographystyle{ieeetr}      

\end{document}